%% file: latex/acl_latex.tex
\pdfoutput=1
\documentclass[11pt]{article}

\usepackage{times}
\usepackage{latexsym}
\usepackage{booktabs}
\usepackage{graphicx}
\usepackage{fontawesome5}
\usepackage{placeins}

\usepackage{rotating}
\usepackage{multirow}
\usepackage{url}
\usepackage{hyperref}
\usepackage[T1]{fontenc}
\usepackage[utf8]{inputenc}

\usepackage{microtype}

\usepackage{inconsolata}

\usepackage{graphicx}
\usepackage[most]{tcolorbox}
\usepackage{xcolor}

\definecolor{findinggreen}{HTML}{E8F9E8}
\newcommand{\researchfinding}[1]{%
  \begin{findingbox}
    \faLightbulb[regular]~#1
  \end{findingbox}
}

\newtcolorbox{findingbox}[1][]{
    breakable,
    enhanced,
    sharp corners,
    boxrule=0pt,
    colback=findinggreen,
    colframe=findinggreen,
    frame hidden,
    borderline west={2pt}{0pt}{green!70!black},
    left=6pt,
    right=6pt,
    top=4pt,
    bottom=4pt,
    before skip=10pt,
    after skip=10pt,
    fontupper=\linespread{1.0}\selectfont,
    #1
}

\usepackage[table]{xcolor}
\definecolor{cvprblue}{RGB}{0,114,225}
\definecolor{cvprred}{RGB}{241,159,158}% 一种常见的蓝色
\definecolor{good}{RGB}{0,150,0}  
\usepackage[most]{tcolorbox}
\usepackage{booktabs}
\usepackage{pifont}  % for checkmarks and crosses
\usepackage{booktabs}
\usepackage{xcolor}
\usepackage[dvipsnames]{xcolor}
\usepackage{makecell}
\usepackage{booktabs}
\usepackage{siunitx}
\usepackage{amssymb}

\usepackage{multirow}
\usepackage[final]{acl}
\usepackage[ruled,vlined]{algorithm2e}
\usepackage{times}
\usepackage{latexsym}
\usepackage{amsmath}
\usepackage{supertabular}

\usepackage{graphicx}  % 导言区
\usepackage[T1]{fontenc}
\usepackage[utf8]{inputenc}

\usepackage{microtype}

\usepackage{inconsolata}
\def \VersionWithComments {}
\ifdefined 
\VersionWithComments
\usepackage{marginnote}

\makeatletter
\DeclareRobustCommand\onedot{\futurelet\@let@token\@onedot}
\def\@onedot{\ifx\@let@token.\else.\null\fi\xspace}

\makeatother
\usepackage{tikz}
\newcommand*\circled[1]{\tikz[baseline=(char.base)]{\node[shape=circle,fill=black,text=white,draw,inner sep=.1pt] (char) {#1};}}

\title{Reason Analogically via Cross-domain Prior Knowledge: An Empirical Study of Cross-domain Knowledge Transfer for In-Context Learning}

\author{
  Le Liu\textsuperscript{1,2}\thanks{Equal Contribution},
  Zhiming Li\textsuperscript{2}\footnotemark[1],
  Jianzhi Yan\textsuperscript{1,2},
  \textbf{Zike Yuan\textsuperscript{1,2}},
  Shiwei Chen\textsuperscript{1,2},
  \textbf{Youcheng Pan\textsuperscript{2}},\\
  \textbf{Buzhou Tang\textsuperscript{1,2}},
  \textbf{Qingcai Chen\textsuperscript{2}}\thanks{Corresponding author},
  \textbf{Yang Xiang\textsuperscript{2}}\footnotemark[2],
  \textbf{Danny Dongning Sun\textsuperscript{2}}\footnotemark[2],\\
  \textsuperscript{1}Harbin Institute of Technology, Shenzhen, China \\
  \textsuperscript{2}Pengcheng Laboratory, Shenzhen, China \\
  }

\begin{document}
\maketitle

\input{latex/tex/abstract}

\input{latex/tex/intro}

\input{latex/tex/preli}

\input{latex/tex/method}

\input{latex/tex/setup}

\input{latex/tex/result}

\input{latex/tex/related}

% \begin{figure}[t!]
%     \centering
%     \includegraphics[width=\linewidth]{figs/wenn_plot.pdf}
%     \caption{\textbf{Sources of improvement in cross-domain in-context learning.} The outer ellipse denotes all 0-shot errors, while the inner circles represent those corrected by Embedding or BM25 retrieval.}
%     \label{fig:wenn_plot}
% \end{figure}

\input{latex/tex/conclusion}

\clearpage
\input{latex/tex/limitations}

\bibliography{custom}

\clearpage
\input{latex/tex/appendix}

\end{document}

%% file: latex/tex/abstract.tex
\begin{abstract}
Despite its success, existing in-context learning (ICL) relies on in-domain expert demonstrations, limiting its applicability when expert annotations are scarce. We posit that different domains may share underlying reasoning structures, enabling source-domain demonstrations to improve target-domain inference despite semantic mismatch. To test this hypothesis, we conduct a comprehensive empirical study of different retrieval methods to validate the feasibility of achieving cross-domain knowledge transfer under the in-context learning setting. Our results demonstrate conditional positive transfer in cross-domain ICL. We identify a clear example absorption threshold: beyond it, positive transfer becomes more likely, and additional demonstrations yield larger gains. Further analysis suggests that these gains stem from reasoning structure repair by retrieved cross-domain examples, rather than semantic cues. Overall, our study validates the feasibility of leveraging cross-domain knowledge transfer to improve cross-domain ICL performance, motivating the community to explore designing more effective retrieval approaches for this novel direction.\footnote{Our implementation is available at \url{https://github.com/littlelaska/ICL-TF4LR}}
\end{abstract}

%% file: latex/tex/intro.tex
\section{Introduction}

In-context learning (ICL)\cite{brown2020lang,radford2019language} allows LLMs to adapt to new tasks using only a small set of demonstrations, without any parameter fine-tuning \cite{dong-etal-2024-survey}. This flexibility has made ICL a central paradigm, motivating extensive in-domain studies on prompt design, example selection, and robustness to mild distribution shifts \cite{mueller2023context,zhou2023explore,wei2022chain,lewkowycz2022solving,tang2023lazy,sun2024shortcut,siska-etal-2024-examining,yuan-etal-2024-llms,honda-oka-2025-exploring,he-etal-2024-using}. However, most existing studies assume access to high-quality in-domain demonstrations curated by human experts. In practice, such annotations are often scarce or unavailable; a natural idea would be to leverage available prior knowledge which is embedded in other domains to improve performance.

As is often the case, different domains can share similar task-solving rationale despite substantial surface-level differences \cite{Besta_2024, Besta_2025, bu-etal-2025-enhanced,zhang2024pathofthoughtsextractingfollowingpaths,li2024surveygraphmeetslarge}. Consequently, a well-designed retrieval method is indispensable for enabling effective transfer learning for ICL. To achieve this goal, retrieval methods must address two major outstanding challenges:
\begin{itemize}
    \item \textcolor{red}{\textbf{Retrieval Expressiveness}}: Is the retrieval method sufficiently expressive to retrieve cross-domain demonstrations that share similar task-solving rationales?
    \item \textcolor{red}{\textbf{Transfer Stability}}: Does the retrieval method remain consistently effective across different large language models (LLMs)?
\end{itemize}
To demonstrate the feasibility of cross-domain transfer learning for LLMs and explore its future potential, we conduct a large-scale empirical study of existing, widely used retrieval methods for cross-domain in-context learning under a unified evaluation framework. Specifically, top-k demonstrations retrieved from source tasks are used to prompt a frozen LLM for target queries. Across six reasoning benchmarks and multiple model families and scales, we evaluate performance using Exact Match and gains over zero-shot. Our study is structured around four core research questions: \circled{1} How is the overall performance of cross-domain knowledge transfer via ICL? \circled{2} How does model scale influence cross-domain transferability?—We analyze how smaller and larger models differ in their ability to leverage retrieved demonstrations, including susceptibility to negative transfer. \circled{3} How does the number of shots (ie, demonstrations) affect cross-domain transferability?—We examine whether increasing shot counts consistently improves performance or instead introduces instability and interference. \circled{4} What's the source of improvement?—We unveil that the source of improvement stems from models' rectification of reasoning structure based on cross-domain demonstrations.

The contributions of this work are summarized as follows.
\begin{itemize}
    \item We conduct a systematic empirical study that examines the effectiveness of different retrieval approaches, their applicability across model families, and their transferability across domains in the setting of cross-domain in-context learning.

    \item We conduct detailed ablation analysis regarding the number of demonstrations and model size, which provides an in-depth understanding of the influence of different variables involved.

    \item To the best of our knowledge, this is the first trial to explore the feasibility of cross-domain transfer learning via in-context learning.
    These comprehensive evaluations may motivate the community to explore more effective retrieval methods that explicitly model logical structures and better support cross-domain generalization.
\end{itemize}

%% file: latex/tex/preli.tex
\section{Preliminaries}

\begin{figure*}[t!]
  \includegraphics[scale=0.31]{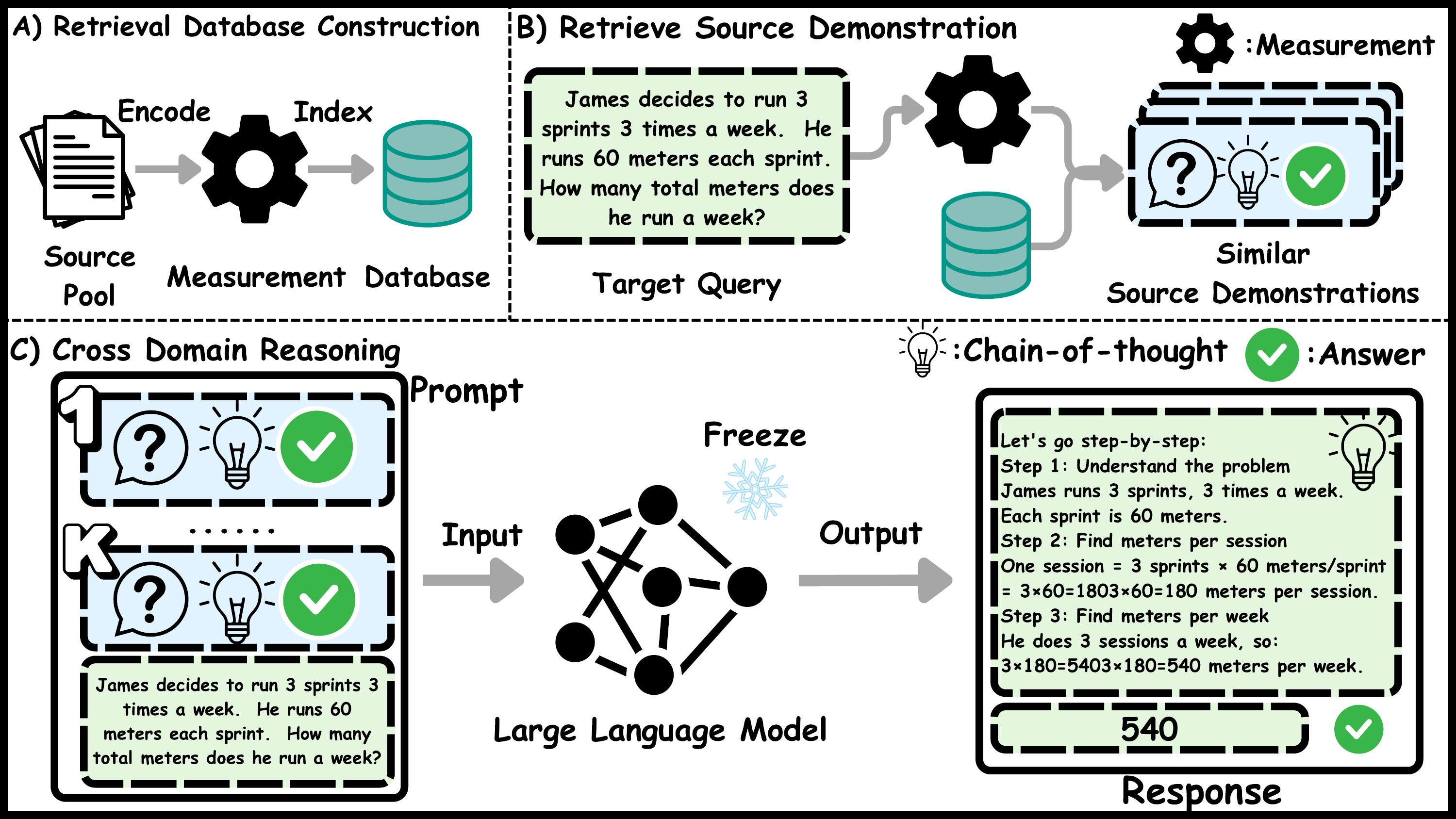}
  \caption{\textbf{Overview of our cross-domain ICL evaluation workflow.} (\textbf{A}) A source demonstration pool is encoded and indexed into a retrieval database. (\textbf{B}) For a target query, the system retrieves semantically similar source demonstrations. (\textbf{C}) Retrieved demonstrations are composed into a prompt and fed to a frozen LLM to produce step-by-step reasoning and the final answer.}
  \label{fig:overview}
\end{figure*}

\subsection{Transfer Learning}
Transfer learning studies how knowledge from a source dataset can be leveraged to improve performance on a target dataset, particularly when labeled target data are scarce. 
Let $\mathcal{D}_S = \{(x_i^S, y_i^S)\}_{i=1}^{n_S}$ and 
$\mathcal{D}_T = \{(x_j^T, y_j^T)\}_{j=1}^{n_T}$ denote the source and target datasets, where $(x,y)\in\mathcal{X}\times\mathcal{Y}$.
In general, the source and target data may follow different input distributions,
\[
P_S(x) \neq P_T(x),
\]
and may correspond to different prediction functions.

The goal of transfer learning is to exploit information learned from $\mathcal{D}_S$ to improve generalization on $\mathcal{D}_T$. 
A common approach is to map both source and target inputs into a shared representation space via an encoder $h_\theta:\mathcal{X}\rightarrow\mathbb{R}^d$, such that transferable structures are preserved:

\begin{equation}
\begin{aligned}
\min_{\theta}\quad
& \mathcal{L}_S\big(f_\theta(h_\theta(x^S)), y^S\big) \\
& + \lambda\,\mathcal{R}\!\left(
    \{h_\theta(x_i^S)\},
    \{h_\theta(x_j^T)\}
\right).
\end{aligned}
\end{equation}

where $\mathcal{L}_S$ is the source supervision loss and $\mathcal{R}$ encourages cross-dataset transferability.

\subsection{Retrieval-Augmented Generation}
Retrieval-Augmented Generation enhances model inference by retrieving relevant documents from a corpus $\mathcal{D}={d_i}_{i=1}^N$ via vector similarity:
\[
\phi: \mathcal{D} \rightarrow \mathbb{R}^k,
\]
and indexed into a retrieval database $\mathcal{I} = \{\phi(d_i)\}_{i=1}^{N}$. The top-$K$ documents are retrieved based on the query representation $\phi(q)$:
\[
\mathcal{R}(q) = \operatorname*{arg\,topK}_{d_i \in \mathcal{D}} \; \mathrm{sim}(\phi(q), \phi(d_i)),
\]
where $\mathrm{sim}(\cdot,\cdot)$ denotes a similarity measure such as inner product or cosine similarity.

The retrieved documents $\{d_{(1)}, \dots, d_{(K)}\}$ are combined with the query via a direct sum operation to form the augmented input:
\[
\tilde{q} = d_{(1)} \oplus d_{(2)} \oplus \cdots \oplus d_{(K)} \oplus q,
\]
which is then fed into the language model to produce the final prediction:
\[
y = f_\theta(\tilde{q}).
\]
In this framework, retrieval serves as an explicit information access mechanism, enabling the model to condition its reasoning on external evidence rather than relying solely on parametric knowledge.

\subsection{In-Context Learning}
In-context learning (ICL) allows LLMs to infer new tasks from contextual examples \cite{wei2022chain, brown2020lang}. Unlike in-weights learning, which relies on gradient-based parameter updates, ICL adapts behavior without modifying model weights.

Formally, each training instance is linearized into an input sequence $\mathbf{x} = (x_1,\ldots,x_{|\mathbf{x}|})$ and an output sequence $\mathbf{y} = (y_1,\ldots,y_{|\mathbf{y}|})$, where each token belongs to the model vocabulary $\mathcal{V}$.  
Given a test input $\mathbf{x}_{\text{test}}$, in-context learning defines its prediction as
\[
\mathbf{y}_{\text{test}} \sim 
\mathcal{P}_{\text{LM}}\!\left(
\mathbf{y}_{\text{test}}
\;\middle|\;
\underbrace{
\mathbf{x}_1,\mathbf{y}_1,\ldots,\mathbf{x}_K,\mathbf{y}_K,\mathbf{x}_{\text{test}}
}_{\textbf{In-context prompt}}
\right),
\]
where the sampling operator denotes the decoding method. Each demonstration $e_i = (\mathbf{x}_i,\mathbf{y}_i)$ is drawn from a dataset
\[
\mathcal{D}=\{(\mathbf{x}_i,\mathbf{y}_i)\}_{i=1}^N.
\]

This formulation allows the model to condition on the provided examples without updating its parameters, enabling fast adaptation to new tasks without additional training cost.

%% file: latex/tex/method.tex
\input{latex/tex/table1}

\section{Method}

As shown in Figure~\ref{fig:overview}, we investigate the feasibility of cross-domain knowledge transfer using in-context learning through a three-step pipeline: embedding database construction, demonstration retrieval, and cross-domain inference.

\subsection{Retrieval Database Construction}
As shown in Part A of Figure~\ref{fig:overview}, given a \textbf{source-domain dataset} $\mathcal{D}_S = \{(x_i^S, y_i^S)\}$, we construct a retrieval metric which embeds each instance to a corresponding representation for the downstream retrieval function to each instance:
\[
m_i = \mathcal{M}(x_i^S, y_i^S),
\]
where $\mathcal{M}(\cdot)$ denotes a retrieval metric function that maps each source example into a searchable form.  
The resulting database is defined as
\[
\mathcal{B}_S = \{(m_i, x_i^S, y_i^S)\},
\]
which supports efficient retrieval under different similarity or matching criteria.

\subsection{Demonstration Retrieval}

As shown in Part B of Figure~\ref{fig:overview}, given a target query $x^T_{test}$, we first compute its corresponding representation
\[
m^T_{test} = \mathcal{M}(x^T_{test}),
\]
where $\mathcal{M}(\cdot)$ denotes the same measurement function used in database construction.

We retrieve source demonstrations by ranking all indexed instances according to a similarity function $s(m^T_{test}, m_i)$ and selecting the top-$k$ results:
\[
\mathcal{R}(x^T_{test}) = \operatorname{TopK}_{s(\cdot,\cdot)}\left(m^T_{test}, \mathcal{B}_S\right).
\]

% The similarity function $s(\cdot,\cdot)$ supports dense, sparse, and hybrid retrieval mechanisms.

The similarity function $s(\cdot,\cdot)$ is used to measure the semantic distance between query and documents; inner-product, cosine similarity, and BM25 are all viable candidates.

\subsection{Cross-domain ICL}
Finally, the retrieved source demonstrations $\{(x_i^{S}, y_i^{S})\}_{i=1}^k$ and the target query $x^T_{\text{test}}$ are concatenated to form the input prompt, the model prediction follows the standard ICL formulation (Part C in Figure~\ref{fig:overview}):

\begin{equation}
\hat{y} \sim 
\mathcal{P}_{\text{LM}}
\left(
\underbrace{[(x_1^{S}, y_1^{S}), \dots, (x_k^{S}, y_k^{S})]}_{\textbf{Source Domain}},
\ \underbrace{x^T_{\text{test}}}_{\textbf{Target Domain}}
\right)
\end{equation}

We evaluate accuracy on $\mathcal{D}_T$ under varying source domains, target domains, shot counts, and model sizes. This framework enables controlled and systematic analysis of factors that influence cross-domain ICL performance.

%% file: latex/tex/table1.tex
\begin{table*}[h!]
\centering
\small
\setlength{\tabcolsep}{4pt}
\begin{tabular}{lllcccccc}
\toprule
\textbf{Model} & \textbf{Source} & \textbf{Method} & \textbf{ProntoQA} & \textbf{FOLIO} & \textbf{ProofWriter} & \textbf{\makecell{Logical\\Deduction}} & \textbf{AR-LSAT} & \textbf{GSM8K} \\
\midrule
\multirow{15}{*}{\textbf{Gemma3-4B}}
& \multirow{3}{*}{\textbf{ProntoQA}} & BM25 & \textemdash & \textbf{$60.8_{\cellcolor{red!15}\underline{-2.5}}$} & \textbf{$66.2_{\cellcolor{cyan!15}\textcolor{red}{+2.7}}$} & \textbf{$36.7_{\cellcolor{red!25}\underline{-19.7}}$} & \textbf{\textemdash} & \textbf{\textemdash} \\
&  & Embed & \textemdash & \textbf{$66.2_{\cellcolor{cyan!15}\textcolor{red}{+2.9}}$} & \textbf{$67.8_{\cellcolor{cyan!25}\textcolor{red}{+4.3}}$} & \textbf{$58.3_{\cellcolor{cyan!15}\textcolor{red}{+2.0}}$} & \textbf{$25.7_{\cellcolor{red!15}\underline{-2.6}}$} & \textbf{$84.8_{\cellcolor{red!5}\underline{-1.3}}$} \\
&  & ConE & \textemdash & \textbf{$60.8_{\cellcolor{red!15}\underline{-2.5}}$} & \textbf{$62.5_{\cellcolor{red!5}\underline{-1.0}}$} & \textbf{$63.3_{\cellcolor{cyan!25}\textcolor{red}{+7.0}}$} & \textbf{$26.1_{\cellcolor{red!15}\underline{-2.2}}$} & \textbf{$85.2_{\underline{-0.9}}$} \\
\cmidrule(lr){2-9}
& \multirow{3}{*}{\textbf{FOLIO}} & BM25 & \textbf{$93.2_{\cellcolor{cyan!5}\textcolor{red}{+1.2}}$} & \textemdash & \textbf{$63.0_{\cellcolor{cyan!5}\textcolor{red}{+1.3}}$} & \textbf{$41.3_{\cellcolor{red!25}\underline{-19.7}}$} & \textbf{$24.8_{\underline{-0.4}}$} & \textbf{$84.0_{\cellcolor{red!15}\underline{-2.4}}$} \\
&  & Embed & \textbf{$93.0_{\cellcolor{cyan!5}\textcolor{red}{+1.0}}$} & \textemdash & \textbf{$65.0_{\cellcolor{cyan!25}\textcolor{red}{+3.3}}$} & \textbf{$58.7_{\cellcolor{red!15}\underline{-2.3}}$} & \textbf{$27.0_{\cellcolor{cyan!5}\textcolor{red}{+1.7}}$} & \textbf{$85.0_{\cellcolor{red!5}\underline{-1.4}}$} \\
&  & ConE & \textbf{$94.4_{\cellcolor{cyan!15}\textcolor{red}{+2.4}}$} & \textemdash & \textbf{$64.3_{\cellcolor{cyan!15}\textcolor{red}{+2.7}}$} & \textbf{$62.7_{\cellcolor{cyan!5}\textcolor{red}{+1.7}}$} & \textbf{$30.0_{\cellcolor{cyan!25}\textcolor{red}{+4.8}}$} & \textbf{\textemdash} \\
\cmidrule(lr){2-9}
& \multirow{3}{*}{\textbf{ProofWriter}} & BM25 & \textbf{$95.2_{\cellcolor{cyan!15}\textcolor{red}{+2.4}}$} & \textbf{$67.6_{\cellcolor{cyan!25}\textcolor{red}{+5.4}}$} & \textemdash & \textbf{$41.3_{\cellcolor{red!25}\underline{-19.3}}$} & \textbf{$28.7_{\cellcolor{cyan!25}\textcolor{red}{+3.5}}$} & \textbf{$85.7_{\underline{-0.4}}$} \\
&  & Embed & \textbf{$95.6_{\cellcolor{cyan!15}\textcolor{red}{+2.8}}$} & \textbf{$65.2_{\cellcolor{cyan!15}\textcolor{red}{+2.9}}$} & \textemdash & \textbf{$61.0_{\textcolor{red}{+0.3}}$} & \textbf{$26.1_{\textcolor{red}{+0.9}}$} & \textbf{$84.5_{\cellcolor{red!5}\underline{-1.6}}$} \\
&  & ConE & \textbf{\textemdash} & \textbf{$65.2_{\cellcolor{cyan!15}\textcolor{red}{+2.9}}$} & \textemdash & \textbf{$62.3_{\cellcolor{cyan!5}\textcolor{red}{+1.7}}$} & \textbf{$27.0_{\cellcolor{cyan!5}\textcolor{red}{+1.7}}$} & \textbf{\textemdash} \\
\cmidrule(lr){2-9}
& \multirow{3}{*}{\textbf{LogicalDeduction}} & BM25 & \textbf{$65.4_{\cellcolor{red!25}\underline{-27.6}}$} & \textbf{$63.7_{\cellcolor{cyan!5}\textcolor{red}{+2.0}}$} & \textbf{$60.8_{\cellcolor{red!5}\underline{-1.8}}$} & \textemdash & \textbf{$27.0_{\underline{-0.9}}$} & \textbf{$84.1_{\cellcolor{red!5}\underline{-1.8}}$} \\
&  & Embed & \textbf{$90.6_{\cellcolor{red!15}\underline{-2.4}}$} & \textbf{$63.7_{\cellcolor{cyan!5}\textcolor{red}{+2.0}}$} & \textbf{$61.5_{\cellcolor{red!5}\underline{-1.2}}$} & \textemdash & \textbf{$26.1_{\cellcolor{red!5}\underline{-1.7}}$} & \textbf{$84.5_{\cellcolor{red!5}\underline{-1.4}}$} \\
&  & ConE & \textbf{$89.2_{\cellcolor{red!25}\underline{-3.8}}$} & \textbf{$61.8$} & \textbf{$56.2_{\cellcolor{red!25}\underline{-6.5}}$} & \textemdash & \textbf{$28.3_{\textcolor{red}{+0.4}}$} & \textbf{\textemdash} \\
\cmidrule(lr){2-9}
& \multirow{3}{*}{\textbf{GSM8K}} & BM25 & \textbf{$86.6_{\cellcolor{red!25}\underline{-6.0}}$} & \textbf{$62.7_{\textcolor{red}{+1.0}}$} & \textbf{$60.5_{\cellcolor{red!25}\underline{-3.8}}$} & \textbf{$62.7$} & \textbf{$26.1_{\cellcolor{red!15}\underline{-2.6}}$} & \textemdash \\
&  & Embed & \textbf{$86.6_{\cellcolor{red!25}\underline{-6.0}}$} & \textbf{$60.8_{\underline{-1.0}}$} & \textbf{$58.2_{\cellcolor{red!25}\underline{-6.2}}$} & \textbf{$63.0_{\textcolor{red}{+0.3}}$} & \textbf{$27.4_{\cellcolor{red!5}\underline{-1.3}}$} & \textemdash \\
&  & ConE & \textbf{$86.8_{\cellcolor{red!25}\underline{-5.8}}$} & \textbf{$62.3_{\textcolor{red}{+0.5}}$} & \textbf{$57.8_{\cellcolor{red!25}\underline{-6.5}}$} & \textbf{$64.7_{\cellcolor{cyan!5}\textcolor{red}{+2.0}}$} & \textbf{$26.5_{\cellcolor{red!15}\underline{-2.2}}$} & \textemdash \\
\midrule
\multirow{15}{*}{\textbf{Gemma3-12B}}
& \multirow{3}{*}{\textbf{ProntoQA}} & BM25 & \textemdash & \textbf{$71.6_{\cellcolor{red!5}\underline{-1.5}}$} & \textbf{$64.8_{\cellcolor{red!25}\underline{-9.0}}$} & \textbf{$64.0_{\cellcolor{red!25}\underline{-9.0}}$} & \textbf{$31.7_{\cellcolor{red!5}\underline{-1.3}}$} & \textbf{$91.3_{\underline{-0.8}}$} \\
&  & Embed & \textemdash & \textbf{$76.5_{\cellcolor{cyan!25}\textcolor{red}{+3.4}}$} & \textbf{$76.2_{\cellcolor{cyan!15}\textcolor{red}{+2.3}}$} & \textbf{$78.7_{\cellcolor{cyan!25}\textcolor{red}{+5.7}}$} & \textbf{$30.9_{\cellcolor{red!15}\underline{-2.2}}$} & \textbf{$91.2_{\underline{-0.8}}$} \\
&  & ConE & \textemdash & \textbf{$71.6_{\cellcolor{red!5}\underline{-1.5}}$} & \textbf{$65.2_{\cellcolor{red!25}\underline{-8.7}}$} & \textbf{$75.7_{\cellcolor{cyan!15}\textcolor{red}{+2.7}}$} & \textbf{$36.1_{\cellcolor{cyan!25}\textcolor{red}{+3.0}}$} & \textbf{$91.6_{\underline{-0.5}}$} \\
\cmidrule(lr){2-9}
& \multirow{3}{*}{\textbf{FOLIO}} & BM25 & \textbf{$99.4_{\cellcolor{cyan!5}\textcolor{red}{+1.2}}$} & \textemdash & \textbf{$77.5_{\cellcolor{cyan!25}\textcolor{red}{+3.7}}$} & \textbf{$66.7_{\cellcolor{red!25}\underline{-6.7}}$} & \textbf{$35.2_{\cellcolor{cyan!25}\textcolor{red}{+5.2}}$} & \textbf{$91.1_{\underline{-0.9}}$} \\
&  & Embed & \textbf{$98.8_{\textcolor{red}{+0.6}}$} & \textemdash & \textbf{$76.5_{\cellcolor{cyan!15}\textcolor{red}{+2.7}}$} & \textbf{$78.0_{\cellcolor{cyan!25}\textcolor{red}{+4.7}}$} & \textbf{$33.5_{\cellcolor{cyan!25}\textcolor{red}{+3.5}}$} & \textbf{$90.8_{\cellcolor{red!5}\underline{-1.2}}$} \\
&  & ConE & \textbf{$98.8_{\textcolor{red}{+0.6}}$} & \textemdash & \textbf{$77.8_{\cellcolor{cyan!25}\textcolor{red}{+4.0}}$} & \textbf{$77.0_{\cellcolor{cyan!25}\textcolor{red}{+3.7}}$} & \textbf{$33.0_{\cellcolor{cyan!25}\textcolor{red}{+3.0}}$} & \textbf{\textemdash} \\
\cmidrule(lr){2-9}
& \multirow{3}{*}{\textbf{ProofWriter}} & BM25 & \textbf{$99.6_{\cellcolor{cyan!5}\textcolor{red}{+1.4}}$} & \textbf{$78.9_{\cellcolor{cyan!25}\textcolor{red}{+6.9}}$} & \textemdash & \textbf{$67.0_{\cellcolor{red!25}\underline{-6.0}}$} & \textbf{$32.6_{\underline{-0.4}}$} & \textbf{$91.0_{\cellcolor{red!5}\underline{-1.1}}$} \\
&  & Embed & \textbf{$99.2_{\cellcolor{cyan!5}\textcolor{red}{+1.0}}$} & \textbf{$77.5_{\cellcolor{cyan!25}\textcolor{red}{+5.4}}$} & \textemdash & \textbf{$78.3_{\cellcolor{cyan!25}\textcolor{red}{+5.3}}$} & \textbf{$32.6_{\underline{-0.4}}$} & \textbf{$91.0_{\cellcolor{red!5}\underline{-1.1}}$} \\
&  & ConE & \textbf{\textemdash} & \textbf{$77.9_{\cellcolor{cyan!25}\textcolor{red}{+5.9}}$} & \textemdash & \textbf{$80.0_{\cellcolor{cyan!25}\textcolor{red}{+7.0}}$} & \textbf{$33.5_{\textcolor{red}{+0.4}}$} & \textbf{\textemdash} \\
\cmidrule(lr){2-9}
& \multirow{3}{*}{\textbf{LogicalDeduction}} & BM25 & \textbf{$94.2_{\cellcolor{red!25}\underline{-4.0}}$} & \textbf{$75.0_{\cellcolor{cyan!25}\textcolor{red}{+3.9}}$} & \textbf{$75.7_{\cellcolor{cyan!5}\textcolor{red}{+1.8}}$} & \textemdash & \textbf{$32.6_{\cellcolor{cyan!15}\textcolor{red}{+2.6}}$} & \textbf{$92.0_{\textcolor{red}{+0.2}}$} \\
&  & Embed & \textbf{$98.0_{\underline{-0.2}}$} & \textbf{$73.0_{\cellcolor{cyan!5}\textcolor{red}{+2.0}}$} & \textbf{$76.2_{\cellcolor{cyan!15}\textcolor{red}{+2.3}}$} & \textemdash & \textbf{$30.9_{\textcolor{red}{+0.9}}$} & \textbf{$91.7_{\underline{-0.1}}$} \\
&  & ConE & \textbf{$99.0_{\textcolor{red}{+0.8}}$} & \textbf{$74.0_{\cellcolor{cyan!15}\textcolor{red}{+2.9}}$} & \textbf{$76.8_{\cellcolor{cyan!25}\textcolor{red}{+3.0}}$} & \textemdash & \textbf{$32.2_{\cellcolor{cyan!15}\textcolor{red}{+2.2}}$} & \textbf{\textemdash} \\
\cmidrule(lr){2-9}
& \multirow{3}{*}{\textbf{GSM8K}} & BM25 & \textbf{$98.8_{\textcolor{red}{+0.6}}$} & \textbf{$72.1$} & \textbf{$75.3_{\cellcolor{cyan!5}\textcolor{red}{+1.5}}$} & \textbf{$77.7_{\cellcolor{cyan!25}\textcolor{red}{+4.7}}$} & \textbf{$33.9_{\textcolor{red}{+0.9}}$} & \textemdash \\
&  & Embed & \textbf{$98.0_{\underline{-0.2}}$} & \textbf{$70.1_{\cellcolor{red!5}\underline{-2.0}}$} & \textbf{$74.3_{\textcolor{red}{+0.5}}$} & \textbf{$75.0_{\cellcolor{cyan!15}\textcolor{red}{+2.0}}$} & \textbf{$32.2_{\underline{-0.9}}$} & \textemdash \\
&  & ConE & \textbf{$99.6_{\cellcolor{cyan!5}\textcolor{red}{+1.4}}$} & \textbf{$70.1_{\cellcolor{red!5}\underline{-2.0}}$} & \textbf{$74.7_{\textcolor{red}{+0.8}}$} & \textbf{$79.3_{\cellcolor{cyan!25}\textcolor{red}{+6.3}}$} & \textbf{$33.5_{\textcolor{red}{+0.4}}$} & \textemdash \\
\midrule
\multirow{15}{*}{\textbf{Gemma3-27B}}
& \multirow{3}{*}{\textbf{ProntoQA}} & BM25 & \textemdash & \textbf{$75.0_{\cellcolor{cyan!25}\textcolor{red}{+3.9}}$} & \textbf{$77.3$} & \textbf{$84.0_{\cellcolor{cyan!25}\textcolor{red}{+6.3}}$} & \textbf{$38.7_{\cellcolor{red!5}\underline{-1.3}}$} & \textbf{$93.1_{\underline{-0.1}}$} \\
&  & Embed & \textemdash & \textbf{$77.0_{\cellcolor{cyan!25}\textcolor{red}{+5.9}}$} & \textbf{$82.5_{\cellcolor{cyan!25}\textcolor{red}{+5.2}}$} & \textbf{$88.0_{\cellcolor{cyan!25}\textcolor{red}{+10.3}}$} & \textbf{$39.1_{\underline{-0.9}}$} & \textbf{$93.3_{\textcolor{red}{+0.1}}$} \\
&  & ConE & \textemdash & \textbf{$73.5_{\cellcolor{cyan!15}\textcolor{red}{+2.5}}$} & \textbf{$75.7_{\cellcolor{red!5}\underline{-1.7}}$} & \textbf{$87.0_{\cellcolor{cyan!25}\textcolor{red}{+9.3}}$} & \textbf{$37.8_{\cellcolor{red!15}\underline{-2.2}}$} & \textbf{$93.4_{\textcolor{red}{+0.2}}$} \\
\cmidrule(lr){2-9}
& \multirow{3}{*}{\textbf{FOLIO}} & BM25 & \textbf{$99.2_{\textcolor{red}{+0.8}}$} & \textemdash & \textbf{$81.3_{\cellcolor{cyan!25}\textcolor{red}{+4.0}}$} & \textbf{$86.7_{\cellcolor{cyan!25}\textcolor{red}{+9.0}}$} & \textbf{$39.1_{\cellcolor{cyan!25}\textcolor{red}{+4.3}}$} & \textbf{$93.7_{\textcolor{red}{+0.5}}$} \\
&  & Embed & \textbf{$99.0_{\textcolor{red}{+0.6}}$} & \textemdash & \textbf{$83.2_{\cellcolor{cyan!25}\textcolor{red}{+5.8}}$} & \textbf{$86.0_{\cellcolor{cyan!25}\textcolor{red}{+8.3}}$} & \textbf{$37.0_{\cellcolor{cyan!15}\textcolor{red}{+2.2}}$} & \textbf{$93.4_{\textcolor{red}{+0.2}}$} \\
&  & ConE & \textbf{$99.6_{\cellcolor{cyan!5}\textcolor{red}{+1.2}}$} & \textemdash & \textbf{$82.0_{\cellcolor{cyan!25}\textcolor{red}{+4.7}}$} & \textbf{$89.0_{\cellcolor{cyan!25}\textcolor{red}{+11.3}}$} & \textbf{$40.9_{\cellcolor{cyan!25}\textcolor{red}{+6.1}}$} & \textbf{\textemdash} \\
\cmidrule(lr){2-9}
& \multirow{3}{*}{\textbf{ProofWriter}} & BM25 & \textbf{$100.0_{\cellcolor{cyan!5}\textcolor{red}{+1.8}}$} & \textbf{$76.5_{\cellcolor{cyan!25}\textcolor{red}{+5.4}}$} & \textemdash & \textbf{$87.0_{\cellcolor{cyan!25}\textcolor{red}{+9.3}}$} & \textbf{$40.4_{\textcolor{red}{+0.9}}$} & \textbf{$93.6_{\textcolor{red}{+0.4}}$} \\
&  & Embed & \textbf{$99.6_{\cellcolor{cyan!5}\textcolor{red}{+1.4}}$} & \textbf{$78.4_{\cellcolor{cyan!25}\textcolor{red}{+7.4}}$} & \textemdash & \textbf{$90.0_{\cellcolor{cyan!25}\textcolor{red}{+12.3}}$} & \textbf{$40.0_{\textcolor{red}{+0.4}}$} & \textbf{$93.6_{\textcolor{red}{+0.4}}$} \\
&  & ConE & \textbf{\textemdash} & \textbf{$78.9_{\cellcolor{cyan!25}\textcolor{red}{+7.8}}$} & \textemdash & \textbf{$87.7_{\cellcolor{cyan!25}\textcolor{red}{+10.0}}$} & \textbf{$40.0_{\textcolor{red}{+0.4}}$} & \textbf{\textemdash} \\
\cmidrule(lr){2-9}
& \multirow{3}{*}{\textbf{LogicalDeduction}} & BM25 & \textbf{$99.2_{\textcolor{red}{+0.8}}$} & \textbf{$74.5_{\cellcolor{cyan!25}\textcolor{red}{+3.4}}$} & \textbf{$78.5_{\cellcolor{cyan!15}\textcolor{red}{+2.5}}$} & \textemdash & \textbf{$41.7_{\cellcolor{cyan!5}\textcolor{red}{+1.7}}$} & \textbf{$93.4_{\textcolor{red}{+0.2}}$} \\
&  & Embed & \textbf{$99.2_{\textcolor{red}{+0.8}}$} & \textbf{$73.5_{\cellcolor{cyan!15}\textcolor{red}{+2.5}}$} & \textbf{$78.8_{\cellcolor{cyan!15}\textcolor{red}{+2.8}}$} & \textemdash & \textbf{$43.9_{\cellcolor{cyan!25}\textcolor{red}{+3.9}}$} & \textbf{$93.2_{\underline{-0.1}}$} \\
&  & ConE & \textbf{$99.2_{\textcolor{red}{+0.8}}$} & \textbf{$76.0_{\cellcolor{cyan!25}\textcolor{red}{+4.9}}$} & \textbf{$79.2_{\cellcolor{cyan!25}\textcolor{red}{+3.2}}$} & \textemdash & \textbf{$40.0$} & \textbf{\textemdash} \\
\cmidrule(lr){2-9}
& \multirow{3}{*}{\textbf{GSM8K}} & BM25 & \textbf{$98.2$} & \textbf{$72.5_{\cellcolor{cyan!5}\textcolor{red}{+1.5}}$} & \textbf{$80.0_{\cellcolor{cyan!15}\textcolor{red}{+2.7}}$} & \textbf{$85.7_{\cellcolor{cyan!25}\textcolor{red}{+7.0}}$} & \textbf{$38.7_{\underline{-0.9}}$} & \textemdash \\
&  & Embed & \textbf{$97.4_{\underline{-0.8}}$} & \textbf{$73.5_{\cellcolor{cyan!15}\textcolor{red}{+2.5}}$} & \textbf{$76.8_{\underline{-0.5}}$} & \textbf{$87.0_{\cellcolor{cyan!25}\textcolor{red}{+8.3}}$} & \textbf{$38.7_{\underline{-0.9}}$} & \textemdash \\
&  & ConE & \textbf{$98.2$} & \textbf{$71.6_{\textcolor{red}{+0.5}}$} & \textbf{$74.2_{\cellcolor{red!25}\underline{-3.2}}$} & \textbf{$90.3_{\cellcolor{cyan!25}\textcolor{red}{+11.7}}$} & \textbf{$38.3_{\cellcolor{red!5}\underline{-1.3}}$} & \textemdash \\
\midrule
\bottomrule
\end{tabular}
\caption{Relative performance of different retrieval strategies across source–target domain pairs, measured against the 0-shot baseline. Deeper \colorbox{cyan!25}{blue} indicates larger gains, while deeper \colorbox{red!25}{red} denotes greater degradation.}
\label{tab:transfer_selected_models}
\end{table*}

%% file: latex/tex/setup.tex
\section{Experimental Setup}
% \lzm{Setup}

\subsection{Models and Datasets}
For models, we evaluate a diverse set of backbone models spanning multiple
families and parameter scales, including 
\textbf{Qwen2.5} (3B, 7B, 14B, 32B) \cite{qwen2.5},
\textbf{Qwen3} (4B, 8B, 14B, 32B) \cite{qwen3technicalreport},
\textbf{Gemma~3} (4B, 12B, 27B) \cite{gemma_2025},
and \textbf{Llama~3.1} (8B) \cite{grattafiori2024llama3herdmodels}.

For datasets, we use six reasoning benchmarks:
\textbf{GSM8K} (grade-school math) \cite{cobbe2021gsm8k},
\textbf{ProntoQA} (synthetic multi-hop deduction) \cite{saparov2022language},
\textbf{LogicalDeduction} (symbolic consistency reasoning) \cite{nguyen2025symbolic},
\textbf{FOLIO} (first-order logical inference) \cite{han2022folio},
\textbf{ProofWriter} (deductive proofs) \cite{tafjord-etal-2021-proofwriter}, 
and \textbf{AR-LSAT} (law school analytical reasoning) \cite{zhong2021arlsat}.
These datasets span arithmetic, symbolic logic, rule-based deduction, and 
formal reasoning, providing a broad testbed for evaluating ICL transferability.

\subsection{Baselines}
We evaluate three representative retrieval metric functions for constructing in-context demonstrations: (1) \textbf{Embedding-based retrieval} \cite{lewis2021retrievalaugmentedgenerationknowledgeintensivenlp}, which selects top-k examples via cosine similarity in the model’s embedding space (We choose bge-large-en-v1.5 as the embedding model \cite{bge_embedding}); (2) \textbf{BM25 retrieval} \cite{robertson2009probabilistic}, a lexical matching baseline widely used in IR and prior ICL studies; (3) \textbf{TopK+ConE} \cite{peng-etal-2024-revisiting} which refines retrieved candidates through conditional-entropy–based reranking to identify the most informative demonstrations.

%% file: latex/tex/result.tex
\section{Main Results}

% \input{latex/tex/table1}

% \subsection{Case Study: Structural Repair via Cross-Domain Demonstrations}

\begin{figure*}[t!]
  \includegraphics[scale=0.31]{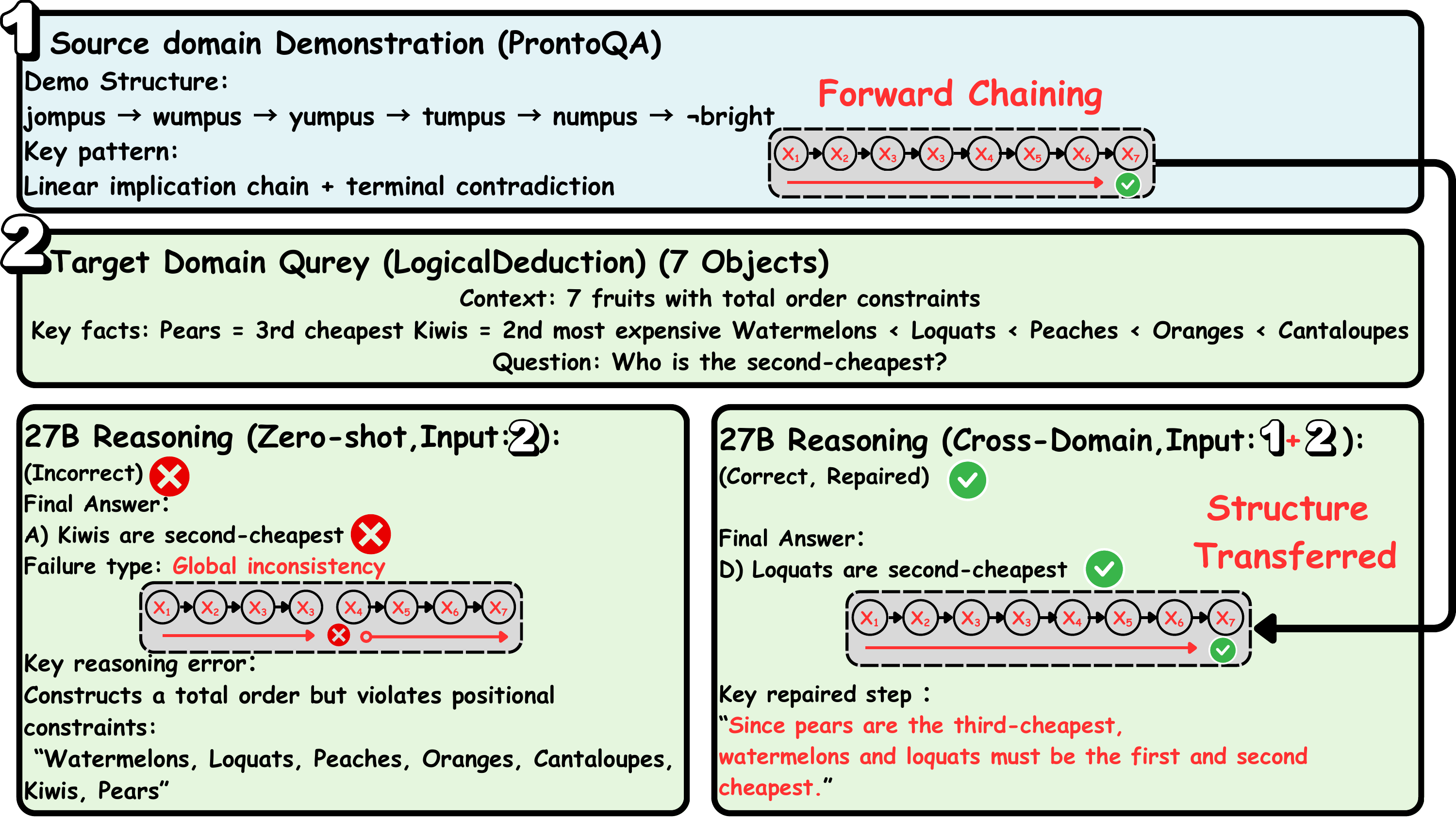}
  \caption{\textbf{Comparison between zero-shot and cross-domain ICL.} Bottom left: Zero-shot reasoning omits a required intermediate link, leading to an incorrect prediction. Bottom right: Cross-domain ICL restores the missing link via a structurally compatible demonstration, yielding the correct answer.}
  \label{case1}
\end{figure*}

\begin{figure*}[t!]
    \centering
    \includegraphics[width=\linewidth]{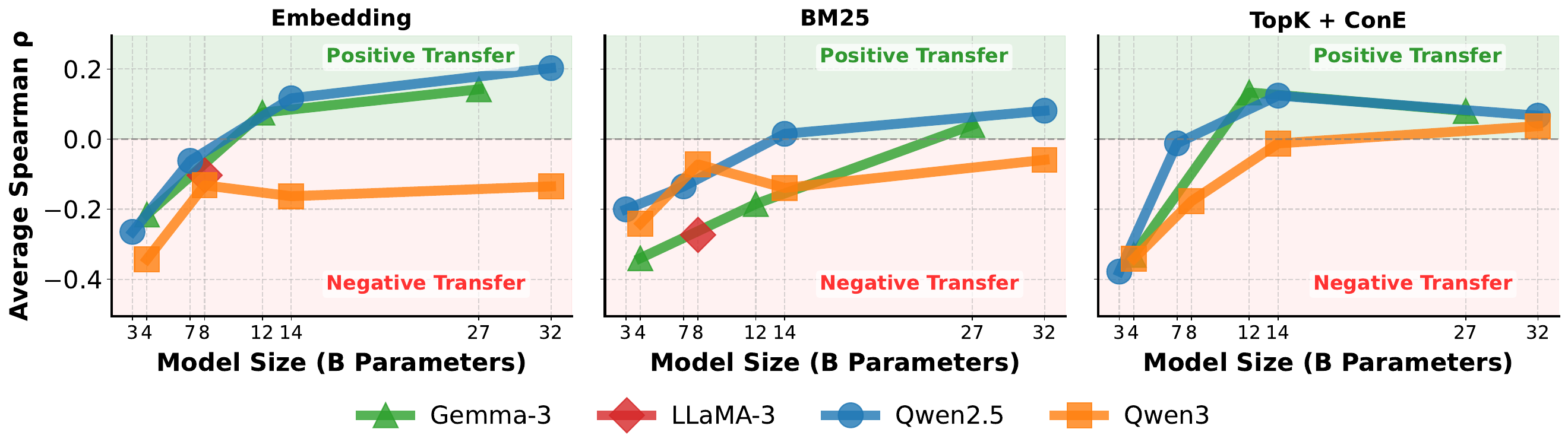}
    \caption{\textbf{Scaling behaviour of cross-domain transfer under different retrieval baselines}. Average Spearman $\rho$ between model size and few-shot performance is plotted for Embedding and BM25 across model families. Shaded regions denote positive and negative transfer. }
    \label{fig:scaling}
\end{figure*}

% Figure \ref{case1} illustrates that cross-domain in-context learning operates at the structural level rather than via surface imitation. Although the ProntoQA source example and the LogicalDeduction target task differ semantically, both rely on sequential constraint propagation.

% In the zero-shot setting, even Gemma-3-27B fails to enforce global ordering constraints, producing locally plausible but globally inconsistent reasoning. With a cross-domain demonstration, the model transfers the forward-chaining structure from the source domain, explicitly repairs inconsistencies, and restores a valid reasoning topology, leading to the correct conclusion.

% This example shows that cross-domain ICL can repair zero-shot reasoning failures by reinstating structurally compatible topologies, beyond surface-level pattern matching

\subsection{RQ1: How is the overall performance of cross-domain knowledge transfer via ICL?}

In order to investigate the overall performance of cross-domain knowledge transfer via ICL, we conducted large-scale experiments using three representative baselines across 25 different transfer directions. Table~\ref{tab:transfer_selected_models} reveals a clear scaling trend: as model size increases, cross-domain knowledge transfer becomes consistently more beneficial, with negative transfer substantially reduced. In particular, Gemma3-27B exhibits negative transfer in only five source--target pairs, whereas smaller variants (4B and 12B) suffer from considerably more failures.

The gains are especially pronounced on the Logical Deduction benchmark. For Gemma3-27B, embedding-based retrieval improves accuracy by $+12.3\%$ for ProofWriter$\rightarrow$Logical Deduction and by $+10.3\%$ for ProntoQA$\rightarrow$Logical Deduction. 

We attribute this effect to model scaling: smaller models often fail to abstract task-solving logic from cross-domain demonstrations and are distracted by surface-level mismatches, leading to negative transfer. In contrast, larger models can better identify and exploit shared reasoning structures across domains. Figure~\ref{case1} illustrates a representative Logical Deduction example where Gemma3-27B recovers the correct reasoning chain after conditioning on an embedding-retrieved cross-domain demonstration.

% \lzm{The point of rq is to derive ``universal research conclusions'' not just stating facts. We could conclude something like ``we can observe from Table 1 that from top to bottom, the reddish areas gradually decrease and the bluish area gradully increase, indicating that as the size of the model increases, the more beneficial cross-domain knowledge transfer is. Concretely, for Gemma3-27B, there are only 5 subtasks where negative transfer happens, whereas for the 4B and 12B variants, there are xxx and xxx, respectively. Besides, we observe that the improvements are particularly noticeable for the Logical Deduction benchmark (target domain) (\eg~for Gemma3-27B, ProofWriter$\rightarrow$Logical Deduction), Embed manages to improve the baseline by $+12.3\%$, and $+10.3\%$ for ProntoQA$\rightarrow$Logical Deduction). We think the scaling effect of model size is due to the fact that the small-scale models are unable to uncover the essential task-solving rationale (\ie~task-solving logic) from the cross-domain demonstrations, instead they are puzzled by the surface-level differences, which elicit the negative transfer. Whereas larger-scale models possess much stronger reasoning ability, allowing it to well use the hidden task-solving logic. Figure 2 shows a concrete Logical Deduction question where the Gemma3-27B model initially done wrong, yet manages to succefully solve by learning from the demonstration retrieved via the Embed metric function. Specifically, ...}

\researchfinding{\textbf{I: }Cross-domain ICL shows systematic and often substantial positive transfer in specific source–target directions.}

\subsection{RQ2: How does model scale influence cross-domain transferability?}
% \lzm{keep it the same as intro}
Figure \ref{fig:scaling} reveals a clear scaling-dependent pattern in cross-domain ICL. Small models (3B–7B) exhibit highly unstable behaviour, where negative transfer is common, and the average Spearman correlation between shot size and performance is near zero or even negative. This indicates that additional demonstrations are often not “absorbed’’ effectively, and may even introduce structural mismatches that degrade reasoning.

\begin{figure*}[t!]
    \centering
    \includegraphics[width=\linewidth]{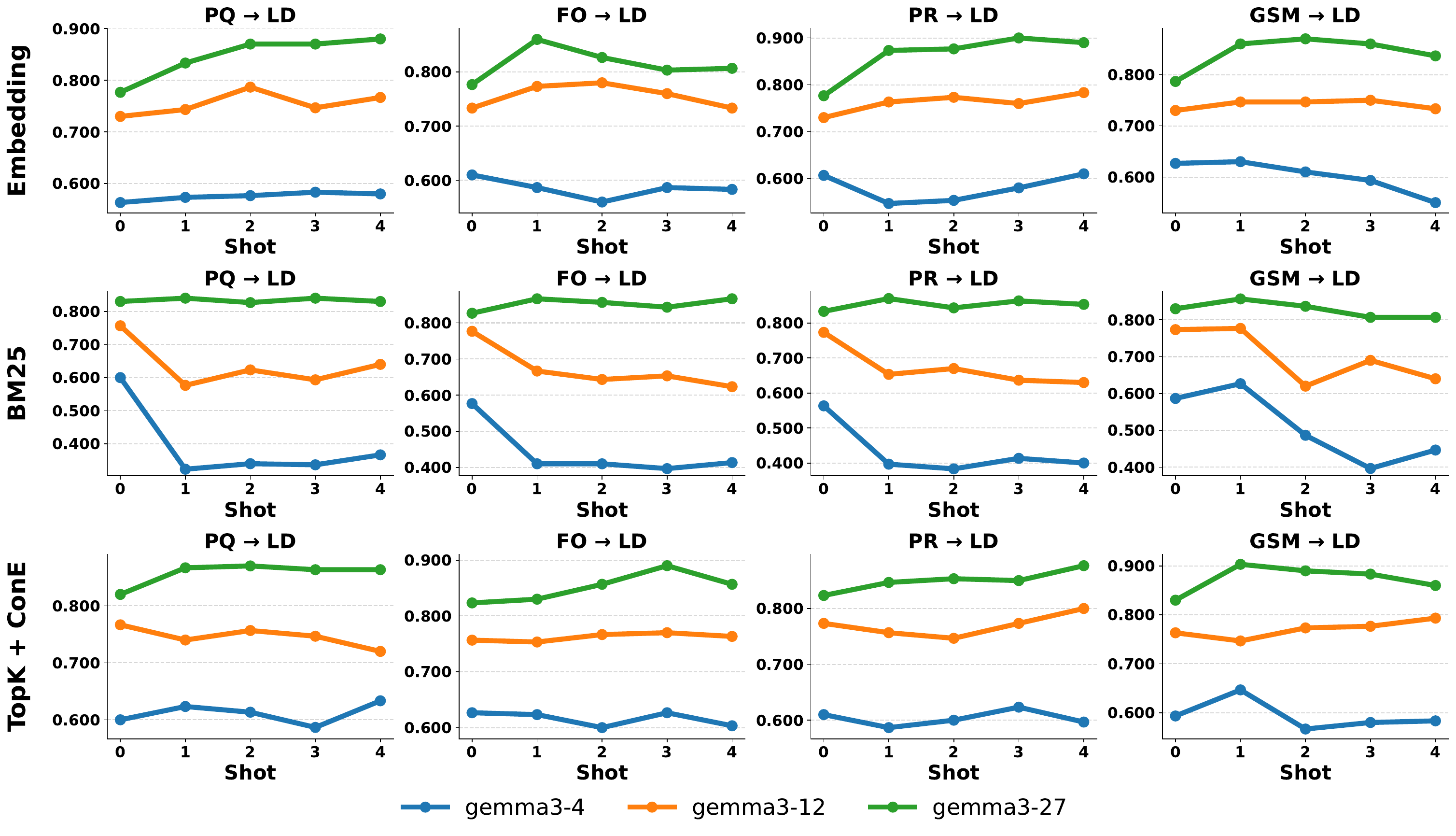}
    \caption{\textbf{Shot–performance scaling across cross-domain ICL settings.}}
    \label{fig:shot}
\end{figure*}

As model size increases, the trend shifts: mid- to large-scale models (12B–32B) show consistently positive correlations and markedly fewer negative-transfer cases. These models not only benefit more reliably from additional examples but also maintain stable improvements across a broader range of source–target directions.

Taken together, the results suggest the presence of an \emph{example absorption threshold}: only models above a certain capacity can reliably leverage cross-domain demonstrations, while smaller models remain highly sensitive and prone to degradation. This highlights model scale as a key factor governing whether ICL exhibits positive or negative shot scaling.

\researchfinding{\textbf{II: }Cross-domain ICL exhibits an absorption threshold: sufficiently large models consistently benefit from cross-domain demonstrations, while smaller models are prone to negative transfer.}

\subsection{RQ3: How does the number of shots affect cross-domain transferability?}

Figure \ref{fig:shot} further examines how model capacity mediates the effect of demonstration quantity in cross-domain ICL. We observe a clear divergence between large and small models in their ability to benefit from additional in-context examples. For the 27B model, performance consistently improves with more demonstrations across all positive transfer pairs, indicating strong capacity for cross-domain integration. In contrast, smaller models (4B and 12B) show unstable scaling behavior, where additional demonstrations often lead to diminishing or even negative returns, especially on structurally mismatched targets such as GSM8K and AR-LSAT. These results suggest that while larger models can effectively exploit richer cross-domain supervision, smaller models lack the representational capacity to accommodate numerous heterogeneous examples, leading to interference rather than benefit. 

Overall, the findings indicate that the effectiveness of increasing shot size is strongly dependent on model scale. While larger models can better leverage additional demonstrations, smaller models often exhibit diminishing or negative returns, indicating that simply increasing the number of demonstrations does not reliably improve performance.

\researchfinding{\textbf{III: }Once the source–target transfer exceeds an absorption threshold, increasing the shot size within a moderate range yields consistently larger cross-domain ICL gains.}

\subsection{RQ4: What's the source of improvement?}
% \lzm{To better understand the source of improvement, we further conduct an in-depth analysis based on the categorization of task-solving logic. Concretely, we first collect the samples that the Gemma3-27b model originally did wrong, and deduce correctly with the introduction of cross-domain demonstrations. We then categorize this set of samples into four categories according to the structures of their task-solving logic, namely L-type, Y-type, ...}

\begin{figure}[t!]
  \includegraphics[scale=0.15]{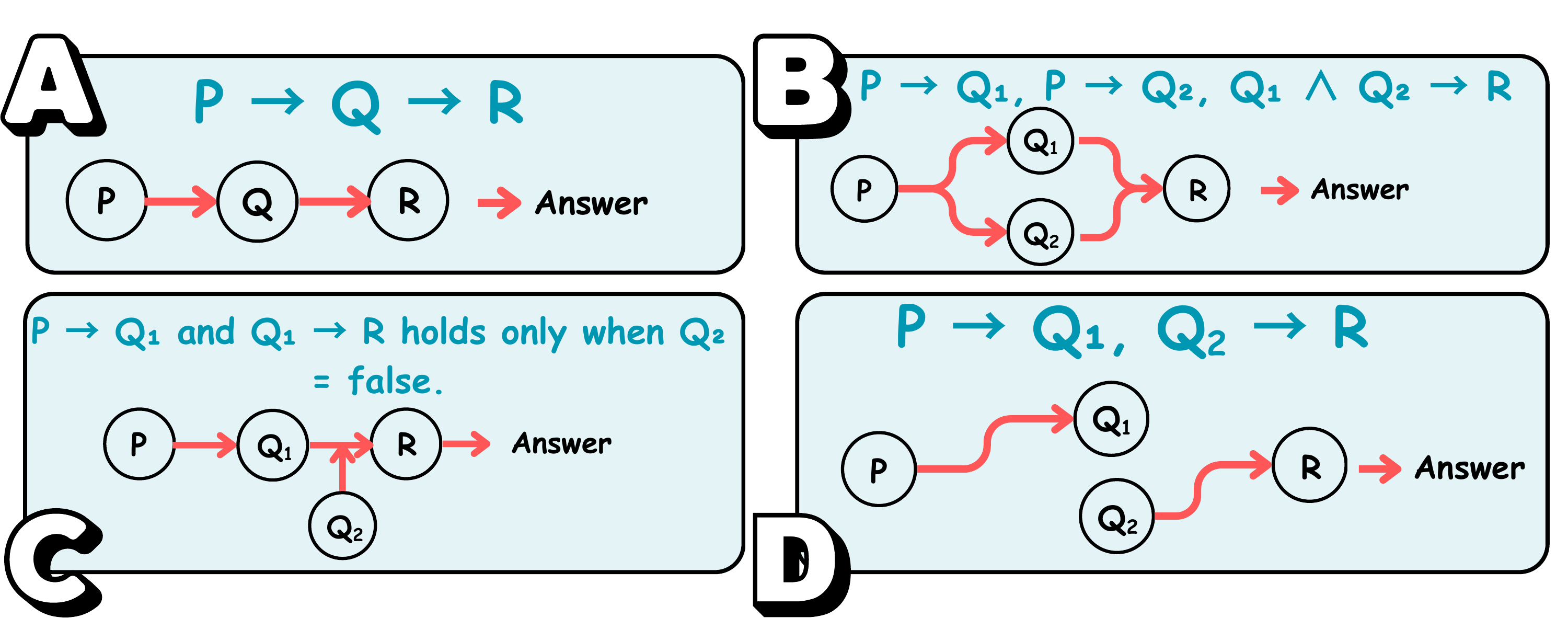}
  \caption{\textbf{Four types of forward chaining.}}
  \label{chaining_type}
\end{figure}

To better understand the source of improvement, we further conduct an in-depth analysis based on the categorization of task-solving logic. Concretely, we first collect the samples that the Gemma3-27b model originally did wrong, and deduce correctly with the introduction of cross-domain demonstrations. As shown in Figure~\ref{chaining_type}, we categorize this set of samples into four categories according to the structures of their task-solving logic, namely L-type (A), Y-type (B), Block-type (C), and other types (D). We then used Deepseek-V3~\cite{deepseekai2024deepseekv3technicalreport} to classify the inference structure of the target query and the retrieved source domain examples. Finally, we plotted the logic structure distribution of all retrieval examples that repaired 0-shot samples when LogicalDeduction was the target domain.

\begin{figure*}[h!]
    \centering
    \includegraphics[width=\linewidth]{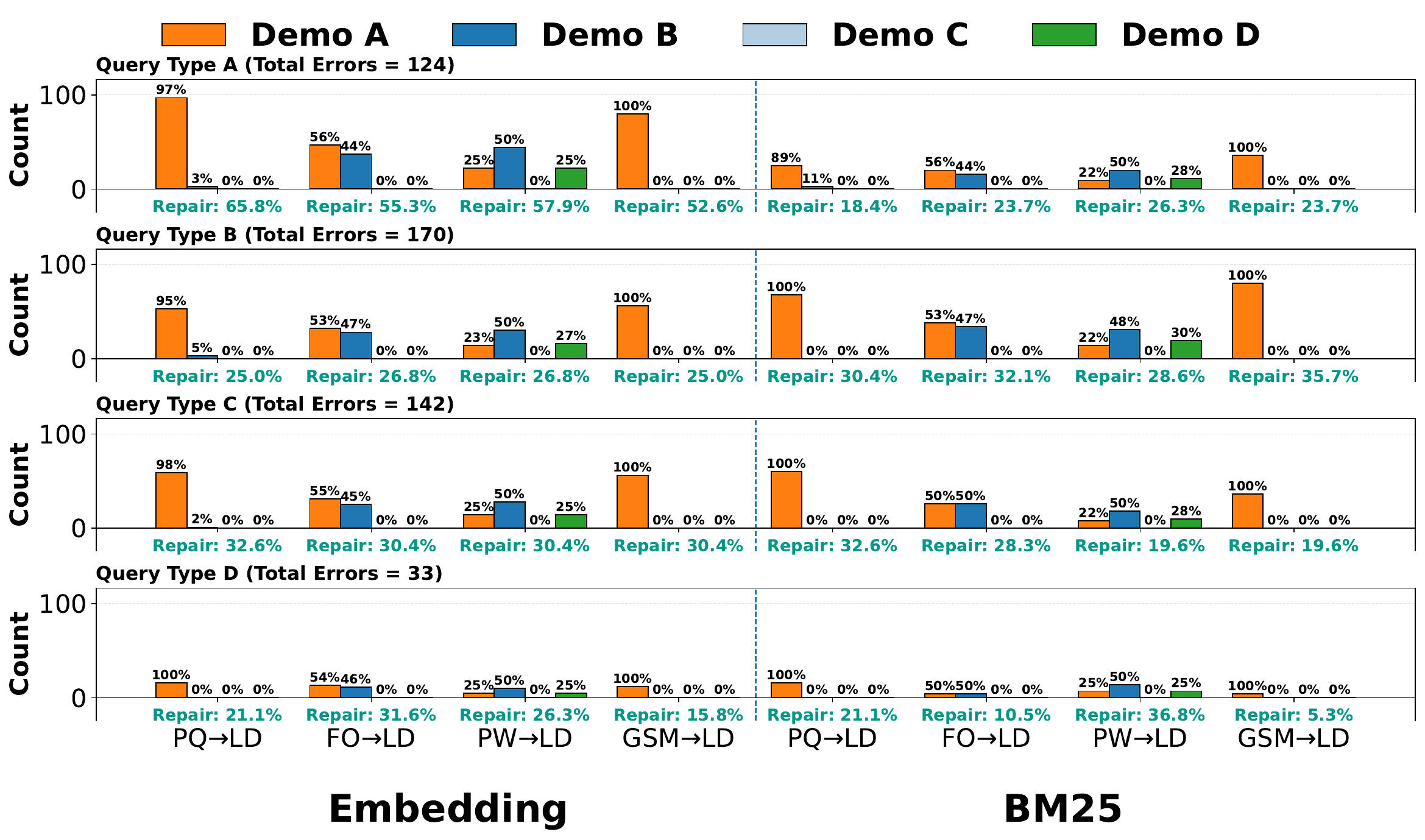}
    \caption{\textbf{Distribution of demonstration topologies among repaired zero-shot errors across retrieval methods and transfer directions.}}
    \label{fig:repair_dist}
\end{figure*}

As shown in Figure~\ref{fig:repair_dist}, the repair ratio varies considerably across retrieval methods and transfer directions. For instance, under embedding-based retrieval, Query Type~A achieves repair ratios of 52.6$\%$–65.8$\%$, whereas the same query type under BM25 drops sharply to 18.4$\%$–26.3$\%$.
Similar sensitivity is observed for Query Type~D, where repair ratios remain below 36.8$\%$ and fall to as low as 5.3$\%$ under mismatched retrieval–demo configurations. These results indicate that cross-domain ICL gains are highly sensitive to the retrieval configuration and depend critically on whether the retrieved demonstrations align with the underlying reasoning structure of the query.

% Figure~\ref{fig:repair_dist} analyzes \emph{where} the observed gains come from by decomposing improvements into (i) the \textbf{query-level repair ratio}---the fraction of unique zero-shot errors corrected by few-shot prompting under each retrieval configuration, and (ii) the \textbf{demo-level topology composition}---the distribution of retrieved demonstration topologies among repaired queries.

At the demonstration level, the retrieved demonstrations for repaired queries exhibit a clear topology bias. Both embedding-based retrieval and BM25 tend to return demonstrations dominated by a small number of chain types (e.g., type A), while other types appear rarely. Such biased retrieval suggests that current retrievers do not fully explore structurally diverse yet potentially transferable demonstrations, which may limit the achievable cross-domain transfer gains. This points to a promising future direction: incorporating chain-structure similarity into retrieval to better align demonstrations with target reasoning structures and improve cross-domain performance.

\researchfinding{\textbf{IV: }Cross-domain ICL gains arise from retrieval-induced repair of a subset of zero-shot failures via structurally compatible demonstrations.}

%% file: latex/tex/related.tex
\section{Related Work}

\subsection{Generalization in LLM}
Large language models (LLMs) often degrade under domain shifts \cite{oncel2024adaptation, oh2025understanding}. Existing approaches—such as data-centric adaptation \cite{wang2024neurons}, prompt calibration \cite{zhao2021calibrate, honda-oka-2025-exploring, he-etal-2024-using}, and parameter-efficient tuning \cite{hu2022lora}. Recent work has begun examining cross-domain representation alignment \cite{aghajanyan2020intrinsic}, including neuron-level alignment in multilingual settings \cite{huang2025neurons}. Neuron-level analyses are well studied \cite{chen2024analyzing, sajjad2022neuron}, but their role in cross-domain in-context learning remains unexplored.

\subsection{Example Selection for In-Context Learning}
In-context learning (ICL) is highly sensitive to demonstration selection \cite{luo2024context}. Existing retrieval methods rely on semantic similarity \cite{rubin2021learning}, dense retrievers \cite{wang2023learning}, uncertainty signals \cite{ling2024uncertainty, huang2024unlocking, margatina2023active}, coverage-based selection \cite{gupta-etal-2023-coverage}, or MMR-based diversification \cite{liu2023lost}. Recent work uses internal representations to analyze LLM reasoning \cite{liu2023lost}. Overall, prior work relies on heuristic retrieval and overlooks reasoning-level alignment.

%% file: latex/tex/conclusion.tex
\section{Conclusion}
We conducted a systematic empirical study of cross-domain in-context learning, showing that transfer performance is sensitive to retrieval quality and demonstration selection, and that only models above an example absorption threshold can reliably benefit from cross-domain demonstrations, while increasing the number of demonstrations does not consistently improve results. Our findings suggest that effective cross-domain transfer mainly stems from structurally compatible demonstrations that repair zero-shot failures, highlighting the need for structure-aware retrieval in future work.

%% file: latex/tex/limitations.tex
\section*{Limitations}
Our study is limited to a specific set of reasoning tasks and retrieval methods, and the findings may not fully generalize to other domains or retrieval paradigms. The notion of reasoning repair is characterized heuristically rather than through a formally defined structural metric. In addition, our analysis focuses on a fixed decoding setup, leaving the interaction between retrieval effects and different decoding strategies underexplored. Finally, while we observe complementary behaviours across retrieval methods, we do not explicitly model or optimize this complementarity.

%% file: latex/tex/appendix.tex
\appendix

\section{Appendix}
\label{sec:appendix}

\subsection{Prompt Construction for In-Context Learning}
\label{app:prompt}

This section describes the prompt templates used for all in-context learning (ICL) experiments. Unless otherwise specified, the same prompt structure is applied across all models, tasks, and retrieval methods to ensure fair comparison.

\subsubsection{Zero-Shot Prompt}

For zero-shot inference, the prompt consists solely of a task instruction followed by the query instance:
\begin{equation}
\texttt{Prompt}_{0\text{-shot}} =
\big[
\texttt{Instruction} \ ;
\ x_q
\big],
\end{equation}
where $x_q$ denotes the target query. The instruction provides a brief task description and remains fixed within each dataset.

\subsubsection{$k$-Shot Prompt}

For $k$-shot ICL, the prompt is constructed by concatenating $k$ retrieved demonstration examples with the query:
\begin{equation}
\small
\texttt{Prompt}_{k\text{-shot}} =
\big[
\texttt{Instruction} \ ;
\ d_{i_1}, d_{i_2}, \dots, d_{i_k} \ ;
\ x_q
\big],
\end{equation}
where $\{d_{i_j}\}_{j=1}^{k}$ are the top-$k$ demonstrations returned by the retrieval module (Appendix~\ref{app:retrieval}). Each demonstration $d_i$ follows a fixed format:
\begin{equation}
d_i = \big[
x_i \ ;\ y_i
\big],
\end{equation}
with $x_i$ denoting the input (e.g., context and question) and $y_i$ the corresponding gold answer.

\subsubsection{Demonstration Ordering}

Retrieved demonstrations are ordered by decreasing retrieval score, such that $d_{i_1}$ is the most similar example to the query. To study order sensitivity, we also evaluate a reversed order setting, where demonstrations are concatenated in ascending order of retrieval score. Apart from the ordering, all other prompt components are kept identical.

\subsubsection{Reasoning and Answer Format}

For tasks that involve explicit reasoning, demonstrations may include intermediate reasoning steps (e.g., chain-of-thought) followed by a final answer. The query instance, however, only requires the model to output the final answer. This design prevents information leakage while preserving the structural guidance provided by the demonstrations.

\subsubsection{Prompt Consistency}

All retrieval methods, shot numbers, and experimental conditions share the same prompt template and textual fields. The total prompt length is constrained by the model context window, and no additional prompt engineering or task-specific tuning is applied beyond the fixed templates described above.

\subsection{Prompt Templates}
\label{app:task_prompts}

We adopt task-specific \emph{system prompts} for each dataset to ensure (i) a consistent reasoning style across domains and (ii) a unified, strictly formatted final answer for reliable automatic evaluation. All prompts follow a two-stage structure: the model is first instructed to provide a short step-by-step reasoning, and then output the final answer on a separate line using a standardized format (e.g., \texttt{Final answer: A}). This design minimizes ambiguity in option extraction and avoids dataset-specific parsing heuristics.

\paragraph{ProntoQA.}
ProntoQA is a binary logical reasoning task with two answer options. We use the following system prompt:
\begin{tcolorbox}[
    colback=gray!4,
    colframe=gray!40,
    boxrule=0.3pt,
    arc=2pt,
    left=4pt,right=4pt,top=4pt,bottom=4pt]
\small\texttt{%
You are a careful reasoner. Think step by step concisely.\\
Then on a new line, output exactly: `Final answer: A' or `Final answer: B'.
}
\end{tcolorbox}

\paragraph{FOLIO.}
FOLIO requires three-way classification (\emph{entailment} / \emph{contradiction} / \emph{unknown}). We extend the same reasoning template to support three options:
\begin{tcolorbox}[
    colback=gray!4,
    colframe=gray!40,
    boxrule=0.3pt,
    arc=2pt,
    left=4pt,right=4pt,top=4pt,bottom=4pt]
\small\texttt{%
You are a careful reasoner. Think step by step concisely.\\
Then on a new line, output exactly:\\
`Final answer: A' or `Final answer: B' or `Final answer: C'.
}
\end{tcolorbox}

\paragraph{ProofWriter.}
ProofWriter is a logical reasoning task typically formulated as entailment-style verification. We adopt the same binary template as ProntoQA:
\begin{tcolorbox}[
    colback=gray!4,
    colframe=gray!40,
    boxrule=0.3pt,
    arc=2pt,
    left=4pt,right=4pt,top=4pt,bottom=4pt]
\small\texttt{%
You are a careful reasoner. Think step by step concisely.\\
Then on a new line, output exactly: `Final answer: A' or `Final answer: B'.
}
\end{tcolorbox}

\paragraph{LogicalDeduction.}
LogicalDeduction is a multiple-choice deductive reasoning task. Let $\mathcal{O}=\{A,B,C,D\}$ denote the option set (dataset-dependent). We instruct the model to output the selected option letter in a strict format:
\begin{tcolorbox}[
    colback=gray!4,
    colframe=gray!40,
    boxrule=0.3pt,
    arc=2pt,
    left=4pt,right=4pt,top=4pt,bottom=4pt]
\small\texttt{%
You are a careful reasoner. Think step by step concisely.\\
Then on a new line, output exactly: `Final answer: <option letter>'.
}
\end{tcolorbox}

\paragraph{AR-LSAT.}
AR-LSAT involves multi-step analytical reasoning with multiple-choice answers. We use the same standardized option-letter output format to enable robust evaluation:
\begin{tcolorbox}[
    colback=gray!4,
    colframe=gray!40,
    boxrule=0.3pt,
    arc=2pt,
    left=4pt,right=4pt,top=4pt,bottom=4pt]
\small\texttt{%
You are a careful analytical reasoner. Think step by step concisely.\\
Then on a new line, output exactly: `Final answer: <option letter>'.
}
\end{tcolorbox}

\paragraph{GSM8K.}
For math word problems, we follow the conventional Chain-of-Thought prompting format and enforce a strict numeric final answer:
\begin{tcolorbox}[
    colback=gray!4,
    colframe=gray!40,
    boxrule=0.3pt,
    arc=2pt,
    left=4pt,right=4pt,top=4pt,bottom=4pt]
\small\texttt{%
You are a careful math reasoner. Solve step by step concisely.\\
Then on a new line, output exactly: `Final answer: <number>'.
}
\end{tcolorbox}

\noindent\textbf{Remark.}
Across all datasets, we keep the reasoning instruction intentionally brief and enforce a single-line final answer in a fixed pattern. During evaluation, we extract the prediction by matching the last occurrence of the \texttt{Final answer:} prefix and parsing the subsequent token(s) according to the task type (option letter or number).

\subsection{Implementation Details}
\subsubsection{Model Inference}
All experiments are conducted using vLLM \cite{kwon2023efficient} as the inference backend to ensure efficient serving of large models and fast hidden-state extraction. Unless otherwise specified, model precision is set to FP16, following the default mixed-precision configuration of vLLM. We use HuggingFace Transformers \cite{wolf-etal-2020-transformers} for model loading, tokenization, and hidden-state access.

\subsubsection{Generation Hyperparameters}
As shown in Table~\ref{hyperparam}, across all experiments—including cross-domain ICL evaluation, DIN retrieval, and case studies—we use the following decoding configuration:
\begin{table}[h]
\centering

\begin{tabular}{ll}
\toprule
\textbf{Category} & \textbf{Setting} \\
\midrule
Temperature           & \textbf{0.0(Greedy)} \\
Max Gen Length        & 8192 tokens \\
Random Seed           & 1-30 \\
\bottomrule
\end{tabular}
\caption{Decoding setup used throughout all experiments.}
\label{hyperparam}
\end{table}

\subsection{Retrieval Methods}
\label{app:retrieval}

This section details the retrieval methods used to select in-context demonstrations for cross-domain ICL, including dense vector retrieval and sparse lexical retrieval (BM25). All retrieval methods operate on the source-domain demonstration pool $\mathcal{D}_S = \{x_i\}_{i=1}^{N}$ and return the top-$k$ examples for a given target query $x_q$.

\subsubsection{Dense Vector Retrieval}
\label{app:dense_retrieval}

In dense retrieval, each example $x$ is mapped to a continuous embedding vector via an encoder $f_\theta(\cdot)$:
\begin{equation}
\mathbf{h}_x = f_\theta(x) \in \mathbb{R}^d,
\end{equation}
where $f_\theta$ is a pretrained sentence or text encoder. The target query $x_q$ is embedded in the same space as $\mathbf{h}_{x_q}$.

Retrieval is performed by computing a similarity score between the query embedding and each candidate example. We adopt cosine similarity:
\begin{equation}
\mathrm{sim}_{\text{dense}}(x_q, x_i) =
\frac{\mathbf{h}_{x_q}^\top \mathbf{h}_{x_i}}
{\|\mathbf{h}_{x_q}\| \, \|\mathbf{h}_{x_i}\|}.
\end{equation}

The top-$k$ demonstrations are selected by:
\begin{equation}
\mathcal{R}_k^{\text{dense}}(x_q)
=
\operatorname*{arg\,topk}_{x_i \in \mathcal{D}_S}
\ \mathrm{sim}_{\text{dense}}(x_q, x_i).
\end{equation}

Dense retrieval captures semantic similarity in a continuous representation space, enabling soft matching beyond exact lexical overlap.

\subsubsection{BM25 Retrieval}
\label{app:bm25}

BM25 is a classical sparse retrieval method based on term-level exact matching. Each example $x$ is treated as a bag of tokens. Given a query $x_q$, the BM25 score between $x_q$ and a candidate example $x_i$ is computed as:
\begin{equation}
\begin{aligned}
\mathrm{sim}_{\text{BM25}}(x_q, x_i)
&= \sum_{t \in x_q} \mathrm{IDF}(t) \cdot \\[-0.2em]
&\!\!
\frac{f(t, x_i)\,(k_1 + 1)}
{f(t, x_i)
+ k_1 \left(1 - b
+ b \cdot \frac{|x_i|}{\mathrm{avgdl}}\right)}
\end{aligned}
\end{equation}

where $f(t, x_i)$ denotes the term frequency of token $t$ in $x_i$, $|x_i|$ is the length of $x_i$, and $\mathrm{avgdl}$ is the average document length in $\mathcal{D}_S$. The inverse document frequency is defined as:
\begin{equation}
\mathrm{IDF}(t) = \log \frac{N - n_t + 0.5}{n_t + 0.5},
\end{equation}
with $n_t$ being the number of examples containing token $t$. We follow standard practice and set $k_1$ and $b$ to fixed constants.

The retrieved set is obtained by:
\begin{equation}
\mathcal{R}_k^{\text{BM25}}(x_q)
=
\operatorname*{arg\,topk}_{x_i \in \mathcal{D}_S}
\ \mathrm{sim}_{\text{BM25}}(x_q, x_i).
\end{equation}

Unlike dense retrieval, BM25 relies on exact lexical overlap and does not use learned representations, providing a strong non-neural baseline.

\subsection{TopK+ConE: Data- \& Model-Dependent Demonstration Selection}
Given a test input $x$ and a demonstration pool $\mathcal{D}$, we aim to select $N$ in-context demonstrations $c$.
We formalize effective demonstration selection as minimizing the conditional entropy of the test input $x$ under an inference model $p_\theta$:

\begin{equation}
c^{*}=\arg\min_{c\in \mathcal{C}} \; H_{\theta}(x\mid c),
\label{eq:cone_obj}
\end{equation}
where $\mathcal{C}$ denotes the candidate set of demonstration groups and
\begin{equation}
H_{\theta}(x\mid c)
\;=\;
-\mathbb{E}_{x}\big[\log p_{\theta}(x\mid c)\big].
\label{eq:cond_entropy_def}
\end{equation}
Using the chain rule of (cross-)entropy, Eq.~\eqref{eq:cone_obj} can be rewritten as
\begin{equation}
H_{\theta}(x\mid c)
=
H_{\theta}(x,c)-H_{\theta}(c),
\label{eq:cone_rewrite}
\end{equation}
where $H_{\theta}(x,c)$ is the cross-entropy of the concatenated prompt (demonstrations plus test input) and $H_{\theta}(c)$ is the cross-entropy of the demonstrations alone, both estimated by the same inference model $p_\theta$.
Thus, TopK+ConE ranks candidate demonstrations by the difference $H_{\theta}(x,c)-H_{\theta}(c)$.

\paragraph{Select-then-Rerank.}
Enumerating $\mathcal{C}$ is infeasible, so we employ a two-stage pipeline.
First, a data-dependent retriever $r(\cdot)$ (e.g., embedding-based nearest neighbors) selects a candidate set $\mathcal{D}_K(x)\subset\mathcal{D}$ with $|\mathcal{D}_K(x)|=K$:
\begin{equation}
\mathcal{D}_K(x)
=
\operatorname{TopK}_{d\in\mathcal{D}}\; s\!\left(r(x),r(d)\right),
\label{eq:topk_select}
\end{equation}
where $s(\cdot,\cdot)$ is a similarity function.
Then we form a small set of candidate demonstration groups $\mathcal{C}(x)\subseteq \binom{\mathcal{D}_K(x)}{N}$ and rerank each $c\in\mathcal{C}(x)$ by ConE:
\begin{equation}
\begin{aligned}
\operatorname{score}(c;x)
=
H_{\theta}(x,c)-H_{\theta}(c),
\qquad \\
c^{*}=\arg\min_{c\in\mathcal{C}(x)} \operatorname{score}(c;x).
\label{eq:cone_score}
\end{aligned}
\end{equation}
Finally, the prompt $\langle c^{*},x\rangle$ is fed into the inference model to produce the prediction $\hat{y}$.

\subsubsection{Usage in ICL}
\label{app:retrieval_usage}

For all experiments, the retrieved demonstrations $\mathcal{R}_k(\cdot)$ are concatenated with the query in a fixed prompt template and fed into the frozen language model. Apart from the retrieval mechanism, all other components of the ICL pipeline are kept identical to ensure fair comparison across retrieval methods.

\subsection{Results of BM25 Retrieval}
\label{app:bm25}

\begin{figure*}[t!]
    \centering
    \includegraphics[width=\linewidth]{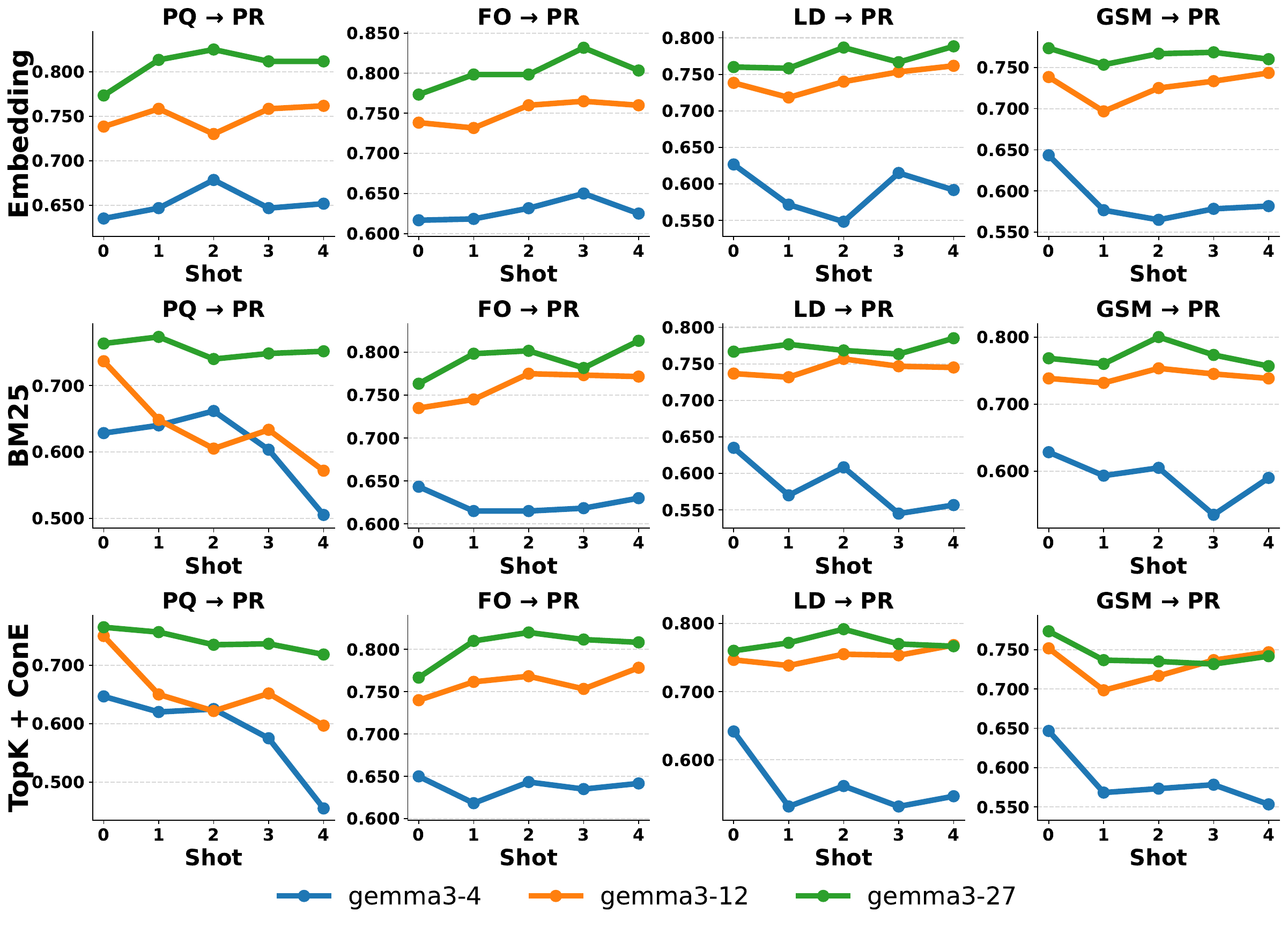}
    \caption{\textbf{Shot–performance scaling across cross-domain ICL settings.} Accuracy curves for multiple models and transfer directions show that increasing the number of demonstrations does not yield reliable positive scaling. While structurally aligned transfers exhibit eAR-LSATy saturation, mismatched pairs often experience instability or negative transfer as shot size increases.}
    \label{fig:shot_gemma_PW}
\end{figure*}

Figures~11 and~12 report cross-domain ICL performance when demonstrations are retrieved using BM25 lexical matching, under the same experimental protocol as the embedding-based retriever in the main text.

Overall, BM25 exhibits substantially weaker and less stable cross-domain transfer compared to dense embedding retrieval. As shown in Figure~11, BM25 rarely achieves consistent improvements over zero-shot baselines across source--target pairs, and negative transfer is more frequent, especially when the source and target domains differ in surface form or vocabulary. This behavior contrasts with embedding-based retrieval, which demonstrates more reliable gains under similar settings.

The performance gap becomes more pronounced as the number of demonstrations increases. Figure~12 shows that BM25 does not benefit from shot scaling in most cross-domain configurations. In several cases, increasing the number of BM25-retrieved demonstrations leads to flat or even degraded performance, indicating that lexical similarity alone is insufficient to identify demonstrations that provide transferable reasoning patterns. This phenomenon is particularly evident for targets requiring abstract logical structures, such as FOLIO and AR-LSAT.

Nevertheless, BM25 is not uniformly ineffective. In cases where source and target domains share strong lexical overlap and similar surface templates (e.g., within closely related logical reasoning datasets), BM25 can occasionally match or approach the performance of embedding-based retrieval, especially at small shot numbers. This suggests that BM25 may still serve as a competitive baseline when reasoning structures and language realizations are highly aligned.

Taken together, these results highlight a key limitation of purely lexical retrieval for cross-domain ICL. While BM25 can retrieve demonstrations that are textually similar, it often fails to capture the structural compatibility required for effective reasoning transfer. This finding further supports our main conclusion that cross-domain ICL performance depends more critically on structural and difficulty alignment than on surface-level similarity.

\subsection{Results of Top-K+ConE Retrieval}

\subsection{Statistical Significance Tests}
\FloatBarrier
\label{app:stats}

This appendix reports the statistical tests used throughout the empirical study and summarizes the corresponding numerical results. Unless otherwise stated, all tests are conducted on paired experimental outcomes under identical settings (same model, source--target pair, and shot number).

\subsubsection{Dense Retrieval vs.\ BM25 Retrieval (RQ1)}
\label{app:stats_rq1}

We compare embedding-based dense retrieval and BM25 sparse retrieval using paired Exact Match (EM) scores across all models and cross-domain transfer directions. Since EM differences are not guaranteed to be normally distributed, we primarily adopt the Wilcoxon signed-rank test, supplemented by paired $t$-tests for reference.

\begin{table}[htbp]
\centering
\small
\begin{tabular}{l c}
\toprule
Metric & Value \\
\midrule
Mean EM (Dense Retrieval) & 0.6658 \\
Mean EM (BM25) & 0.6671 \\
Mean Difference (BM25 $-$ Dense) & +0.0013 \\
Wilcoxon statistic & 29845.0 \\
Wilcoxon $p$-value & 0.2803 \\
Paired $t$-test $t$ & 1.1458 \\
Paired $t$-test $p$-value & 0.2526 \\
\bottomrule
\end{tabular}
\caption{Overall significance test comparing BM25 and dense retrieval.}
\label{tab:bm25_overall}
\end{table}

Overall, BM25 does not yield a statistically significant improvement over dense retrieval (Wilcoxon $p=0.28$), with an average gain of only $+0.13$ percentage points in EM.

\paragraph{Per-model analysis.}
Table~\ref{tab:bm25_model} reports significance tests stratified by model.

\begin{table}[h!]
\centering
\small
\begin{tabular}{l c c c c}
\toprule
Model & $n$ & Mean Diff & Wilcoxon $p$ & $t$-test $p$ \\
\midrule
LLaMA3-8B & 100 & +0.0029 & 0.818 & 0.306 \\
Qwen2.5-14B & 100 & +0.0031 & 0.093 & 0.157 \\
Qwen3-4B & 100 & +0.0053 & 0.135 & 0.024 \\
Qwen2.5-32B & 145 & $-$0.0009 & 0.692 & 0.626 \\
Qwen2.5-7B & 100 & +0.0005 & 0.850 & 0.790 \\
\bottomrule
\end{tabular}
\caption{Per-model significance tests for BM25 vs.\ dense retrieval.}
\label{tab:bm25_model}
\end{table}

Only the smallest model (Qwen3-4B) shows weak sensitivity under the $t$-test, while all other models exhibit no significant differences.

\paragraph{Per-direction analysis.}
Certain source--target transfer directions exhibit statistically significant differences (Table~\ref{tab:bm25_direction}), suggesting that dense retrieval is more effective when semantic distance is large and reasoning structure is complex.

\begin{table}[h!]
\centering
\small
\begin{tabular}{l c c c}
\toprule
Source $\rightarrow$ Target & Mean Diff & Wilcoxon $p$ & $t$-test $p$ \\
\midrule
FOL $\rightarrow$ GSM & $-$0.0039 & 0.0030 & 0.0035 \\
LD $\rightarrow$ FOL & $-$0.0108 & 0.0089 & 0.0071 \\
PQA $\rightarrow$ PW & $-$0.0497 & 0.0679 & 0.0333 \\
PW $\rightarrow$ GSM & $-$0.0055 & 0.0254 & 0.0179 \\
PW $\rightarrow$ LD & +0.0407 & 0.0625 & 0.0021 \\
\bottomrule
\end{tabular}
\caption{Per-direction significance tests for BM25 vs.\ dense retrieval.}
\label{tab:bm25_direction}
\end{table}

\subsubsection{Model Size and Monotonicity (RQ2)}
\label{app:stats_rq6}

We further examine the relationship between model size and few-shot monotonicity.

\begin{table}[h!]
\centering
\small
\begin{tabular}{l c c}
\toprule
Metric & Coefficient & $p$-value \\
\midrule
Pearson $r$ & 0.2070 & $1.56\times10^{-4}$ \\
Spearman $\rho$ & 0.2079 & $1.46\times10^{-4}$ \\
Kendall $\tau$ & 0.1510 & $1.50\times10^{-4}$ \\
\bottomrule
\end{tabular}
\caption{Correlation between model size and few-shot monotonicity.}
\label{tab:modelsize_corr}
\end{table}

All correlations are significantly positive, supporting the existence of a model-size-dependent ``example absorption'' threshold for effective cross-domain few-shot learning.

\subsubsection{Few-shot Monotonicity (RQ3)}
\label{app:stats_rq2}

We analyze the relationship between task complexity (defined as the product of source and target option counts) and few-shot monotonicity, measured by the Spearman correlation between shot number and EM.

\begin{table}[h!]
\centering
\small
\begin{tabular}{l c c}
\toprule
Correlation Metric & Coefficient & $p$-value \\
\midrule
Pearson $r$ & $-$0.2336 & $1.87\times10^{-5}$ \\
Spearman $\rho$ & $-$0.3864 & $3.72\times10^{-13}$ \\
Kendall $\tau$ & $-$0.2808 & $1.22\times10^{-12}$ \\
\bottomrule
\end{tabular}
\caption{Overall correlation between task complexity and few-shot monotonicity.}
\label{tab:difficulty_overall}
\end{table}

All three correlation coefficients are significantly negative, indicating that higher task complexity is associated with less reliable few-shot scaling.

\subsection{Full Results}
\label{app:full_results}

Tables~10--13 report the complete cross-domain ICL results across all evaluated models, source--target domain pairs, retrieval methods, and shot settings. These tables serve as comprehensive supplementary evidence for the main empirical observations discussed in the paper, and are intended to demonstrate that our conclusions are not driven by a small subset of tasks or configurations.

\paragraph{Non-monotonic shot scaling.}
Across Tables~10--13, increasing the number of demonstrations does not consistently lead to performance gains in cross-domain ICL. While moderate numbers of shots can be beneficial in some settings, higher shot counts frequently exhibit diminishing returns or even performance degradation. This non-monotonic behavior is especially pronounced when the source and target domains differ substantially in reasoning structure or difficulty, supporting our claim that cross-domain ICL is not governed by simple shot scaling laws.

\paragraph{Strong dependence on model capacity.}
The full results further reveal a clear interaction between model size and demonstration effectiveness. Smaller models tend to show high variance and are more susceptible to negative transfer when additional demonstrations are introduced, whereas larger models exhibit more stable improvements and are better able to absorb cross-domain demonstrations. This pattern consistently appears across multiple targets and retrieval settings, indicating the presence of a model-dependent example absorption threshold.

\paragraph{Target-domain sensitivity.}
Another salient pattern observed in Tables~10--13 is the strong dependence of cross-domain ICL gains on the target domain. Certain target tasks benefit more reliably from retrieved demonstrations, while others show limited or inconsistent improvements regardless of the source domain or shot count. This target-specific sensitivity suggests that the effectiveness of demonstrations is constrained by the reasoning requirements of the target task, rather than by the availability of semantically similar examples alone.

\paragraph{Retrieval method comparison.}
Comparing dense retrieval and BM25 across the full results, we observe that neither method universally dominates the other. Dense retrieval often provides stronger gains when semantic alignment between source and target exists, while BM25 remains competitive or more stable in settings where lexical overlap is informative. The variability across Tables~10--13 highlights that retrieval quality alone cannot guarantee effective cross-domain transfer without considering task structure and difficulty.

Overall, the comprehensive results in Tables~10--13 reinforce our central findings: cross-domain ICL performance is highly heterogeneous, sensitive to model capacity and target-domain characteristics, and cannot be reliably improved by increasing the number of demonstrations or by relying on a single retrieval strategy.

\clearpage

\begin{figure*}[t!]
    \centering
    \includegraphics[width=\linewidth]{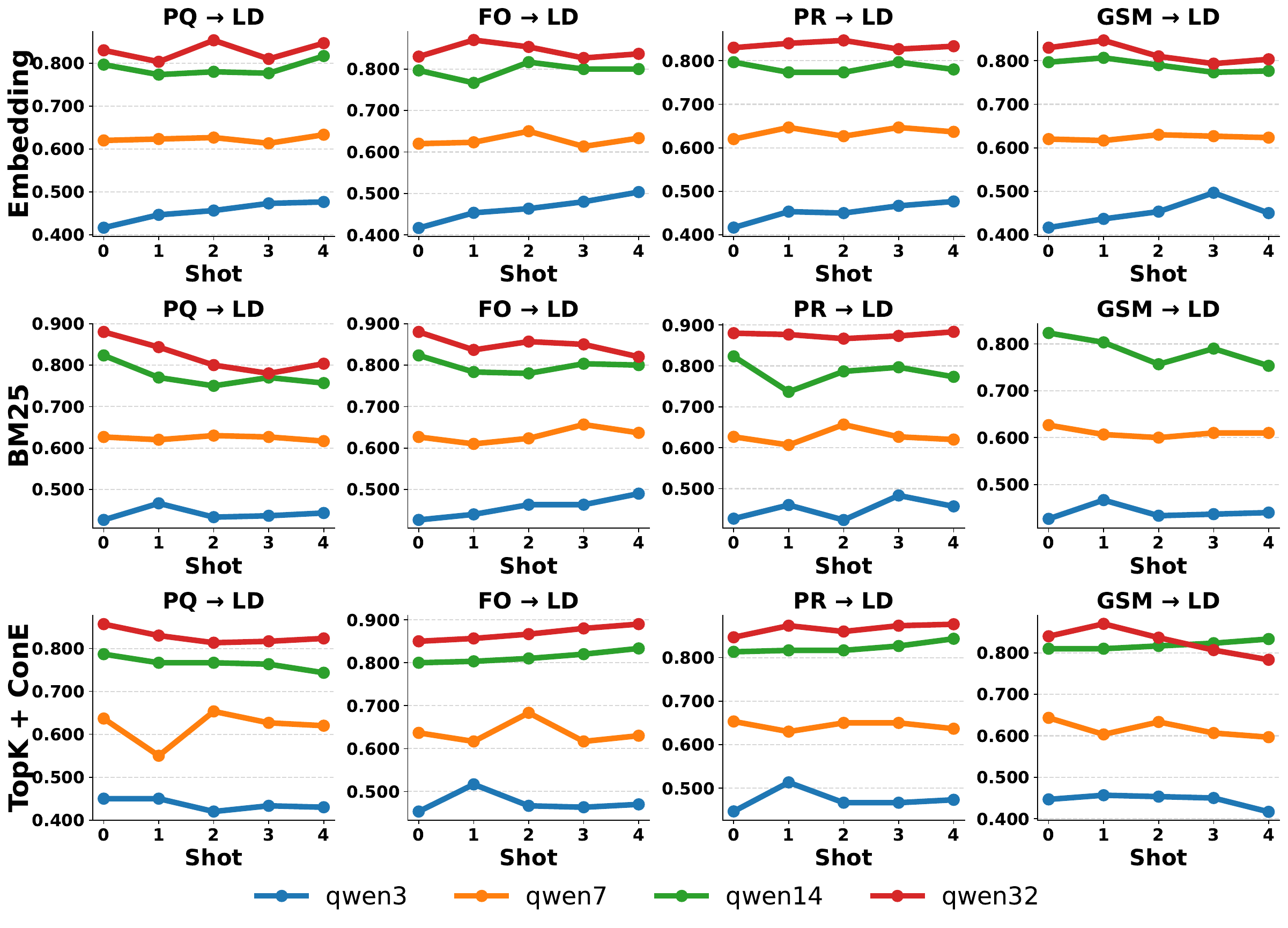}
    \caption{\textbf{Shot–performance scaling across cross-domain ICL settings.} Accuracy curves for multiple models and transfer directions show that increasing the number of demonstrations does not yield reliable positive scaling. While structurally aligned transfers exhibit eAR-LSATy saturation, mismatched pairs often experience instability or negative transfer as shot size increases.}
    \label{fig:shot_bm25}
\end{figure*}

\begin{figure*}[t!]
    \centering
    \includegraphics[width=\linewidth]{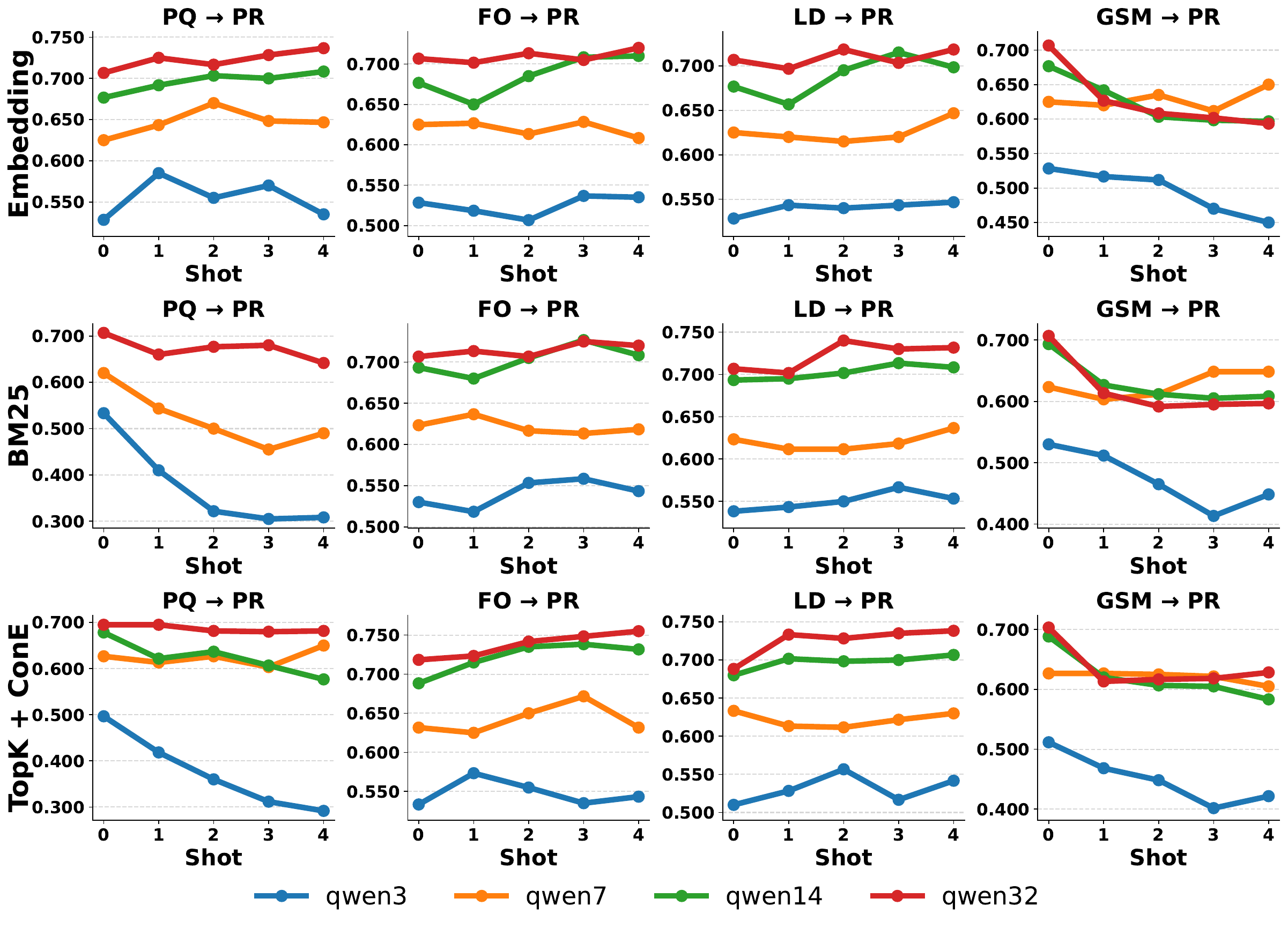}
    \caption{\textbf{Shot–performance scaling across cross-domain ICL settings.} Accuracy curves for multiple models and transfer directions show that increasing the number of demonstrations does not yield reliable positive scaling. While structurally aligned transfers exhibit eAR-LSATy saturation, mismatched pairs often experience instability or negative transfer as shot size increases.}
    \label{fig:shot_qwen_PW}
\end{figure*}

\begin{figure*}[t!]
    \centering
    \includegraphics[width=\linewidth]{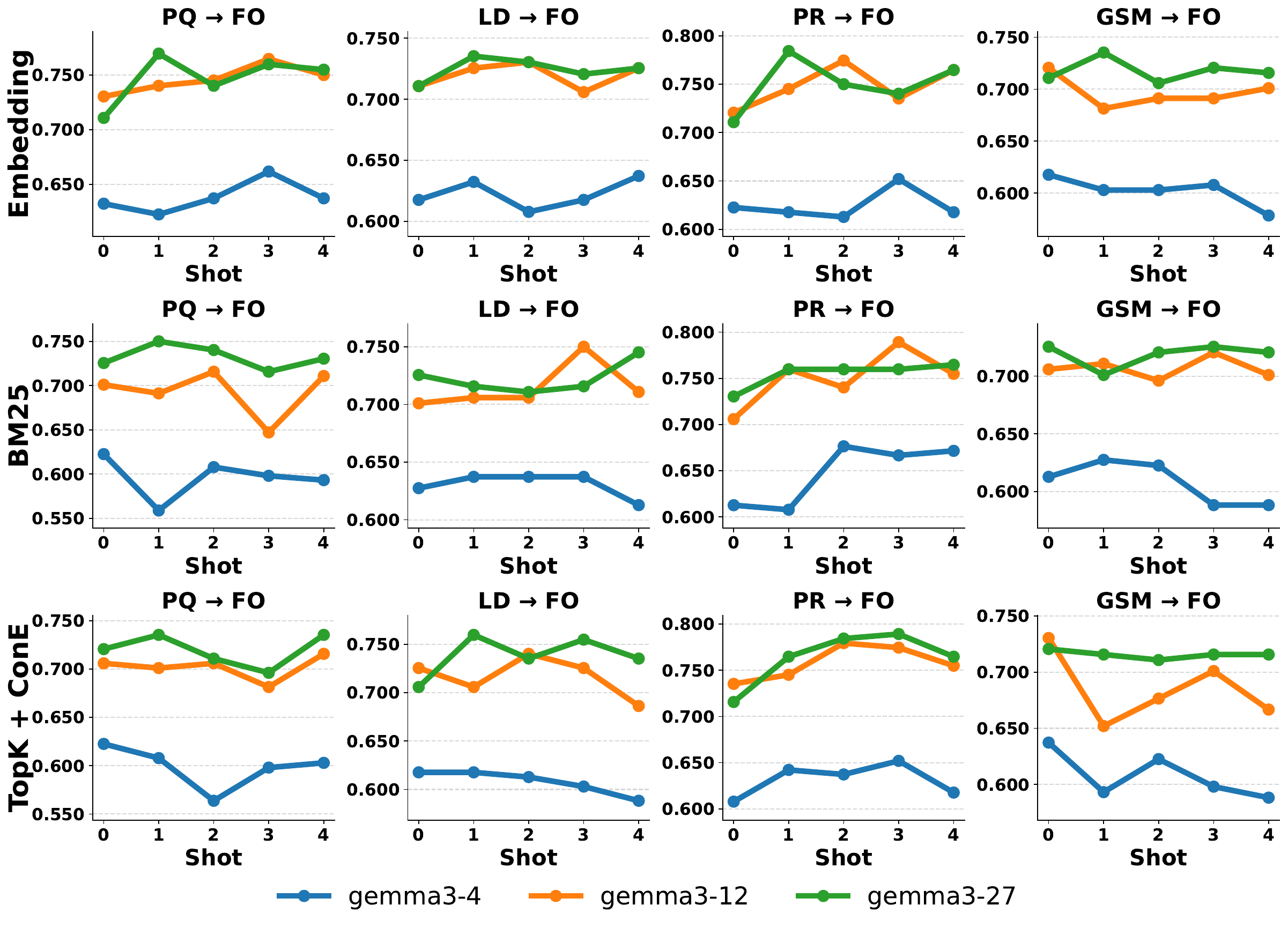}
    \caption{\textbf{Shot–performance scaling across cross-domain ICL settings.} Accuracy curves for multiple models and transfer directions show that increasing the number of demonstrations does not yield reliable positive scaling. While structurally aligned transfers exhibit eAR-LSATy saturation, mismatched pairs often experience instability or negative transfer as shot size increases.}
    \label{fig:shot_gemma_FO}
\end{figure*}

\begin{figure*}[t!]
    \centering
    \includegraphics[width=\linewidth]{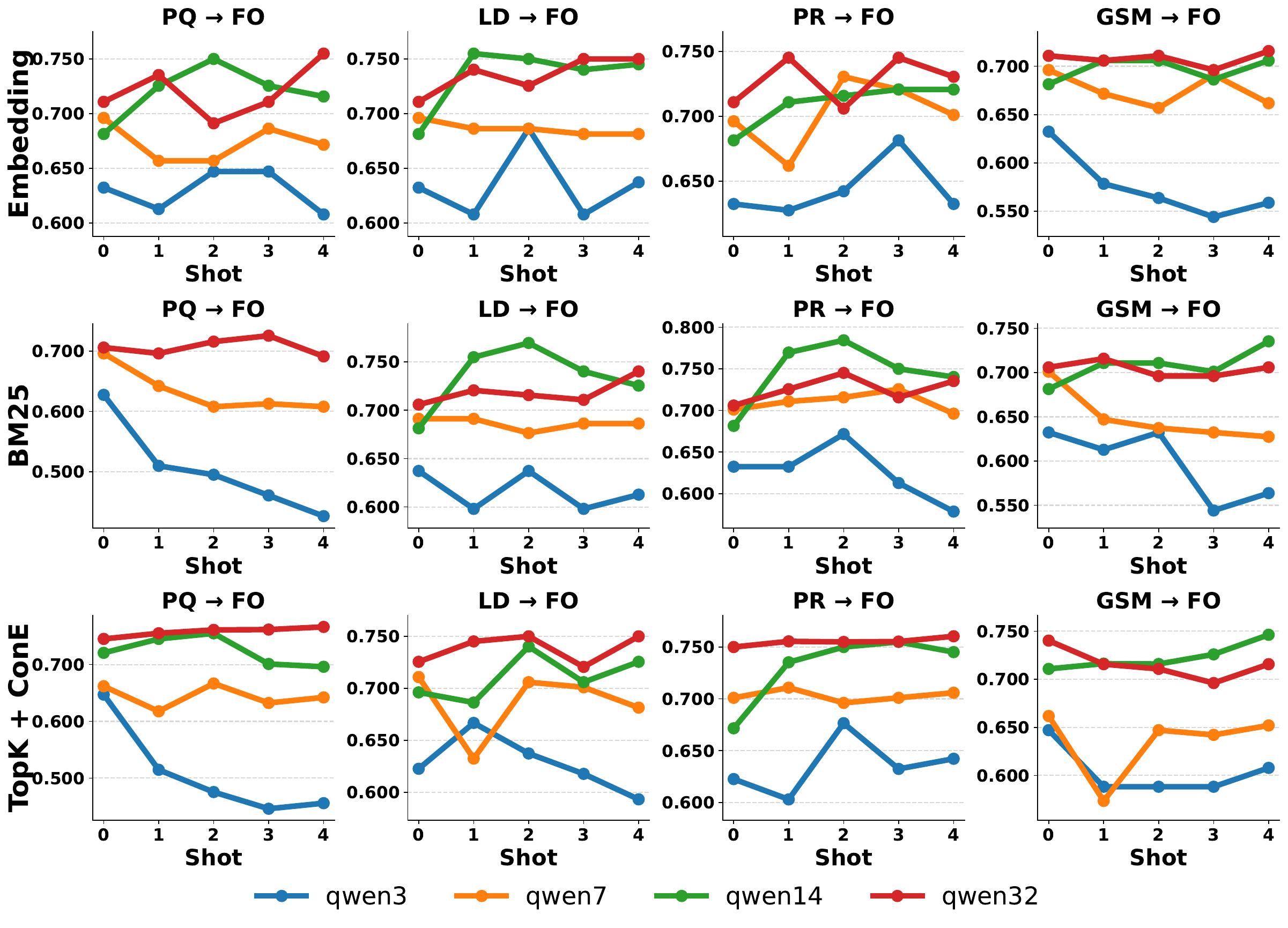}
    \caption{\textbf{Shot–performance scaling across cross-domain ICL settings.} Accuracy curves for multiple models and transfer directions show that increasing the number of demonstrations does not yield reliable positive scaling. While structurally aligned transfers exhibit eAR-LSATy saturation, mismatched pairs often experience instability or negative transfer as shot size increases.}
    \label{fig:shot_qwen_FO}
\end{figure*}

\begin{sidewaystable*}[t]
\centering
\small
\resizebox{1.05\textwidth}{!}{%
\setlength{\tabcolsep}{3pt}
\begin{tabular}{ll
ccccc
ccccc
ccccc
ccccc
ccccc
ccccc}
\toprule
\textbf{Model} & \textbf{Source}
& \multicolumn{5}{c}{\textbf{AR-LSAT}}
& \multicolumn{5}{c}{\textbf{FOLIO}}
& \multicolumn{5}{c}{\textbf{LogicalDeduction}}
& \multicolumn{5}{c}{\textbf{ProntoQA}}
& \multicolumn{5}{c}{\textbf{ProofWriter}}
& \multicolumn{5}{c}{\textbf{gsm8k}} \\
\cmidrule(lr){3-7}
\cmidrule(lr){8-12}
\cmidrule(lr){13-17}
\cmidrule(lr){18-22}
\cmidrule(lr){23-27}
\cmidrule(lr){28-32}
& & 0 & 1 & 2 & 3 & 4
  & 0 & 1 & 2 & 3 & 4
  & 0 & 1 & 2 & 3 & 4
  & 0 & 1 & 2 & 3 & 4
  & 0 & 1 & 2 & 3 & 4
  & 0 & 1 & 2 & 3 & 4 \\
\midrule
\multirow{5}{*}{qwen3} & FOLIO & 0.235 & \textbf{\textcolor{red}{0.252}} & 0.178 & \textbf{\textcolor{red}{0.257}} & 0.209 & 0.632 & 0.593 & 0.627 & \textbf{\textcolor{red}{0.642}} & 0.623 & 0.417 & \textbf{\textcolor{red}{0.453}} & \textbf{\textcolor{red}{0.463}} & \textbf{\textcolor{red}{0.480}} & \textbf{\textcolor{red}{0.503}} & 0.858 & 0.846 & 0.846 & 0.844 & 0.844 & 0.528 & 0.518 & 0.507 & \textbf{\textcolor{red}{0.537}} & \textbf{\textcolor{red}{0.535}} & 0.737 & 0.726 & 0.682 & 0.659 & 0.645 \\
& LogicalDeduction & 0.235 & 0.191 & 0.235 & 0.209 & \textbf{\textcolor{red}{0.265}} & 0.632 & 0.608 & \textbf{\textcolor{red}{0.686}} & 0.608 & \textbf{\textcolor{red}{0.637}} & 0.417 & \textbf{\textcolor{red}{0.487}} & \textbf{\textcolor{red}{0.513}} & \textbf{\textcolor{red}{0.470}} & \textbf{\textcolor{red}{0.450}} & 0.858 & 0.856 & 0.820 & 0.832 & 0.832 & 0.528 & \textbf{\textcolor{red}{0.543}} & \textbf{\textcolor{red}{0.540}} & \textbf{\textcolor{red}{0.543}} & \textbf{\textcolor{red}{0.547}} & 0.737 & 0.726 & 0.717 & 0.699 & 0.663 \\
& ProntoQA & 0.235 & 0.187 & 0.187 & -- & -- & 0.632 & 0.613 & \textbf{\textcolor{red}{0.647}} & \textbf{\textcolor{red}{0.647}} & 0.608 & 0.417 & \textbf{\textcolor{red}{0.447}} & \textbf{\textcolor{red}{0.457}} & \textbf{\textcolor{red}{0.473}} & \textbf{\textcolor{red}{0.477}} & 0.858 & 0.854 & \textbf{\textcolor{red}{0.906}} & \textbf{\textcolor{red}{0.896}} & \textbf{\textcolor{red}{0.876}} & 0.528 & \textbf{\textcolor{red}{0.585}} & \textbf{\textcolor{red}{0.555}} & \textbf{\textcolor{red}{0.570}} & \textbf{\textcolor{red}{0.535}} & 0.737 & 0.733 & 0.714 & 0.723 & 0.710 \\
& ProofWriter & 0.235 & 0.213 & 0.226 & 0.196 & 0.187 & 0.632 & 0.627 & \textbf{\textcolor{red}{0.642}} & \textbf{\textcolor{red}{0.681}} & 0.632 & 0.417 & \textbf{\textcolor{red}{0.453}} & \textbf{\textcolor{red}{0.450}} & \textbf{\textcolor{red}{0.467}} & \textbf{\textcolor{red}{0.477}} & 0.858 & 0.856 & \textbf{\textcolor{red}{0.872}} & 0.846 & \textbf{\textcolor{red}{0.860}} & 0.528 & 0.528 & \textbf{\textcolor{red}{0.548}} & 0.513 & 0.510 & 0.737 & \textbf{\textcolor{red}{0.742}} & 0.689 & 0.651 & 0.624 \\
& gsm8k & 0.235 & 0.174 & 0.226 & 0.217 & 0.230 & 0.632 & 0.578 & 0.564 & 0.544 & 0.559 & 0.417 & \textbf{\textcolor{red}{0.437}} & \textbf{\textcolor{red}{0.453}} & \textbf{\textcolor{red}{0.497}} & \textbf{\textcolor{red}{0.450}} & 0.858 & 0.752 & 0.734 & 0.592 & 0.574 & 0.528 & 0.517 & 0.512 & 0.470 & 0.450 & 0.737 & 0.729 & \textbf{\textcolor{red}{0.753}} & 0.721 & 0.665 \\
\midrule
\multirow{5}{*}{qwen7} & FOLIO & 0.257 & \textbf{\textcolor{red}{0.270}} & 0.230 & 0.222 & 0.217 & 0.696 & 0.676 & 0.696 & \textbf{\textcolor{red}{0.711}} & \textbf{\textcolor{red}{0.701}} & 0.620 & \textbf{\textcolor{red}{0.623}} & \textbf{\textcolor{red}{0.650}} & 0.613 & \textbf{\textcolor{red}{0.633}} & 0.968 & 0.958 & 0.964 & 0.968 & \textbf{\textcolor{red}{0.978}} & 0.625 & \textbf{\textcolor{red}{0.627}} & 0.613 & \textbf{\textcolor{red}{0.628}} & 0.608 & 0.890 & 0.885 & 0.867 & 0.867 & 0.828 \\
& LogicalDeduction & 0.257 & 0.243 & 0.257 & 0.257 & 0.248 & 0.696 & 0.686 & 0.686 & 0.681 & 0.681 & 0.620 & \textbf{\textcolor{red}{0.647}} & \textbf{\textcolor{red}{0.683}} & \textbf{\textcolor{red}{0.640}} & \textbf{\textcolor{red}{0.630}} & 0.968 & \textbf{\textcolor{red}{0.980}} & \textbf{\textcolor{red}{0.980}} & \textbf{\textcolor{red}{0.986}} & \textbf{\textcolor{red}{0.988}} & 0.625 & 0.620 & 0.615 & 0.620 & \textbf{\textcolor{red}{0.647}} & 0.890 & 0.880 & 0.876 & 0.880 & 0.862 \\
& ProntoQA & 0.257 & 0.226 & 0.200 & 0.222 & 0.226 & 0.696 & 0.657 & 0.657 & 0.686 & 0.672 & 0.620 & \textbf{\textcolor{red}{0.623}} & \textbf{\textcolor{red}{0.627}} & 0.613 & \textbf{\textcolor{red}{0.633}} & 0.968 & 0.944 & 0.950 & \textbf{\textcolor{red}{0.972}} & 0.964 & 0.625 & \textbf{\textcolor{red}{0.643}} & \textbf{\textcolor{red}{0.670}} & \textbf{\textcolor{red}{0.648}} & \textbf{\textcolor{red}{0.647}} & 0.890 & 0.864 & 0.877 & 0.867 & 0.859 \\
& ProofWriter & 0.257 & 0.230 & \textbf{\textcolor{red}{0.287}} & 0.213 & \textbf{\textcolor{red}{0.283}} & 0.696 & 0.662 & \textbf{\textcolor{red}{0.730}} & \textbf{\textcolor{red}{0.721}} & \textbf{\textcolor{red}{0.701}} & 0.620 & \textbf{\textcolor{red}{0.647}} & \textbf{\textcolor{red}{0.627}} & \textbf{\textcolor{red}{0.647}} & \textbf{\textcolor{red}{0.637}} & 0.968 & \textbf{\textcolor{red}{0.970}} & \textbf{\textcolor{red}{0.988}} & \textbf{\textcolor{red}{0.984}} & \textbf{\textcolor{red}{0.992}} & 0.625 & \textbf{\textcolor{red}{0.627}} & \textbf{\textcolor{red}{0.655}} & \textbf{\textcolor{red}{0.630}} & \textbf{\textcolor{red}{0.658}} & 0.890 & 0.860 & 0.867 & 0.839 & 0.826 \\
& gsm8k & 0.257 & 0.226 & 0.217 & 0.213 & 0.230 & 0.696 & 0.672 & 0.657 & 0.691 & 0.662 & 0.620 & 0.617 & \textbf{\textcolor{red}{0.630}} & \textbf{\textcolor{red}{0.627}} & \textbf{\textcolor{red}{0.623}} & 0.968 & 0.942 & 0.924 & 0.924 & 0.934 & 0.625 & 0.620 & \textbf{\textcolor{red}{0.635}} & 0.612 & \textbf{\textcolor{red}{0.650}} & 0.890 & 0.882 & 0.889 & 0.885 & 0.877 \\
\midrule
\multirow{5}{*}{qwen14} & FOLIO & 0.348 & 0.274 & 0.335 & \textbf{\textcolor{red}{0.374}} & 0.322 & 0.681 & \textbf{\textcolor{red}{0.750}} & \textbf{\textcolor{red}{0.745}} & \textbf{\textcolor{red}{0.765}} & \textbf{\textcolor{red}{0.760}} & 0.797 & 0.767 & \textbf{\textcolor{red}{0.817}} & \textbf{\textcolor{red}{0.800}} & \textbf{\textcolor{red}{0.800}} & 0.948 & \textbf{\textcolor{red}{0.978}} & \textbf{\textcolor{red}{0.978}} & \textbf{\textcolor{red}{0.970}} & \textbf{\textcolor{red}{0.978}} & 0.677 & 0.650 & \textbf{\textcolor{red}{0.685}} & \textbf{\textcolor{red}{0.708}} & \textbf{\textcolor{red}{0.710}} & 0.932 & 0.927 & 0.927 & 0.929 & 0.930 \\
& LogicalDeduction & 0.348 & 0.296 & 0.317 & 0.265 & 0.317 & 0.681 & \textbf{\textcolor{red}{0.755}} & \textbf{\textcolor{red}{0.750}} & \textbf{\textcolor{red}{0.740}} & \textbf{\textcolor{red}{0.745}} & 0.797 & \textbf{\textcolor{red}{0.810}} & 0.777 & \textbf{\textcolor{red}{0.810}} & \textbf{\textcolor{red}{0.807}} & 0.948 & \textbf{\textcolor{red}{0.978}} & \textbf{\textcolor{red}{0.956}} & \textbf{\textcolor{red}{0.968}} & \textbf{\textcolor{red}{0.974}} & 0.677 & 0.657 & \textbf{\textcolor{red}{0.695}} & \textbf{\textcolor{red}{0.715}} & \textbf{\textcolor{red}{0.698}} & 0.932 & 0.929 & 0.926 & 0.932 & 0.925 \\
& ProntoQA & 0.348 & 0.317 & \textbf{\textcolor{red}{0.357}} & 0.326 & \textbf{\textcolor{red}{0.352}} & 0.681 & \textbf{\textcolor{red}{0.725}} & \textbf{\textcolor{red}{0.750}} & \textbf{\textcolor{red}{0.725}} & \textbf{\textcolor{red}{0.716}} & 0.797 & 0.773 & 0.780 & 0.777 & \textbf{\textcolor{red}{0.817}} & 0.948 & \textbf{\textcolor{red}{0.972}} & \textbf{\textcolor{red}{0.978}} & \textbf{\textcolor{red}{0.982}} & \textbf{\textcolor{red}{0.984}} & 0.677 & \textbf{\textcolor{red}{0.692}} & \textbf{\textcolor{red}{0.703}} & \textbf{\textcolor{red}{0.700}} & \textbf{\textcolor{red}{0.708}} & 0.932 & 0.929 & 0.927 & 0.929 & 0.930 \\
& ProofWriter & 0.348 & 0.287 & 0.330 & \textbf{\textcolor{red}{0.365}} & 0.348 & 0.681 & \textbf{\textcolor{red}{0.711}} & \textbf{\textcolor{red}{0.716}} & \textbf{\textcolor{red}{0.721}} & \textbf{\textcolor{red}{0.721}} & 0.797 & 0.773 & 0.773 & 0.797 & 0.780 & 0.948 & \textbf{\textcolor{red}{0.968}} & \textbf{\textcolor{red}{0.982}} & \textbf{\textcolor{red}{0.980}} & \textbf{\textcolor{red}{0.976}} & 0.677 & \textbf{\textcolor{red}{0.738}} & \textbf{\textcolor{red}{0.752}} & \textbf{\textcolor{red}{0.762}} & \textbf{\textcolor{red}{0.737}} & 0.932 & 0.932 & 0.929 & \textbf{\textcolor{red}{0.936}} & 0.932 \\
& gsm8k & 0.348 & 0.326 & 0.330 & 0.300 & 0.296 & 0.681 & \textbf{\textcolor{red}{0.706}} & \textbf{\textcolor{red}{0.706}} & \textbf{\textcolor{red}{0.686}} & \textbf{\textcolor{red}{0.706}} & 0.797 & \textbf{\textcolor{red}{0.807}} & 0.790 & 0.773 & 0.777 & 0.948 & 0.942 & 0.920 & 0.892 & 0.896 & 0.677 & 0.642 & 0.603 & 0.598 & 0.597 & 0.932 & 0.922 & 0.924 & 0.923 & 0.926 \\
\midrule
\multirow{5}{*}{qwen32} & FOLIO & 0.322 & \textbf{\textcolor{red}{0.335}} & 0.278 & \textbf{\textcolor{red}{0.365}} & \textbf{\textcolor{red}{0.343}} & 0.711 & \textbf{\textcolor{red}{0.740}} & \textbf{\textcolor{red}{0.745}} & \textbf{\textcolor{red}{0.735}} & \textbf{\textcolor{red}{0.745}} & 0.830 & \textbf{\textcolor{red}{0.870}} & \textbf{\textcolor{red}{0.853}} & 0.827 & \textbf{\textcolor{red}{0.837}} & 1.000 & 1.000 & 1.000 & 1.000 & 1.000 & 0.707 & 0.702 & \textbf{\textcolor{red}{0.713}} & 0.705 & \textbf{\textcolor{red}{0.720}} & 0.883 & \textbf{\textcolor{red}{0.900}} & \textbf{\textcolor{red}{0.907}} & \textbf{\textcolor{red}{0.900}} & \textbf{\textcolor{red}{0.896}} \\
& LogicalDeduction & 0.322 & \textbf{\textcolor{red}{0.343}} & \textbf{\textcolor{red}{0.335}} & 0.283 & 0.313 & 0.711 & \textbf{\textcolor{red}{0.740}} & \textbf{\textcolor{red}{0.725}} & \textbf{\textcolor{red}{0.750}} & \textbf{\textcolor{red}{0.750}} & 0.830 & \textbf{\textcolor{red}{0.867}} & \textbf{\textcolor{red}{0.860}} & \textbf{\textcolor{red}{0.853}} & 0.820 & 1.000 & 0.998 & 0.998 & 1.000 & 1.000 & 0.707 & 0.697 & \textbf{\textcolor{red}{0.718}} & 0.703 & \textbf{\textcolor{red}{0.718}} & 0.883 & \textbf{\textcolor{red}{0.902}} & \textbf{\textcolor{red}{0.917}} & \textbf{\textcolor{red}{0.914}} & \textbf{\textcolor{red}{0.909}} \\
& ProntoQA & 0.322 & \textbf{\textcolor{red}{0.343}} & 0.313 & \textbf{\textcolor{red}{0.339}} & \textbf{\textcolor{red}{0.383}} & 0.711 & \textbf{\textcolor{red}{0.735}} & 0.691 & 0.711 & \textbf{\textcolor{red}{0.755}} & 0.830 & 0.803 & \textbf{\textcolor{red}{0.853}} & 0.810 & \textbf{\textcolor{red}{0.847}} & 1.000 & 0.998 & 0.998 & 1.000 & 0.998 & 0.707 & \textbf{\textcolor{red}{0.725}} & \textbf{\textcolor{red}{0.717}} & \textbf{\textcolor{red}{0.728}} & \textbf{\textcolor{red}{0.737}} & 0.883 & \textbf{\textcolor{red}{0.903}} & \textbf{\textcolor{red}{0.908}} & \textbf{\textcolor{red}{0.910}} & \textbf{\textcolor{red}{0.899}} \\
& ProofWriter & 0.322 & \textbf{\textcolor{red}{0.343}} & 0.322 & \textbf{\textcolor{red}{0.343}} & \textbf{\textcolor{red}{0.352}} & 0.711 & \textbf{\textcolor{red}{0.745}} & 0.706 & \textbf{\textcolor{red}{0.745}} & \textbf{\textcolor{red}{0.730}} & 0.830 & \textbf{\textcolor{red}{0.840}} & \textbf{\textcolor{red}{0.847}} & 0.827 & \textbf{\textcolor{red}{0.833}} & 1.000 & 0.998 & 1.000 & 0.998 & 1.000 & 0.707 & \textbf{\textcolor{red}{0.732}} & \textbf{\textcolor{red}{0.735}} & \textbf{\textcolor{red}{0.753}} & \textbf{\textcolor{red}{0.757}} & 0.883 & \textbf{\textcolor{red}{0.889}} & \textbf{\textcolor{red}{0.910}} & \textbf{\textcolor{red}{0.885}} & 0.860 \\
& gsm8k & 0.322 & \textbf{\textcolor{red}{0.348}} & 0.322 & \textbf{\textcolor{red}{0.361}} & \textbf{\textcolor{red}{0.352}} & 0.711 & 0.706 & 0.711 & 0.696 & \textbf{\textcolor{red}{0.716}} & 0.830 & \textbf{\textcolor{red}{0.847}} & 0.810 & 0.793 & 0.803 & 1.000 & 0.992 & 0.984 & 0.984 & 0.982 & 0.707 & 0.627 & 0.608 & 0.602 & 0.593 & 0.883 & \textbf{\textcolor{red}{0.910}} & \textbf{\textcolor{red}{0.920}} & \textbf{\textcolor{red}{0.917}} & \textbf{\textcolor{red}{0.920}} \\
\midrule
\multirow{5}{*}{gemma3-4} & FOLIO & 0.252 & \textbf{\textcolor{red}{0.270}} & \textbf{\textcolor{red}{0.265}} & 0.243 & 0.222 & 0.652 & 0.623 & 0.652 & 0.618 & 0.642 & 0.610 & 0.587 & 0.560 & 0.587 & 0.583 & 0.920 & \textbf{\textcolor{red}{0.930}} & \textbf{\textcolor{red}{0.924}} & 0.920 & 0.910 & 0.617 & \textbf{\textcolor{red}{0.618}} & \textbf{\textcolor{red}{0.632}} & \textbf{\textcolor{red}{0.650}} & \textbf{\textcolor{red}{0.625}} & 0.864 & 0.850 & 0.845 & 0.837 & 0.831 \\
& LogicalDeduction & 0.278 & 0.235 & 0.226 & 0.248 & 0.261 & 0.618 & \textbf{\textcolor{red}{0.632}} & 0.608 & 0.618 & \textbf{\textcolor{red}{0.637}} & 0.613 & \textbf{\textcolor{red}{0.650}} & \textbf{\textcolor{red}{0.637}} & \textbf{\textcolor{red}{0.637}} & 0.573 & 0.930 & 0.880 & 0.892 & 0.906 & 0.898 & 0.627 & 0.572 & 0.548 & 0.615 & 0.592 & 0.859 & 0.843 & 0.843 & 0.845 & 0.840 \\
& ProntoQA & 0.283 & 0.257 & 0.222 & 0.252 & 0.252 & 0.632 & 0.623 & \textbf{\textcolor{red}{0.637}} & \textbf{\textcolor{red}{0.662}} & \textbf{\textcolor{red}{0.637}} & 0.563 & \textbf{\textcolor{red}{0.573}} & \textbf{\textcolor{red}{0.577}} & \textbf{\textcolor{red}{0.583}} & \textbf{\textcolor{red}{0.580}} & 0.910 & \textbf{\textcolor{red}{0.934}} & \textbf{\textcolor{red}{0.946}} & \textbf{\textcolor{red}{0.934}} & \textbf{\textcolor{red}{0.954}} & 0.635 & \textbf{\textcolor{red}{0.647}} & \textbf{\textcolor{red}{0.678}} & \textbf{\textcolor{red}{0.647}} & \textbf{\textcolor{red}{0.652}} & 0.861 & 0.848 & 0.845 & 0.845 & 0.841 \\
& ProofWriter & 0.252 & 0.239 & \textbf{\textcolor{red}{0.261}} & 0.252 & \textbf{\textcolor{red}{0.257}} & 0.623 & 0.618 & 0.613 & \textbf{\textcolor{red}{0.652}} & 0.618 & 0.607 & 0.547 & 0.553 & 0.580 & \textbf{\textcolor{red}{0.610}} & 0.928 & \textbf{\textcolor{red}{0.956}} & \textbf{\textcolor{red}{0.930}} & \textbf{\textcolor{red}{0.930}} & 0.914 & 0.638 & \textbf{\textcolor{red}{0.673}} & \textbf{\textcolor{red}{0.648}} & 0.635 & 0.623 & 0.860 & 0.832 & 0.842 & 0.845 & 0.835 \\
& gsm8k & 0.287 & 0.252 & 0.265 & 0.270 & 0.274 & 0.618 & 0.603 & 0.603 & 0.608 & 0.578 & 0.627 & \textbf{\textcolor{red}{0.630}} & 0.610 & 0.593 & 0.550 & 0.926 & 0.866 & 0.840 & 0.808 & 0.782 & 0.643 & 0.577 & 0.565 & 0.578 & 0.582 & 0.860 & 0.831 & 0.819 & 0.786 & 0.754 \\
\midrule
\multirow{5}{*}{gemma3-12} & FOLIO & 0.300 & \textbf{\textcolor{red}{0.335}} & \textbf{\textcolor{red}{0.335}} & \textbf{\textcolor{red}{0.304}} & \textbf{\textcolor{red}{0.317}} & 0.725 & 0.725 & \textbf{\textcolor{red}{0.770}} & \textbf{\textcolor{red}{0.740}} & \textbf{\textcolor{red}{0.770}} & 0.733 & \textbf{\textcolor{red}{0.773}} & \textbf{\textcolor{red}{0.780}} & \textbf{\textcolor{red}{0.760}} & 0.733 & 0.982 & 0.978 & 0.980 & \textbf{\textcolor{red}{0.988}} & 0.976 & 0.738 & 0.732 & \textbf{\textcolor{red}{0.760}} & \textbf{\textcolor{red}{0.765}} & \textbf{\textcolor{red}{0.760}} & 0.920 & 0.908 & 0.905 & 0.892 & 0.899 \\
& LogicalDeduction & 0.300 & 0.248 & 0.270 & \textbf{\textcolor{red}{0.304}} & \textbf{\textcolor{red}{0.309}} & 0.711 & \textbf{\textcolor{red}{0.725}} & \textbf{\textcolor{red}{0.730}} & 0.706 & \textbf{\textcolor{red}{0.725}} & 0.747 & \textbf{\textcolor{red}{0.847}} & \textbf{\textcolor{red}{0.827}} & \textbf{\textcolor{red}{0.837}} & \textbf{\textcolor{red}{0.807}} & 0.982 & 0.980 & 0.980 & 0.972 & 0.978 & 0.738 & 0.718 & \textbf{\textcolor{red}{0.740}} & \textbf{\textcolor{red}{0.753}} & \textbf{\textcolor{red}{0.762}} & 0.918 & 0.917 & 0.907 & 0.900 & 0.899 \\
& ProntoQA & 0.330 & 0.274 & 0.257 & 0.304 & 0.309 & 0.730 & \textbf{\textcolor{red}{0.740}} & \textbf{\textcolor{red}{0.745}} & \textbf{\textcolor{red}{0.765}} & \textbf{\textcolor{red}{0.750}} & 0.730 & \textbf{\textcolor{red}{0.743}} & \textbf{\textcolor{red}{0.787}} & \textbf{\textcolor{red}{0.747}} & \textbf{\textcolor{red}{0.767}} & 0.982 & -- & \textbf{\textcolor{red}{0.990}} & \textbf{\textcolor{red}{0.996}} & \textbf{\textcolor{red}{0.996}} & 0.738 & \textbf{\textcolor{red}{0.758}} & 0.730 & \textbf{\textcolor{red}{0.758}} & \textbf{\textcolor{red}{0.762}} & 0.920 & 0.910 & 0.912 & 0.900 & 0.899 \\
& ProofWriter & 0.330 & 0.296 & 0.261 & 0.326 & 0.270 & 0.721 & \textbf{\textcolor{red}{0.745}} & \textbf{\textcolor{red}{0.775}} & \textbf{\textcolor{red}{0.735}} & \textbf{\textcolor{red}{0.765}} & 0.730 & \textbf{\textcolor{red}{0.763}} & \textbf{\textcolor{red}{0.773}} & \textbf{\textcolor{red}{0.760}} & \textbf{\textcolor{red}{0.783}} & 0.982 & \textbf{\textcolor{red}{0.986}} & \textbf{\textcolor{red}{0.992}} & \textbf{\textcolor{red}{0.986}} & \textbf{\textcolor{red}{0.992}} & 0.738 & \textbf{\textcolor{red}{0.752}} & \textbf{\textcolor{red}{0.763}} & \textbf{\textcolor{red}{0.792}} & \textbf{\textcolor{red}{0.790}} & 0.921 & 0.910 & 0.905 & 0.898 & 0.895 \\
& gsm8k & 0.330 & 0.322 & 0.291 & 0.322 & 0.313 & 0.721 & 0.681 & 0.691 & 0.691 & 0.701 & 0.730 & \textbf{\textcolor{red}{0.747}} & \textbf{\textcolor{red}{0.747}} & \textbf{\textcolor{red}{0.750}} & \textbf{\textcolor{red}{0.733}} & 0.982 & 0.978 & 0.980 & 0.976 & 0.978 & 0.738 & 0.697 & 0.725 & 0.733 & \textbf{\textcolor{red}{0.743}} & 0.920 & 0.871 & 0.608 & 0.624 & 0.569 \\
\midrule
\multirow{5}{*}{gemma3-27} & FOLIO & 0.348 & \textbf{\textcolor{red}{0.370}} & \textbf{\textcolor{red}{0.361}} & \textbf{\textcolor{red}{0.357}} & \textbf{\textcolor{red}{0.365}} & 0.711 & \textbf{\textcolor{red}{0.750}} & \textbf{\textcolor{red}{0.725}} & 0.711 & \textbf{\textcolor{red}{0.770}} & 0.777 & \textbf{\textcolor{red}{0.860}} & \textbf{\textcolor{red}{0.827}} & \textbf{\textcolor{red}{0.803}} & \textbf{\textcolor{red}{0.807}} & 0.984 & \textbf{\textcolor{red}{0.988}} & \textbf{\textcolor{red}{0.988}} & 0.984 & \textbf{\textcolor{red}{0.990}} & 0.773 & \textbf{\textcolor{red}{0.798}} & \textbf{\textcolor{red}{0.798}} & \textbf{\textcolor{red}{0.832}} & \textbf{\textcolor{red}{0.803}} & 0.932 & \textbf{\textcolor{red}{0.933}} & 0.931 & \textbf{\textcolor{red}{0.934}} & 0.924 \\
& LogicalDeduction & 0.400 & 0.374 & \textbf{\textcolor{red}{0.426}} & 0.400 & \textbf{\textcolor{red}{0.439}} & 0.711 & \textbf{\textcolor{red}{0.735}} & \textbf{\textcolor{red}{0.730}} & \textbf{\textcolor{red}{0.721}} & \textbf{\textcolor{red}{0.725}} & 0.777 & \textbf{\textcolor{red}{0.880}} & \textbf{\textcolor{red}{0.883}} & \textbf{\textcolor{red}{0.890}} & \textbf{\textcolor{red}{0.897}} & 0.984 & 0.984 & 0.982 & \textbf{\textcolor{red}{0.988}} & \textbf{\textcolor{red}{0.992}} & 0.760 & 0.758 & \textbf{\textcolor{red}{0.787}} & \textbf{\textcolor{red}{0.767}} & \textbf{\textcolor{red}{0.788}} & 0.932 & 0.932 & 0.926 & 0.922 & 0.923 \\
& ProntoQA & 0.400 & 0.391 & 0.378 & 0.357 & 0.383 & 0.711 & \textbf{\textcolor{red}{0.770}} & \textbf{\textcolor{red}{0.740}} & \textbf{\textcolor{red}{0.760}} & \textbf{\textcolor{red}{0.755}} & 0.777 & \textbf{\textcolor{red}{0.833}} & \textbf{\textcolor{red}{0.870}} & \textbf{\textcolor{red}{0.870}} & \textbf{\textcolor{red}{0.880}} & 0.982 & \textbf{\textcolor{red}{0.996}} & \textbf{\textcolor{red}{0.992}} & \textbf{\textcolor{red}{0.998}} & \textbf{\textcolor{red}{0.994}} & 0.773 & \textbf{\textcolor{red}{0.813}} & \textbf{\textcolor{red}{0.825}} & \textbf{\textcolor{red}{0.812}} & \textbf{\textcolor{red}{0.812}} & 0.932 & \textbf{\textcolor{red}{0.933}} & 0.925 & 0.918 & 0.923 \\
& ProofWriter & 0.396 & 0.387 & \textbf{\textcolor{red}{0.400}} & 0.378 & 0.374 & 0.711 & \textbf{\textcolor{red}{0.784}} & \textbf{\textcolor{red}{0.750}} & \textbf{\textcolor{red}{0.740}} & \textbf{\textcolor{red}{0.765}} & 0.777 & \textbf{\textcolor{red}{0.873}} & \textbf{\textcolor{red}{0.877}} & \textbf{\textcolor{red}{0.900}} & \textbf{\textcolor{red}{0.890}} & 0.982 & \textbf{\textcolor{red}{0.996}} & \textbf{\textcolor{red}{0.990}} & \textbf{\textcolor{red}{0.992}} & \textbf{\textcolor{red}{0.992}} & 0.773 & \textbf{\textcolor{red}{0.780}} & \textbf{\textcolor{red}{0.808}} & \textbf{\textcolor{red}{0.812}} & \textbf{\textcolor{red}{0.820}} & 0.932 & \textbf{\textcolor{red}{0.936}} & 0.924 & 0.923 & 0.923 \\
& gsm8k & 0.396 & 0.383 & 0.370 & 0.352 & 0.387 & 0.711 & \textbf{\textcolor{red}{0.735}} & 0.706 & \textbf{\textcolor{red}{0.721}} & \textbf{\textcolor{red}{0.716}} & 0.787 & \textbf{\textcolor{red}{0.860}} & \textbf{\textcolor{red}{0.870}} & \textbf{\textcolor{red}{0.860}} & \textbf{\textcolor{red}{0.837}} & 0.982 & 0.970 & 0.972 & 0.974 & 0.974 & 0.773 & 0.753 & 0.767 & 0.768 & 0.760 & 0.932 & 0.929 & 0.841 & 0.804 & 0.803 \\

\bottomrule
\end{tabular}
}%
\caption{Embedding 1}
\end{sidewaystable*}

\begin{sidewaystable*}[t]
\centering
\small
\resizebox{1.05\textwidth}{!}{%
\setlength{\tabcolsep}{3pt}
\begin{tabular}{ll
ccccc
ccccc
ccccc
ccccc
ccccc
ccccc}
\toprule
\textbf{Model} & \textbf{Source}
& \multicolumn{5}{c}{\textbf{AR-LSAT}}
& \multicolumn{5}{c}{\textbf{FOLIO}}
& \multicolumn{5}{c}{\textbf{LogicalDeduction}}
& \multicolumn{5}{c}{\textbf{ProntoQA}}
& \multicolumn{5}{c}{\textbf{ProofWriter}}
& \multicolumn{5}{c}{\textbf{gsm8k}} \\
\cmidrule(lr){3-7}
\cmidrule(lr){8-12}
\cmidrule(lr){13-17}
\cmidrule(lr){18-22}
\cmidrule(lr){23-27}
\cmidrule(lr){28-32}
& & 0 & 1 & 2 & 3 & 4
  & 0 & 1 & 2 & 3 & 4
  & 0 & 1 & 2 & 3 & 4
  & 0 & 1 & 2 & 3 & 4
  & 0 & 1 & 2 & 3 & 4
  & 0 & 1 & 2 & 3 & 4 \\
\midrule
\multirow{5}{*}{qwen3-4} & FOLIO & 0.696 & 0.657 & 0.643 & 0.635 & 0.600 & 0.804 & \textbf{\textcolor{red}{0.809}} & 0.794 & \textbf{\textcolor{red}{0.824}} & \textbf{\textcolor{red}{0.814}} & 0.963 & 0.963 & 0.943 & 0.943 & 0.933 & 0.998 & 0.996 & 0.994 & 0.992 & 0.992 & 0.827 & 0.807 & 0.808 & 0.815 & 0.818 & 0.916 & 0.901 & 0.905 & 0.910 & 0.907 \\
& LogicalDeduction & 0.696 & 0.670 & 0.678 & 0.622 & 0.652 & 0.804 & 0.789 & 0.799 & 0.794 & 0.779 & 0.963 & 0.957 & 0.960 & \textbf{\textcolor{red}{0.967}} & 0.940 & 0.998 & \textbf{\textcolor{red}{1.000}} & \textbf{\textcolor{red}{1.000}} & 0.998 & \textbf{\textcolor{red}{1.000}} & 0.827 & 0.795 & 0.820 & 0.823 & 0.798 & 0.917 & 0.906 & 0.907 & 0.913 & 0.910 \\
& ProntoQA & 0.696 & 0.674 & 0.674 & 0.652 & 0.613 & 0.804 & \textbf{\textcolor{red}{0.809}} & \textbf{\textcolor{red}{0.809}} & 0.799 & 0.789 & 0.963 & 0.940 & 0.950 & 0.933 & 0.950 & 0.998 & 0.990 & 0.996 & 0.992 & 0.992 & 0.827 & 0.797 & 0.793 & 0.788 & 0.803 & 0.917 & 0.910 & 0.900 & 0.889 & 0.895 \\
& ProofWriter & 0.696 & 0.665 & 0.596 & 0.604 & 0.548 & 0.804 & 0.779 & \textbf{\textcolor{red}{0.809}} & 0.804 & 0.804 & 0.963 & 0.930 & 0.937 & 0.910 & 0.923 & 0.998 & 0.992 & 0.994 & \textbf{\textcolor{red}{1.000}} & \textbf{\textcolor{red}{1.000}} & 0.827 & 0.805 & 0.815 & 0.817 & 0.800 & 0.917 & 0.894 & 0.902 & 0.910 & 0.910 \\
& gsm8k & 0.696 & 0.678 & 0.652 & \textbf{\textcolor{red}{0.709}} & 0.661 & 0.804 & 0.779 & \textbf{\textcolor{red}{0.819}} & 0.799 & 0.789 & 0.963 & 0.963 & 0.960 & 0.960 & \textbf{\textcolor{red}{0.967}} & 0.998 & 0.992 & 0.994 & 0.996 & 0.998 & 0.827 & 0.810 & 0.807 & 0.823 & 0.815 & 0.917 & \textbf{\textcolor{red}{0.920}} & 0.912 & \textbf{\textcolor{red}{0.919}} & 0.907 \\
\midrule
\multirow{5}{*}{qwen3-8} & FOLIO & 0.643 & \textbf{\textcolor{red}{0.687}} & \textbf{\textcolor{red}{0.665}} & 0.613 & 0.600 & 0.784 & \textbf{\textcolor{red}{0.809}} & \textbf{\textcolor{red}{0.794}} & \textbf{\textcolor{red}{0.804}} & \textbf{\textcolor{red}{0.833}} & 0.920 & \textbf{\textcolor{red}{0.980}} & \textbf{\textcolor{red}{0.940}} & \textbf{\textcolor{red}{0.960}} & \textbf{\textcolor{red}{0.950}} & 1.000 & 1.000 & 1.000 & 1.000 & 0.996 & 0.882 & 0.858 & 0.845 & 0.865 & 0.850 & 0.929 & 0.920 & 0.917 & 0.914 & 0.914 \\
& LogicalDeduction & 0.643 & \textbf{\textcolor{red}{0.661}} & \textbf{\textcolor{red}{0.674}} & 0.630 & \textbf{\textcolor{red}{0.661}} & 0.784 & \textbf{\textcolor{red}{0.794}} & \textbf{\textcolor{red}{0.804}} & \textbf{\textcolor{red}{0.804}} & 0.775 & 0.920 & \textbf{\textcolor{red}{0.983}} & \textbf{\textcolor{red}{0.980}} & \textbf{\textcolor{red}{0.983}} & \textbf{\textcolor{red}{0.967}} & 1.000 & 1.000 & 0.996 & 0.998 & 1.000 & 0.882 & 0.860 & 0.860 & 0.848 & 0.847 & 0.929 & \textbf{\textcolor{red}{0.932}} & 0.926 & 0.920 & 0.916 \\
& ProntoQA & 0.643 & \textbf{\textcolor{red}{0.657}} & 0.639 & \textbf{\textcolor{red}{0.678}} & 0.613 & 0.784 & \textbf{\textcolor{red}{0.828}} & \textbf{\textcolor{red}{0.814}} & \textbf{\textcolor{red}{0.809}} & \textbf{\textcolor{red}{0.819}} & 0.920 & \textbf{\textcolor{red}{0.950}} & \textbf{\textcolor{red}{0.930}} & \textbf{\textcolor{red}{0.947}} & \textbf{\textcolor{red}{0.947}} & 1.000 & 0.984 & 0.996 & 0.998 & 0.994 & 0.882 & 0.840 & 0.830 & 0.848 & 0.857 & 0.929 & 0.917 & 0.928 & 0.920 & 0.927 \\
& ProofWriter & 0.643 & \textbf{\textcolor{red}{0.652}} & \textbf{\textcolor{red}{0.657}} & 0.600 & 0.561 & 0.784 & \textbf{\textcolor{red}{0.828}} & \textbf{\textcolor{red}{0.824}} & \textbf{\textcolor{red}{0.814}} & \textbf{\textcolor{red}{0.843}} & 0.920 & \textbf{\textcolor{red}{0.960}} & \textbf{\textcolor{red}{0.950}} & \textbf{\textcolor{red}{0.960}} & \textbf{\textcolor{red}{0.950}} & 1.000 & 0.998 & 1.000 & 0.998 & 1.000 & 0.882 & 0.857 & 0.872 & 0.860 & 0.842 & 0.929 & 0.923 & 0.920 & 0.917 & 0.915 \\
& gsm8k & 0.643 & \textbf{\textcolor{red}{0.704}} & \textbf{\textcolor{red}{0.709}} & \textbf{\textcolor{red}{0.661}} & \textbf{\textcolor{red}{0.700}} & 0.784 & \textbf{\textcolor{red}{0.814}} & 0.784 & \textbf{\textcolor{red}{0.804}} & \textbf{\textcolor{red}{0.833}} & 0.920 & \textbf{\textcolor{red}{0.960}} & \textbf{\textcolor{red}{0.940}} & \textbf{\textcolor{red}{0.967}} & \textbf{\textcolor{red}{0.953}} & 1.000 & 0.996 & 0.998 & 0.996 & 1.000 & 0.882 & 0.865 & 0.868 & 0.865 & 0.865 & 0.929 & 0.928 & \textbf{\textcolor{red}{0.934}} & \textbf{\textcolor{red}{0.936}} & \textbf{\textcolor{red}{0.940}} \\
\midrule
\multirow{5}{*}{qwen3-14} & FOLIO & 0.774 & 0.752 & 0.743 & 0.704 & 0.670 & 0.824 & \textbf{\textcolor{red}{0.848}} & 0.824 & 0.814 & 0.814 & 0.990 & \textbf{\textcolor{red}{0.993}} & 0.983 & 0.977 & \textbf{\textcolor{red}{0.993}} & 0.990 & \textbf{\textcolor{red}{0.992}} & \textbf{\textcolor{red}{0.994}} & \textbf{\textcolor{red}{0.996}} & \textbf{\textcolor{red}{0.998}} & 0.840 & \textbf{\textcolor{red}{0.842}} & 0.827 & 0.813 & 0.833 & 0.878 & \textbf{\textcolor{red}{0.915}} & \textbf{\textcolor{red}{0.914}} & \textbf{\textcolor{red}{0.911}} & \textbf{\textcolor{red}{0.913}} \\
& LogicalDeduction & 0.774 & 0.757 & 0.752 & 0.717 & 0.713 & 0.824 & \textbf{\textcolor{red}{0.848}} & \textbf{\textcolor{red}{0.843}} & \textbf{\textcolor{red}{0.828}} & 0.819 & 0.990 & \textbf{\textcolor{red}{0.997}} & 0.990 & 0.990 & 0.980 & 0.990 & 0.990 & \textbf{\textcolor{red}{0.994}} & 0.984 & 0.988 & 0.840 & 0.822 & 0.815 & 0.812 & 0.810 & 0.878 & \textbf{\textcolor{red}{0.914}} & \textbf{\textcolor{red}{0.910}} & \textbf{\textcolor{red}{0.908}} & \textbf{\textcolor{red}{0.908}} \\
& ProntoQA & 0.774 & 0.739 & 0.674 & 0.652 & 0.657 & 0.824 & 0.824 & 0.824 & \textbf{\textcolor{red}{0.848}} & \textbf{\textcolor{red}{0.833}} & 0.990 & 0.990 & \textbf{\textcolor{red}{0.993}} & 0.967 & 0.987 & 0.990 & \textbf{\textcolor{red}{1.000}} & \textbf{\textcolor{red}{0.994}} & \textbf{\textcolor{red}{0.996}} & \textbf{\textcolor{red}{1.000}} & 0.840 & 0.825 & 0.802 & 0.822 & 0.827 & 0.878 & \textbf{\textcolor{red}{0.915}} & \textbf{\textcolor{red}{0.914}} & \textbf{\textcolor{red}{0.913}} & \textbf{\textcolor{red}{0.915}} \\
& ProofWriter & 0.774 & 0.743 & 0.687 & 0.657 & 0.643 & 0.824 & 0.824 & \textbf{\textcolor{red}{0.833}} & 0.824 & 0.824 & 0.990 & 0.983 & 0.983 & 0.980 & \textbf{\textcolor{red}{0.993}} & 0.990 & \textbf{\textcolor{red}{0.992}} & \textbf{\textcolor{red}{0.992}} & \textbf{\textcolor{red}{0.992}} & \textbf{\textcolor{red}{0.998}} & 0.840 & \textbf{\textcolor{red}{0.843}} & 0.825 & \textbf{\textcolor{red}{0.853}} & \textbf{\textcolor{red}{0.857}} & 0.878 & \textbf{\textcolor{red}{0.913}} & \textbf{\textcolor{red}{0.914}} & \textbf{\textcolor{red}{0.912}} & \textbf{\textcolor{red}{0.915}} \\
& gsm8k & 0.774 & 0.752 & 0.757 & 0.757 & 0.739 & 0.824 & 0.809 & 0.824 & 0.809 & 0.814 & 0.990 & 0.990 & 0.987 & 0.980 & \textbf{\textcolor{red}{0.993}} & 0.990 & 0.984 & 0.974 & 0.964 & 0.964 & 0.840 & 0.832 & 0.823 & 0.832 & 0.832 & 0.878 & \textbf{\textcolor{red}{0.917}} & \textbf{\textcolor{red}{0.924}} & \textbf{\textcolor{red}{0.927}} & \textbf{\textcolor{red}{0.930}} \\
\midrule
\multirow{5}{*}{qwen3-32} & FOLIO & 0.791 & 0.783 & 0.717 & 0.739 & 0.687 & 0.804 & 0.789 & \textbf{\textcolor{red}{0.814}} & 0.799 & 0.804 & 0.903 & 0.877 & 0.843 & 0.823 & \textbf{\textcolor{red}{0.913}} & 1.000 & 1.000 & 1.000 & 1.000 & 1.000 & 0.838 & \textbf{\textcolor{red}{0.855}} & \textbf{\textcolor{red}{0.843}} & \textbf{\textcolor{red}{0.853}} & \textbf{\textcolor{red}{0.853}} & 0.930 & \textbf{\textcolor{red}{0.942}} & \textbf{\textcolor{red}{0.938}} & \textbf{\textcolor{red}{0.933}} & \textbf{\textcolor{red}{0.932}} \\
& LogicalDeduction & 0.778 & 0.757 & 0.748 & 0.730 & 0.726 & 0.784 & \textbf{\textcolor{red}{0.789}} & \textbf{\textcolor{red}{0.814}} & \textbf{\textcolor{red}{0.794}} & 0.779 & 0.860 & \textbf{\textcolor{red}{0.930}} & \textbf{\textcolor{red}{0.960}} & \textbf{\textcolor{red}{0.957}} & \textbf{\textcolor{red}{0.943}} & 1.000 & 1.000 & 1.000 & 0.998 & 0.998 & 0.860 & 0.850 & 0.818 & 0.838 & 0.828 & 0.931 & \textbf{\textcolor{red}{0.937}} & 0.930 & \textbf{\textcolor{red}{0.933}} & \textbf{\textcolor{red}{0.936}} \\
& ProntoQA & 0.770 & 0.717 & 0.730 & 0.726 & 0.735 & 0.775 & \textbf{\textcolor{red}{0.804}} & \textbf{\textcolor{red}{0.794}} & \textbf{\textcolor{red}{0.794}} & \textbf{\textcolor{red}{0.809}} & 0.907 & 0.903 & 0.833 & 0.827 & 0.723 & 1.000 & 1.000 & 0.996 & 0.994 & 0.992 & 0.843 & 0.842 & 0.833 & 0.838 & \textbf{\textcolor{red}{0.845}} & 0.929 & \textbf{\textcolor{red}{0.936}} & \textbf{\textcolor{red}{0.939}} & \textbf{\textcolor{red}{0.929}} & \textbf{\textcolor{red}{0.932}} \\
& ProofWriter & 0.752 & 0.700 & -- & -- & -- & 0.799 & 0.770 & 0.775 & \textbf{\textcolor{red}{0.804}} & \textbf{\textcolor{red}{0.814}} & 0.880 & \textbf{\textcolor{red}{0.920}} & \textbf{\textcolor{red}{0.920}} & 0.757 & 0.750 & 1.000 & 1.000 & 1.000 & 1.000 & 1.000 & 0.827 & \textbf{\textcolor{red}{0.877}} & \textbf{\textcolor{red}{0.860}} & \textbf{\textcolor{red}{0.842}} & \textbf{\textcolor{red}{0.863}} & 0.929 & \textbf{\textcolor{red}{0.937}} & \textbf{\textcolor{red}{0.936}} & \textbf{\textcolor{red}{0.937}} & \textbf{\textcolor{red}{0.937}} \\
& gsm8k & 0.757 & 0.735 & 0.748 & 0.739 & 0.748 & 0.775 & 0.770 & \textbf{\textcolor{red}{0.799}} & \textbf{\textcolor{red}{0.789}} & \textbf{\textcolor{red}{0.789}} & 0.887 & 0.770 & 0.730 & 0.647 & 0.650 & 1.000 & 1.000 & 1.000 & 1.000 & 1.000 & 0.843 & 0.833 & 0.833 & \textbf{\textcolor{red}{0.855}} & 0.837 & 0.936 & \textbf{\textcolor{red}{0.944}} & \textbf{\textcolor{red}{0.945}} & \textbf{\textcolor{red}{0.942}} & \textbf{\textcolor{red}{0.942}} \\
\midrule
\multirow{5}{*}{llama3-8} & FOLIO & 0.287 & 0.230 & 0.200 & 0.252 & 0.235 & 0.593 & \textbf{\textcolor{red}{0.608}} & \textbf{\textcolor{red}{0.598}} & 0.574 & \textbf{\textcolor{red}{0.627}} & 0.547 & \textbf{\textcolor{red}{0.557}} & 0.527 & 0.490 & 0.480 & 0.794 & 0.782 & \textbf{\textcolor{red}{0.798}} & 0.794 & \textbf{\textcolor{red}{0.800}} & 0.448 & \textbf{\textcolor{red}{0.528}} & \textbf{\textcolor{red}{0.537}} & \textbf{\textcolor{red}{0.518}} & \textbf{\textcolor{red}{0.552}} & 0.616 & 0.588 & 0.576 & 0.561 & 0.550 \\
& LogicalDeduction & 0.287 & 0.278 & 0.143 & 0.191 & 0.252 & 0.593 & \textbf{\textcolor{red}{0.608}} & 0.593 & 0.574 & \textbf{\textcolor{red}{0.608}} & 0.547 & \textbf{\textcolor{red}{0.597}} & \textbf{\textcolor{red}{0.580}} & \textbf{\textcolor{red}{0.580}} & 0.523 & 0.794 & \textbf{\textcolor{red}{0.840}} & \textbf{\textcolor{red}{0.846}} & \textbf{\textcolor{red}{0.814}} & \textbf{\textcolor{red}{0.850}} & 0.448 & \textbf{\textcolor{red}{0.493}} & \textbf{\textcolor{red}{0.527}} & \textbf{\textcolor{red}{0.520}} & \textbf{\textcolor{red}{0.497}} & 0.616 & 0.597 & 0.584 & 0.580 & 0.582 \\
& ProntoQA & 0.287 & 0.217 & 0.209 & 0.209 & 0.204 & 0.593 & 0.569 & \textbf{\textcolor{red}{0.603}} & \textbf{\textcolor{red}{0.623}} & \textbf{\textcolor{red}{0.623}} & 0.547 & 0.467 & 0.463 & 0.530 & \textbf{\textcolor{red}{0.550}} & 0.794 & \textbf{\textcolor{red}{0.860}} & \textbf{\textcolor{red}{0.892}} & \textbf{\textcolor{red}{0.920}} & \textbf{\textcolor{red}{0.942}} & 0.448 & \textbf{\textcolor{red}{0.568}} & \textbf{\textcolor{red}{0.563}} & \textbf{\textcolor{red}{0.543}} & \textbf{\textcolor{red}{0.570}} & 0.616 & 0.598 & 0.597 & 0.590 & 0.594 \\
& ProofWriter & 0.287 & 0.257 & 0.209 & 0.243 & 0.248 & 0.593 & \textbf{\textcolor{red}{0.618}} & \textbf{\textcolor{red}{0.618}} & 0.574 & \textbf{\textcolor{red}{0.598}} & 0.547 & 0.457 & 0.477 & 0.507 & 0.497 & 0.794 & 0.736 & 0.774 & 0.780 & 0.794 & 0.448 & \textbf{\textcolor{red}{0.538}} & \textbf{\textcolor{red}{0.597}} & \textbf{\textcolor{red}{0.598}} & \textbf{\textcolor{red}{0.560}} & 0.616 & 0.582 & 0.587 & 0.558 & 0.553 \\
& gsm8k & 0.287 & 0.230 & 0.222 & 0.204 & 0.222 & 0.593 & 0.569 & 0.544 & 0.520 & 0.515 & 0.547 & 0.490 & 0.463 & 0.433 & 0.463 & 0.794 & 0.780 & 0.782 & 0.784 & 0.780 & 0.448 & \textbf{\textcolor{red}{0.498}} & \textbf{\textcolor{red}{0.452}} & 0.447 & \textbf{\textcolor{red}{0.490}} & 0.616 & \textbf{\textcolor{red}{0.660}} & \textbf{\textcolor{red}{0.688}} & \textbf{\textcolor{red}{0.707}} & \textbf{\textcolor{red}{0.704}} \\

\bottomrule
\end{tabular}
}%
\caption{Embedding 2}
\label{Embedding_2}
\end{sidewaystable*}

\begin{sidewaystable*}[t]
\centering
\small
\resizebox{1.05\textwidth}{!}{%
\setlength{\tabcolsep}{3pt}
\begin{tabular}{ll
ccccc
ccccc
ccccc
ccccc
ccccc
ccccc}
\toprule
\textbf{Model} & \textbf{Source}
& \multicolumn{5}{c}{\textbf{AR-LSAT}}
& \multicolumn{5}{c}{\textbf{FOLIO}}
& \multicolumn{5}{c}{\textbf{LogicalDeduction}}
& \multicolumn{5}{c}{\textbf{ProntoQA}}
& \multicolumn{5}{c}{\textbf{ProofWriter}}
& \multicolumn{5}{c}{\textbf{gsm8k}} \\
\cmidrule(lr){3-7}
\cmidrule(lr){8-12}
\cmidrule(lr){13-17}
\cmidrule(lr){18-22}
\cmidrule(lr){23-27}
\cmidrule(lr){28-32}
& & 0 & 1 & 2 & 3 & 4
  & 0 & 1 & 2 & 3 & 4
  & 0 & 1 & 2 & 3 & 4
  & 0 & 1 & 2 & 3 & 4
  & 0 & 1 & 2 & 3 & 4
  & 0 & 1 & 2 & 3 & 4 \\
\midrule
\multirow{5}{*}{qwen3} & FOLIO & 0.257 & 0.196 & \textbf{\textcolor{red}{0.283}} & 0.217 & 0.200 & 0.632 & 0.632 & 0.623 & \textbf{\textcolor{red}{0.637}} & \textbf{\textcolor{red}{0.637}} & 0.427 & \textbf{\textcolor{red}{0.440}} & \textbf{\textcolor{red}{0.463}} & \textbf{\textcolor{red}{0.463}} & \textbf{\textcolor{red}{0.490}} & 0.868 & 0.828 & 0.830 & 0.846 & 0.844 & 0.530 & 0.518 & \textbf{\textcolor{red}{0.553}} & \textbf{\textcolor{red}{0.558}} & \textbf{\textcolor{red}{0.543}} & 0.737 & 0.726 & 0.675 & 0.644 & 0.637 \\
& LogicalDeduction & 0.235 & 0.217 & 0.235 & 0.209 & \textbf{\textcolor{red}{0.243}} & 0.637 & 0.598 & 0.637 & 0.598 & 0.613 & 0.427 & \textbf{\textcolor{red}{0.453}} & \textbf{\textcolor{red}{0.477}} & \textbf{\textcolor{red}{0.497}} & \textbf{\textcolor{red}{0.467}} & 0.850 & 0.842 & 0.838 & 0.814 & 0.844 & 0.538 & \textbf{\textcolor{red}{0.543}} & \textbf{\textcolor{red}{0.550}} & \textbf{\textcolor{red}{0.567}} & \textbf{\textcolor{red}{0.553}} & 0.729 & \textbf{\textcolor{red}{0.732}} & 0.712 & 0.692 & 0.677 \\
& ProntoQA & 0.248 & 0.165 & 0.157 & 0.174 & 0.178 & 0.627 & 0.510 & 0.495 & 0.461 & 0.426 & 0.427 & \textbf{\textcolor{red}{0.467}} & \textbf{\textcolor{red}{0.433}} & \textbf{\textcolor{red}{0.437}} & \textbf{\textcolor{red}{0.443}} & 0.854 & 0.742 & 0.758 & 0.768 & 0.774 & 0.533 & 0.410 & 0.322 & 0.305 & 0.308 & 0.737 & 0.702 & 0.718 & \textbf{\textcolor{red}{0.738}} & 0.734 \\
& ProofWriter & 0.257 & 0.209 & 0.204 & 0.235 & 0.213 & 0.632 & 0.632 & \textbf{\textcolor{red}{0.672}} & 0.613 & 0.578 & 0.427 & \textbf{\textcolor{red}{0.460}} & 0.423 & \textbf{\textcolor{red}{0.483}} & \textbf{\textcolor{red}{0.457}} & 0.868 & 0.864 & 0.848 & 0.848 & 0.864 & 0.530 & 0.517 & \textbf{\textcolor{red}{0.552}} & \textbf{\textcolor{red}{0.563}} & 0.527 & 0.737 & 0.719 & 0.684 & 0.623 & 0.602 \\
& gsm8k & 0.257 & 0.226 & 0.196 & 0.222 & 0.157 & 0.632 & 0.613 & 0.632 & 0.544 & 0.564 & 0.427 & \textbf{\textcolor{red}{0.467}} & \textbf{\textcolor{red}{0.433}} & \textbf{\textcolor{red}{0.437}} & \textbf{\textcolor{red}{0.440}} & 0.868 & 0.784 & 0.708 & 0.600 & 0.556 & 0.530 & 0.512 & 0.465 & 0.413 & 0.448 & 0.737 & 0.728 & \textbf{\textcolor{red}{0.757}} & 0.710 & 0.650 \\
\midrule
\multirow{5}{*}{qwen7} & FOLIO & 0.252 & \textbf{\textcolor{red}{0.287}} & 0.226 & 0.222 & 0.235 & 0.701 & 0.696 & 0.676 & \textbf{\textcolor{red}{0.745}} & \textbf{\textcolor{red}{0.735}} & 0.627 & 0.610 & 0.623 & \textbf{\textcolor{red}{0.657}} & \textbf{\textcolor{red}{0.637}} & 0.968 & 0.968 & \textbf{\textcolor{red}{0.970}} & \textbf{\textcolor{red}{0.976}} & \textbf{\textcolor{red}{0.988}} & 0.623 & \textbf{\textcolor{red}{0.637}} & 0.617 & 0.613 & 0.618 & 0.887 & 0.871 & 0.880 & 0.856 & 0.829 \\
& LogicalDeduction & 0.252 & \textbf{\textcolor{red}{0.257}} & 0.252 & \textbf{\textcolor{red}{0.291}} & \textbf{\textcolor{red}{0.278}} & 0.691 & 0.691 & 0.676 & 0.686 & 0.686 & 0.627 & 0.573 & 0.580 & 0.623 & 0.587 & 0.970 & \textbf{\textcolor{red}{0.972}} & 0.966 & \textbf{\textcolor{red}{0.986}} & \textbf{\textcolor{red}{0.986}} & 0.623 & 0.612 & 0.612 & 0.618 & \textbf{\textcolor{red}{0.637}} & 0.889 & 0.881 & 0.880 & 0.878 & 0.867 \\
& ProntoQA & 0.270 & 0.252 & 0.204 & \textbf{\textcolor{red}{0.274}} & 0.235 & 0.696 & 0.642 & 0.608 & 0.613 & 0.608 & 0.627 & 0.620 & \textbf{\textcolor{red}{0.630}} & 0.627 & 0.617 & 0.970 & 0.828 & 0.842 & 0.834 & 0.852 & 0.620 & 0.543 & 0.500 & 0.455 & 0.490 & 0.888 & 0.848 & 0.880 & 0.875 & 0.877 \\
& ProofWriter & 0.252 & 0.243 & 0.222 & \textbf{\textcolor{red}{0.265}} & \textbf{\textcolor{red}{0.278}} & 0.701 & \textbf{\textcolor{red}{0.711}} & \textbf{\textcolor{red}{0.716}} & \textbf{\textcolor{red}{0.725}} & 0.696 & 0.627 & 0.607 & \textbf{\textcolor{red}{0.657}} & 0.627 & 0.620 & 0.968 & 0.964 & \textbf{\textcolor{red}{0.986}} & \textbf{\textcolor{red}{0.984}} & \textbf{\textcolor{red}{0.976}} & 0.623 & \textbf{\textcolor{red}{0.662}} & \textbf{\textcolor{red}{0.635}} & \textbf{\textcolor{red}{0.668}} & \textbf{\textcolor{red}{0.665}} & 0.887 & 0.860 & 0.855 & 0.852 & 0.783 \\
& gsm8k & 0.252 & 0.235 & 0.239 & 0.239 & 0.248 & 0.701 & 0.647 & 0.637 & 0.632 & 0.627 & 0.627 & 0.607 & 0.600 & 0.610 & 0.610 & 0.968 & 0.934 & 0.924 & 0.916 & 0.914 & 0.623 & 0.603 & 0.612 & \textbf{\textcolor{red}{0.648}} & \textbf{\textcolor{red}{0.648}} & 0.887 & 0.882 & 0.884 & 0.884 & 0.863 \\
\midrule
\multirow{5}{*}{qwen14} & FOLIO & 0.317 & 0.313 & 0.304 & 0.300 & \textbf{\textcolor{red}{0.348}} & 0.681 & \textbf{\textcolor{red}{0.755}} & \textbf{\textcolor{red}{0.716}} & \textbf{\textcolor{red}{0.730}} & \textbf{\textcolor{red}{0.730}} & 0.823 & 0.783 & 0.780 & 0.803 & 0.800 & 0.946 & \textbf{\textcolor{red}{0.968}} & \textbf{\textcolor{red}{0.972}} & \textbf{\textcolor{red}{0.966}} & \textbf{\textcolor{red}{0.984}} & 0.693 & 0.680 & \textbf{\textcolor{red}{0.705}} & \textbf{\textcolor{red}{0.727}} & \textbf{\textcolor{red}{0.708}} & 0.932 & 0.927 & 0.932 & 0.921 & 0.923 \\
& LogicalDeduction & 0.317 & 0.296 & \textbf{\textcolor{red}{0.352}} & \textbf{\textcolor{red}{0.322}} & 0.278 & 0.681 & \textbf{\textcolor{red}{0.755}} & \textbf{\textcolor{red}{0.770}} & \textbf{\textcolor{red}{0.740}} & \textbf{\textcolor{red}{0.725}} & 0.823 & 0.740 & 0.750 & 0.780 & 0.783 & 0.946 & 0.944 & \textbf{\textcolor{red}{0.992}} & \textbf{\textcolor{red}{0.974}} & \textbf{\textcolor{red}{0.974}} & 0.693 & \textbf{\textcolor{red}{0.695}} & \textbf{\textcolor{red}{0.702}} & \textbf{\textcolor{red}{0.713}} & \textbf{\textcolor{red}{0.708}} & 0.932 & 0.930 & 0.929 & 0.927 & \textbf{\textcolor{red}{0.936}} \\
& ProntoQA & -- & -- & -- & -- & -- & -- & -- & -- & -- & -- & 0.823 & 0.770 & 0.750 & 0.770 & 0.757 & -- & -- & -- & -- & -- & -- & -- & -- & -- & -- & -- & -- & -- & -- & -- \\
& ProofWriter & 0.317 & \textbf{\textcolor{red}{0.322}} & 0.309 & \textbf{\textcolor{red}{0.374}} & 0.317 & 0.681 & \textbf{\textcolor{red}{0.770}} & \textbf{\textcolor{red}{0.784}} & \textbf{\textcolor{red}{0.750}} & \textbf{\textcolor{red}{0.740}} & 0.823 & 0.737 & 0.787 & 0.797 & 0.773 & 0.946 & \textbf{\textcolor{red}{0.984}} & \textbf{\textcolor{red}{0.996}} & \textbf{\textcolor{red}{0.984}} & \textbf{\textcolor{red}{0.984}} & 0.693 & \textbf{\textcolor{red}{0.752}} & \textbf{\textcolor{red}{0.763}} & \textbf{\textcolor{red}{0.743}} & \textbf{\textcolor{red}{0.773}} & 0.932 & 0.927 & 0.928 & \textbf{\textcolor{red}{0.937}} & \textbf{\textcolor{red}{0.935}} \\
& gsm8k & 0.317 & 0.313 & \textbf{\textcolor{red}{0.322}} & \textbf{\textcolor{red}{0.326}} & 0.313 & 0.681 & \textbf{\textcolor{red}{0.711}} & \textbf{\textcolor{red}{0.711}} & \textbf{\textcolor{red}{0.701}} & \textbf{\textcolor{red}{0.735}} & 0.823 & 0.803 & 0.757 & 0.790 & 0.753 & 0.946 & 0.928 & 0.914 & 0.896 & 0.902 & 0.693 & 0.627 & 0.612 & 0.605 & 0.608 & 0.932 & 0.932 & 0.923 & 0.925 & 0.919 \\
\midrule
\multirow{5}{*}{qwen32} & FOLIO & 0.326 & \textbf{\textcolor{red}{0.343}} & 0.313 & 0.309 & \textbf{\textcolor{red}{0.339}} & 0.706 & \textbf{\textcolor{red}{0.725}} & \textbf{\textcolor{red}{0.740}} & \textbf{\textcolor{red}{0.770}} & \textbf{\textcolor{red}{0.735}} & 0.880 & 0.837 & 0.857 & 0.850 & 0.820 & 1.000 & 0.998 & 1.000 & 1.000 & 1.000 & 0.707 & \textbf{\textcolor{red}{0.713}} & 0.707 & \textbf{\textcolor{red}{0.725}} & \textbf{\textcolor{red}{0.720}} & 0.882 & \textbf{\textcolor{red}{0.898}} & \textbf{\textcolor{red}{0.897}} & \textbf{\textcolor{red}{0.894}} & \textbf{\textcolor{red}{0.892}} \\
& LogicalDeduction & 0.326 & 0.322 & 0.296 & 0.300 & \textbf{\textcolor{red}{0.343}} & 0.706 & \textbf{\textcolor{red}{0.721}} & \textbf{\textcolor{red}{0.716}} & \textbf{\textcolor{red}{0.711}} & \textbf{\textcolor{red}{0.740}} & 0.880 & 0.843 & 0.833 & 0.873 & 0.863 & 1.000 & 1.000 & 1.000 & 1.000 & 0.998 & 0.707 & 0.702 & \textbf{\textcolor{red}{0.740}} & \textbf{\textcolor{red}{0.730}} & \textbf{\textcolor{red}{0.732}} & 0.882 & \textbf{\textcolor{red}{0.904}} & \textbf{\textcolor{red}{0.917}} & \textbf{\textcolor{red}{0.909}} & \textbf{\textcolor{red}{0.914}} \\
& ProntoQA & 0.326 & \textbf{\textcolor{red}{0.343}} & 0.317 & \textbf{\textcolor{red}{0.361}} & \textbf{\textcolor{red}{0.400}} & 0.706 & 0.696 & \textbf{\textcolor{red}{0.716}} & \textbf{\textcolor{red}{0.725}} & 0.691 & 0.880 & 0.843 & 0.800 & 0.780 & 0.803 & 1.000 & 0.978 & 0.990 & 0.994 & 0.996 & 0.707 & 0.660 & 0.677 & 0.680 & 0.642 & 0.882 & \textbf{\textcolor{red}{0.892}} & \textbf{\textcolor{red}{0.905}} & \textbf{\textcolor{red}{0.907}} & \textbf{\textcolor{red}{0.921}} \\
& ProofWriter & 0.326 & \textbf{\textcolor{red}{0.357}} & \textbf{\textcolor{red}{0.357}} & \textbf{\textcolor{red}{0.374}} & 0.317 & 0.706 & \textbf{\textcolor{red}{0.725}} & \textbf{\textcolor{red}{0.745}} & \textbf{\textcolor{red}{0.716}} & \textbf{\textcolor{red}{0.735}} & 0.880 & 0.877 & 0.867 & 0.873 & \textbf{\textcolor{red}{0.883}} & 1.000 & 0.998 & 1.000 & 1.000 & 1.000 & 0.707 & 0.707 & \textbf{\textcolor{red}{0.735}} & \textbf{\textcolor{red}{0.767}} & \textbf{\textcolor{red}{0.768}} & 0.882 & \textbf{\textcolor{red}{0.889}} & \textbf{\textcolor{red}{0.901}} & 0.870 & 0.843 \\
& gsm8k & 0.326 & \textbf{\textcolor{red}{0.357}} & 0.317 & \textbf{\textcolor{red}{0.352}} & 0.326 & 0.706 & \textbf{\textcolor{red}{0.716}} & 0.696 & 0.696 & 0.706 & -- & -- & -- & -- & -- & 1.000 & 0.994 & 0.986 & 0.994 & 0.984 & 0.707 & 0.613 & 0.592 & 0.595 & 0.597 & 0.882 & \textbf{\textcolor{red}{0.915}} & \textbf{\textcolor{red}{0.922}} & \textbf{\textcolor{red}{0.924}} & \textbf{\textcolor{red}{0.919}} \\
\midrule
\multirow{5}{*}{gemma3-4} & FOLIO & 0.261 & 0.248 & 0.230 & 0.235 & 0.213 & 0.598 & \textbf{\textcolor{red}{0.603}} & \textbf{\textcolor{red}{0.603}} & 0.593 & \textbf{\textcolor{red}{0.686}} & 0.577 & 0.410 & 0.410 & 0.397 & 0.413 & 0.918 & 0.862 & 0.916 & \textbf{\textcolor{red}{0.932}} & \textbf{\textcolor{red}{0.922}} & 0.643 & 0.615 & 0.615 & 0.618 & 0.630 & 0.862 & 0.836 & 0.840 & 0.836 & 0.831 \\
& LogicalDeduction & 0.261 & \textbf{\textcolor{red}{0.270}} & 0.235 & 0.257 & 0.235 & 0.627 & \textbf{\textcolor{red}{0.637}} & \textbf{\textcolor{red}{0.637}} & \textbf{\textcolor{red}{0.637}} & 0.613 & 0.590 & 0.357 & 0.370 & 0.423 & 0.387 & 0.912 & 0.590 & 0.638 & 0.654 & 0.644 & 0.635 & 0.570 & 0.608 & 0.545 & 0.557 & 0.863 & 0.841 & 0.830 & 0.838 & 0.836 \\
& ProntoQA & -- & -- & -- & -- & -- & 0.623 & 0.559 & 0.608 & 0.598 & 0.593 & 0.600 & 0.323 & 0.340 & 0.337 & 0.367 & 0.904 & 0.846 & 0.822 & 0.806 & 0.806 & 0.628 & \textbf{\textcolor{red}{0.640}} & \textbf{\textcolor{red}{0.662}} & 0.603 & 0.505 & -- & -- & -- & -- & -- \\
& ProofWriter & 0.252 & 0.222 & \textbf{\textcolor{red}{0.278}} & \textbf{\textcolor{red}{0.287}} & \textbf{\textcolor{red}{0.274}} & 0.613 & 0.608 & \textbf{\textcolor{red}{0.676}} & \textbf{\textcolor{red}{0.667}} & \textbf{\textcolor{red}{0.672}} & 0.563 & 0.397 & 0.383 & 0.413 & 0.400 & 0.916 & \textbf{\textcolor{red}{0.944}} & \textbf{\textcolor{red}{0.936}} & \textbf{\textcolor{red}{0.952}} & \textbf{\textcolor{red}{0.926}} & 0.637 & \textbf{\textcolor{red}{0.680}} & \textbf{\textcolor{red}{0.648}} & \textbf{\textcolor{red}{0.653}} & \textbf{\textcolor{red}{0.657}} & 0.863 & 0.854 & 0.849 & 0.857 & 0.846 \\
& gsm8k & 0.257 & \textbf{\textcolor{red}{0.261}} & 0.222 & 0.257 & 0.235 & 0.613 & \textbf{\textcolor{red}{0.627}} & \textbf{\textcolor{red}{0.623}} & 0.588 & 0.588 & 0.587 & \textbf{\textcolor{red}{0.627}} & 0.487 & 0.397 & 0.447 & 0.924 & 0.866 & 0.840 & 0.810 & 0.826 & 0.628 & 0.593 & 0.605 & 0.535 & 0.590 & 0.861 & 0.851 & 0.824 & 0.801 & 0.745 \\
\midrule
\multirow{5}{*}{gemma3-12} & FOLIO & 0.313 & \textbf{\textcolor{red}{0.326}} & \textbf{\textcolor{red}{0.352}} & \textbf{\textcolor{red}{0.317}} & 0.283 & 0.701 & \textbf{\textcolor{red}{0.721}} & \textbf{\textcolor{red}{0.745}} & \textbf{\textcolor{red}{0.706}} & \textbf{\textcolor{red}{0.711}} & 0.777 & 0.667 & 0.643 & 0.653 & 0.623 & 0.986 & 0.984 & 0.978 & \textbf{\textcolor{red}{0.988}} & \textbf{\textcolor{red}{0.994}} & 0.735 & \textbf{\textcolor{red}{0.745}} & \textbf{\textcolor{red}{0.775}} & \textbf{\textcolor{red}{0.773}} & \textbf{\textcolor{red}{0.772}} & 0.918 & 0.911 & 0.900 & 0.892 & 0.887 \\
& LogicalDeduction & 0.313 & \textbf{\textcolor{red}{0.326}} & \textbf{\textcolor{red}{0.326}} & 0.287 & 0.257 & 0.701 & \textbf{\textcolor{red}{0.706}} & \textbf{\textcolor{red}{0.706}} & \textbf{\textcolor{red}{0.750}} & \textbf{\textcolor{red}{0.711}} & 0.777 & 0.653 & 0.667 & 0.723 & 0.700 & 0.986 & 0.938 & 0.942 & 0.938 & 0.928 & 0.737 & 0.732 & \textbf{\textcolor{red}{0.757}} & \textbf{\textcolor{red}{0.747}} & \textbf{\textcolor{red}{0.745}} & 0.919 & \textbf{\textcolor{red}{0.920}} & 0.909 & 0.891 & 0.898 \\
& ProntoQA & 0.313 & 0.300 & 0.296 & 0.309 & \textbf{\textcolor{red}{0.317}} & 0.701 & 0.691 & \textbf{\textcolor{red}{0.716}} & 0.647 & \textbf{\textcolor{red}{0.711}} & 0.757 & 0.577 & 0.623 & 0.593 & 0.640 & 0.986 & 0.910 & 0.968 & 0.956 & 0.878 & 0.737 & 0.648 & 0.605 & 0.633 & 0.572 & 0.918 & 0.912 & 0.908 & 0.913 & 0.903 \\
& ProofWriter & 0.313 & \textbf{\textcolor{red}{0.326}} & 0.274 & 0.304 & 0.304 & 0.706 & \textbf{\textcolor{red}{0.760}} & \textbf{\textcolor{red}{0.740}} & \textbf{\textcolor{red}{0.789}} & \textbf{\textcolor{red}{0.755}} & 0.773 & 0.653 & 0.670 & 0.637 & 0.630 & 0.986 & \textbf{\textcolor{red}{0.992}} & \textbf{\textcolor{red}{0.996}} & \textbf{\textcolor{red}{0.996}} & \textbf{\textcolor{red}{0.988}} & 0.738 & \textbf{\textcolor{red}{0.757}} & \textbf{\textcolor{red}{0.755}} & \textbf{\textcolor{red}{0.767}} & \textbf{\textcolor{red}{0.773}} & 0.920 & 0.910 & 0.901 & 0.894 & 0.889 \\
& gsm8k & 0.296 & \textbf{\textcolor{red}{0.339}} & \textbf{\textcolor{red}{0.304}} & 0.274 & \textbf{\textcolor{red}{0.322}} & 0.706 & \textbf{\textcolor{red}{0.711}} & 0.696 & \textbf{\textcolor{red}{0.721}} & 0.701 & 0.773 & \textbf{\textcolor{red}{0.777}} & 0.620 & 0.690 & 0.640 & 0.986 & 0.978 & 0.982 & 0.982 & \textbf{\textcolor{red}{0.988}} & 0.738 & 0.732 & \textbf{\textcolor{red}{0.753}} & \textbf{\textcolor{red}{0.745}} & 0.738 & 0.920 & 0.868 & 0.631 & 0.565 & 0.544 \\
\midrule
\multirow{5}{*}{gemma3-27} & FOLIO & 0.330 & \textbf{\textcolor{red}{0.365}} & \textbf{\textcolor{red}{0.391}} & \textbf{\textcolor{red}{0.387}} & 0.326 & 0.721 & \textbf{\textcolor{red}{0.765}} & \textbf{\textcolor{red}{0.779}} & 0.721 & \textbf{\textcolor{red}{0.804}} & 0.827 & \textbf{\textcolor{red}{0.867}} & \textbf{\textcolor{red}{0.857}} & \textbf{\textcolor{red}{0.843}} & \textbf{\textcolor{red}{0.867}} & 0.984 & \textbf{\textcolor{red}{0.992}} & \textbf{\textcolor{red}{0.986}} & \textbf{\textcolor{red}{0.988}} & \textbf{\textcolor{red}{0.990}} & 0.763 & \textbf{\textcolor{red}{0.798}} & \textbf{\textcolor{red}{0.802}} & \textbf{\textcolor{red}{0.782}} & \textbf{\textcolor{red}{0.813}} & 0.936 & \textbf{\textcolor{red}{0.937}} & 0.932 & 0.936 & 0.930 \\
& LogicalDeduction & 0.335 & \textbf{\textcolor{red}{0.417}} & \textbf{\textcolor{red}{0.383}} & \textbf{\textcolor{red}{0.400}} & \textbf{\textcolor{red}{0.370}} & 0.725 & 0.716 & 0.711 & 0.716 & \textbf{\textcolor{red}{0.745}} & 0.833 & 0.793 & 0.790 & \textbf{\textcolor{red}{0.840}} & \textbf{\textcolor{red}{0.837}} & 0.984 & \textbf{\textcolor{red}{0.988}} & \textbf{\textcolor{red}{0.992}} & 0.982 & 0.982 & 0.767 & \textbf{\textcolor{red}{0.777}} & \textbf{\textcolor{red}{0.768}} & 0.763 & \textbf{\textcolor{red}{0.785}} & 0.936 & 0.933 & 0.934 & 0.930 & 0.927 \\
& ProntoQA & 0.330 & \textbf{\textcolor{red}{0.383}} & \textbf{\textcolor{red}{0.352}} & \textbf{\textcolor{red}{0.348}} & \textbf{\textcolor{red}{0.387}} & 0.725 & \textbf{\textcolor{red}{0.750}} & \textbf{\textcolor{red}{0.740}} & 0.716 & \textbf{\textcolor{red}{0.730}} & 0.830 & \textbf{\textcolor{red}{0.840}} & 0.827 & \textbf{\textcolor{red}{0.840}} & 0.830 & 0.986 & \textbf{\textcolor{red}{0.992}} & \textbf{\textcolor{red}{0.988}} & \textbf{\textcolor{red}{0.992}} & \textbf{\textcolor{red}{0.994}} & 0.763 & \textbf{\textcolor{red}{0.773}} & 0.740 & 0.748 & 0.752 & 0.936 & 0.922 & 0.929 & 0.931 & 0.925 \\
& ProofWriter & 0.348 & \textbf{\textcolor{red}{0.361}} & \textbf{\textcolor{red}{0.370}} & \textbf{\textcolor{red}{0.404}} & \textbf{\textcolor{red}{0.378}} & 0.730 & \textbf{\textcolor{red}{0.760}} & \textbf{\textcolor{red}{0.760}} & \textbf{\textcolor{red}{0.760}} & \textbf{\textcolor{red}{0.765}} & 0.833 & \textbf{\textcolor{red}{0.870}} & \textbf{\textcolor{red}{0.843}} & \textbf{\textcolor{red}{0.863}} & \textbf{\textcolor{red}{0.853}} & 0.986 & \textbf{\textcolor{red}{1.000}} & \textbf{\textcolor{red}{0.994}} & \textbf{\textcolor{red}{0.994}} & \textbf{\textcolor{red}{0.990}} & 0.770 & \textbf{\textcolor{red}{0.785}} & \textbf{\textcolor{red}{0.802}} & \textbf{\textcolor{red}{0.790}} & \textbf{\textcolor{red}{0.787}} & 0.936 & 0.929 & 0.936 & 0.926 & 0.929 \\
& gsm8k & 0.330 & \textbf{\textcolor{red}{0.387}} & \textbf{\textcolor{red}{0.348}} & 0.330 & \textbf{\textcolor{red}{0.335}} & 0.725 & 0.701 & 0.721 & 0.725 & 0.721 & 0.830 & \textbf{\textcolor{red}{0.857}} & \textbf{\textcolor{red}{0.837}} & 0.807 & 0.807 & 0.986 & 0.980 & 0.978 & 0.978 & 0.982 & 0.768 & 0.760 & \textbf{\textcolor{red}{0.800}} & \textbf{\textcolor{red}{0.773}} & 0.757 & 0.936 & 0.918 & 0.810 & 0.743 & 0.733 \\

\bottomrule
\end{tabular}
}%
\caption{RAG\_bm25 cross-domain EM of Qwen models. Red numbers indicate improvements over 0-shot.}
\end{sidewaystable*}

\begin{sidewaystable*}[t]
\centering
\small
\resizebox{1.05\textwidth}{!}{%
\setlength{\tabcolsep}{3pt}
\begin{tabular}{ll
ccccc
ccccc
ccccc
ccccc
ccccc
ccccc}
\toprule
\textbf{Model} & \textbf{Source}
& \multicolumn{5}{c}{\textbf{AR-LSAT}}
& \multicolumn{5}{c}{\textbf{FOLIO}}
& \multicolumn{5}{c}{\textbf{LogicalDeduction}}
& \multicolumn{5}{c}{\textbf{ProntoQA}}
& \multicolumn{5}{c}{\textbf{ProofWriter}}
& \multicolumn{5}{c}{\textbf{gsm8k}} \\
\cmidrule(lr){3-7}
\cmidrule(lr){8-12}
\cmidrule(lr){13-17}
\cmidrule(lr){18-22}
\cmidrule(lr){23-27}
\cmidrule(lr){28-32}
& & 0 & 1 & 2 & 3 & 4
  & 0 & 1 & 2 & 3 & 4
  & 0 & 1 & 2 & 3 & 4
  & 0 & 1 & 2 & 3 & 4
  & 0 & 1 & 2 & 3 & 4
  & 0 & 1 & 2 & 3 & 4 \\
\midrule
\multirow{5}{*}{qwen3-4} & FOLIO & 0.726 & 0.665 & 0.674 & 0.678 & 0.678 & 0.804 & \textbf{\textcolor{red}{0.819}} & \textbf{\textcolor{red}{0.809}} & \textbf{\textcolor{red}{0.828}} & \textbf{\textcolor{red}{0.809}} & 0.967 & 0.953 & 0.943 & 0.947 & 0.947 & 0.998 & 0.998 & \textbf{\textcolor{red}{1.000}} & 0.998 & 0.998 & 0.817 & 0.810 & 0.805 & 0.802 & \textbf{\textcolor{red}{0.822}} & 0.917 & 0.914 & 0.902 & 0.906 & 0.901 \\
& LogicalDeduction & 0.696 & \textbf{\textcolor{red}{0.730}} & 0.683 & 0.661 & 0.678 & 0.804 & 0.784 & 0.775 & 0.775 & 0.784 & 0.967 & \textbf{\textcolor{red}{0.980}} & 0.960 & 0.950 & 0.967 & 0.998 & \textbf{\textcolor{red}{1.000}} & \textbf{\textcolor{red}{1.000}} & \textbf{\textcolor{red}{1.000}} & \textbf{\textcolor{red}{1.000}} & 0.817 & 0.788 & 0.810 & 0.813 & 0.805 & 0.914 & 0.908 & 0.905 & \textbf{\textcolor{red}{0.916}} & 0.910 \\
& ProntoQA & -- & -- & -- & -- & -- & -- & -- & -- & -- & -- & 0.967 & 0.930 & 0.943 & 0.933 & 0.943 & -- & -- & -- & -- & -- & -- & -- & -- & -- & -- & -- & -- & -- & -- & -- \\
& ProofWriter & 0.735 & 0.717 & 0.648 & 0.665 & 0.670 & 0.804 & 0.789 & \textbf{\textcolor{red}{0.809}} & 0.804 & \textbf{\textcolor{red}{0.833}} & 0.967 & 0.933 & 0.937 & 0.953 & 0.943 & 0.998 & 0.990 & 0.998 & 0.992 & 0.998 & 0.817 & 0.793 & 0.798 & 0.793 & 0.787 & 0.917 & 0.901 & 0.906 & 0.909 & 0.907 \\
& gsm8k & 0.735 & 0.717 & 0.670 & 0.674 & 0.678 & 0.804 & 0.750 & 0.760 & 0.775 & 0.779 & 0.967 & 0.940 & 0.957 & 0.967 & 0.937 & 0.998 & 0.998 & 0.996 & 0.998 & 0.996 & 0.817 & 0.802 & \textbf{\textcolor{red}{0.820}} & \textbf{\textcolor{red}{0.823}} & 0.800 & 0.917 & -- & \textbf{\textcolor{red}{0.921}} & 0.910 & 0.909 \\
\midrule
\multirow{4}{*}{qwen3-8} & FOLIO & 0.683 & -- & -- & -- & -- & -- & -- & -- & -- & -- & -- & -- & -- & -- & -- & 1.000 & 0.996 & 1.000 & 1.000 & 1.000 & -- & -- & -- & -- & -- & 0.932 & 0.922 & 0.920 & 0.919 & 0.916 \\
& LogicalDeduction & 0.683 & \textbf{\textcolor{red}{0.691}} & 0.683 & \textbf{\textcolor{red}{0.713}} & \textbf{\textcolor{red}{0.730}} & 0.779 & \textbf{\textcolor{red}{0.789}} & \textbf{\textcolor{red}{0.809}} & \textbf{\textcolor{red}{0.799}} & \textbf{\textcolor{red}{0.799}} & 0.950 & \textbf{\textcolor{red}{0.980}} & \textbf{\textcolor{red}{0.990}} & \textbf{\textcolor{red}{0.977}} & \textbf{\textcolor{red}{0.973}} & 1.000 & 1.000 & 0.998 & 0.998 & 0.998 & 0.872 & 0.860 & 0.818 & 0.840 & 0.857 & 0.932 & 0.920 & 0.920 & 0.919 & 0.923 \\
& ProntoQA & 0.683 & 0.674 & \textbf{\textcolor{red}{0.704}} & \textbf{\textcolor{red}{0.704}} & 0.661 & 0.779 & \textbf{\textcolor{red}{0.828}} & \textbf{\textcolor{red}{0.814}} & \textbf{\textcolor{red}{0.819}} & \textbf{\textcolor{red}{0.848}} & 0.950 & 0.887 & 0.890 & 0.860 & 0.907 & 1.000 & 0.982 & 0.992 & 0.998 & 0.996 & 0.872 & 0.838 & 0.835 & 0.825 & 0.823 & 0.932 & 0.924 & 0.932 & 0.920 & 0.918 \\
& gsm8k & 0.683 & \textbf{\textcolor{red}{0.722}} & \textbf{\textcolor{red}{0.713}} & \textbf{\textcolor{red}{0.739}} & \textbf{\textcolor{red}{0.717}} & 0.779 & \textbf{\textcolor{red}{0.804}} & \textbf{\textcolor{red}{0.789}} & \textbf{\textcolor{red}{0.799}} & \textbf{\textcolor{red}{0.789}} & 0.950 & 0.943 & 0.930 & 0.947 & \textbf{\textcolor{red}{0.973}} & 1.000 & 0.998 & 0.996 & 0.996 & 0.994 & 0.872 & 0.855 & 0.855 & \textbf{\textcolor{red}{0.885}} & \textbf{\textcolor{red}{0.887}} & 0.932 & 0.930 & \textbf{\textcolor{red}{0.936}} & 0.931 & \textbf{\textcolor{red}{0.939}} \\
\midrule
\multirow{5}{*}{qwen3-14} & FOLIO & 0.791 & 0.757 & 0.726 & 0.743 & 0.730 & 0.828 & 0.828 & 0.828 & \textbf{\textcolor{red}{0.833}} & \textbf{\textcolor{red}{0.838}} & 0.993 & 0.990 & 0.983 & 0.973 & 0.990 & 0.982 & \textbf{\textcolor{red}{0.996}} & \textbf{\textcolor{red}{1.000}} & \textbf{\textcolor{red}{0.998}} & \textbf{\textcolor{red}{0.996}} & 0.838 & 0.820 & 0.815 & 0.820 & 0.820 & 0.876 & \textbf{\textcolor{red}{0.912}} & \textbf{\textcolor{red}{0.916}} & \textbf{\textcolor{red}{0.910}} & \textbf{\textcolor{red}{0.910}} \\
& LogicalDeduction & 0.791 & 0.783 & 0.774 & 0.743 & 0.739 & 0.828 & \textbf{\textcolor{red}{0.838}} & \textbf{\textcolor{red}{0.848}} & \textbf{\textcolor{red}{0.833}} & 0.824 & 0.993 & 0.983 & 0.987 & 0.983 & 0.987 & 0.982 & \textbf{\textcolor{red}{0.986}} & 0.982 & 0.982 & 0.970 & 0.838 & 0.802 & 0.798 & 0.815 & 0.807 & 0.876 & \textbf{\textcolor{red}{0.914}} & \textbf{\textcolor{red}{0.912}} & \textbf{\textcolor{red}{0.913}} & \textbf{\textcolor{red}{0.907}} \\
& ProntoQA & 0.791 & 0.748 & 0.761 & 0.761 & 0.722 & 0.828 & \textbf{\textcolor{red}{0.833}} & \textbf{\textcolor{red}{0.848}} & 0.824 & 0.828 & 0.993 & 0.990 & 0.973 & 0.993 & 0.990 & 0.982 & \textbf{\textcolor{red}{0.998}} & \textbf{\textcolor{red}{1.000}} & \textbf{\textcolor{red}{0.998}} & \textbf{\textcolor{red}{1.000}} & 0.838 & 0.798 & 0.807 & 0.822 & 0.808 & 0.876 & \textbf{\textcolor{red}{0.918}} & \textbf{\textcolor{red}{0.914}} & \textbf{\textcolor{red}{0.908}} & \textbf{\textcolor{red}{0.920}} \\
& ProofWriter & 0.791 & 0.704 & 0.722 & 0.700 & 0.726 & 0.828 & \textbf{\textcolor{red}{0.838}} & \textbf{\textcolor{red}{0.833}} & 0.819 & \textbf{\textcolor{red}{0.838}} & 0.993 & -- & -- & -- & -- & 0.982 & \textbf{\textcolor{red}{1.000}} & \textbf{\textcolor{red}{1.000}} & \textbf{\textcolor{red}{0.994}} & \textbf{\textcolor{red}{0.998}} & 0.838 & 0.837 & \textbf{\textcolor{red}{0.847}} & \textbf{\textcolor{red}{0.840}} & 0.835 & 0.876 & \textbf{\textcolor{red}{0.913}} & \textbf{\textcolor{red}{0.910}} & \textbf{\textcolor{red}{0.907}} & \textbf{\textcolor{red}{0.911}} \\
& gsm8k & 0.791 & 0.770 & 0.752 & \textbf{\textcolor{red}{0.800}} & 0.770 & 0.828 & 0.828 & 0.814 & 0.824 & 0.794 & 0.993 & 0.993 & \textbf{\textcolor{red}{0.997}} & 0.987 & 0.987 & 0.982 & 0.976 & 0.980 & 0.974 & 0.964 & 0.838 & 0.833 & 0.823 & \textbf{\textcolor{red}{0.843}} & -- & 0.876 & \textbf{\textcolor{red}{0.920}} & \textbf{\textcolor{red}{0.927}} & \textbf{\textcolor{red}{0.926}} & \textbf{\textcolor{red}{0.923}} \\
\midrule
\multirow{5}{*}{qwen3-32} & FOLIO & 0.752 & 0.739 & 0.748 & \textbf{\textcolor{red}{0.757}} & 0.752 & -- & -- & -- & -- & -- & 0.913 & 0.600 & 0.637 & 0.683 & 0.707 & 1.000 & 1.000 & 1.000 & 1.000 & 1.000 & 0.860 & \textbf{\textcolor{red}{0.863}} & \textbf{\textcolor{red}{0.863}} & 0.857 & 0.845 & 0.929 & \textbf{\textcolor{red}{0.932}} & \textbf{\textcolor{red}{0.933}} & \textbf{\textcolor{red}{0.932}} & \textbf{\textcolor{red}{0.939}} \\
& LogicalDeduction & 0.757 & \textbf{\textcolor{red}{0.770}} & 0.743 & 0.735 & \textbf{\textcolor{red}{0.778}} & 0.765 & \textbf{\textcolor{red}{0.799}} & \textbf{\textcolor{red}{0.794}} & \textbf{\textcolor{red}{0.799}} & \textbf{\textcolor{red}{0.799}} & -- & -- & -- & -- & -- & 0.998 & 0.584 & 0.642 & 0.760 & 0.596 & 0.847 & \textbf{\textcolor{red}{0.848}} & \textbf{\textcolor{red}{0.857}} & 0.843 & \textbf{\textcolor{red}{0.852}} & 0.928 & \textbf{\textcolor{red}{0.934}} & \textbf{\textcolor{red}{0.932}} & \textbf{\textcolor{red}{0.930}} & \textbf{\textcolor{red}{0.929}} \\
& ProntoQA & 0.787 & 0.739 & 0.717 & 0.730 & 0.674 & 0.804 & 0.770 & \textbf{\textcolor{red}{0.819}} & \textbf{\textcolor{red}{0.824}} & \textbf{\textcolor{red}{0.819}} & 0.890 & 0.610 & 0.527 & 0.490 & 0.520 & -- & -- & -- & -- & -- & 0.828 & \textbf{\textcolor{red}{0.852}} & \textbf{\textcolor{red}{0.857}} & \textbf{\textcolor{red}{0.855}} & \textbf{\textcolor{red}{0.848}} & 0.927 & 0.926 & \textbf{\textcolor{red}{0.932}} & \textbf{\textcolor{red}{0.932}} & \textbf{\textcolor{red}{0.929}} \\
& ProofWriter & 0.761 & 0.757 & 0.713 & 0.687 & 0.691 & 0.775 & \textbf{\textcolor{red}{0.794}} & \textbf{\textcolor{red}{0.833}} & \textbf{\textcolor{red}{0.799}} & \textbf{\textcolor{red}{0.824}} & 0.910 & 0.690 & 0.580 & 0.623 & 0.573 & 1.000 & 0.998 & 1.000 & 1.000 & 1.000 & -- & -- & -- & -- & -- & 0.932 & \textbf{\textcolor{red}{0.936}} & 0.930 & 0.931 & 0.930 \\
& gsm8k & 0.774 & 0.735 & 0.730 & 0.700 & 0.713 & 0.804 & 0.770 & 0.789 & 0.794 & 0.794 & 0.887 & \textbf{\textcolor{red}{0.927}} & 0.563 & \textbf{\textcolor{red}{1000.000}} & 0.577 & 0.998 & \textbf{\textcolor{red}{1.000}} & \textbf{\textcolor{red}{1.000}} & \textbf{\textcolor{red}{1.000}} & 0.996 & 0.828 & \textbf{\textcolor{red}{0.872}} & \textbf{\textcolor{red}{0.848}} & \textbf{\textcolor{red}{0.847}} & \textbf{\textcolor{red}{0.855}} & -- & -- & -- & -- & -- \\
\midrule
\multirow{5}{*}{llama3-8} & FOLIO & 0.261 & 0.226 & 0.209 & 0.217 & 0.226 & 0.603 & \textbf{\textcolor{red}{0.627}} & 0.588 & \textbf{\textcolor{red}{0.608}} & \textbf{\textcolor{red}{0.618}} & 0.527 & 0.507 & 0.480 & 0.440 & 0.467 & 0.782 & \textbf{\textcolor{red}{0.848}} & 0.782 & 0.766 & \textbf{\textcolor{red}{0.794}} & 0.457 & \textbf{\textcolor{red}{0.520}} & \textbf{\textcolor{red}{0.523}} & \textbf{\textcolor{red}{0.520}} & \textbf{\textcolor{red}{0.523}} & 0.614 & 0.575 & 0.575 & 0.546 & 0.541 \\
& LogicalDeduction & 0.261 & 0.257 & 0.213 & 0.204 & 0.235 & 0.603 & 0.549 & 0.569 & 0.593 & 0.574 & 0.527 & 0.510 & 0.500 & \textbf{\textcolor{red}{0.530}} & 0.480 & 0.784 & \textbf{\textcolor{red}{0.834}} & \textbf{\textcolor{red}{0.792}} & \textbf{\textcolor{red}{0.846}} & \textbf{\textcolor{red}{0.838}} & 0.457 & \textbf{\textcolor{red}{0.463}} & \textbf{\textcolor{red}{0.513}} & \textbf{\textcolor{red}{0.525}} & \textbf{\textcolor{red}{0.518}} & 0.620 & 0.604 & 0.591 & 0.576 & 0.570 \\
& ProntoQA & -- & -- & -- & -- & -- & -- & -- & -- & -- & -- & 0.527 & 0.500 & 0.450 & 0.457 & 0.493 & -- & -- & -- & -- & -- & -- & -- & -- & -- & -- & -- & -- & -- & -- & -- \\
& ProofWriter & 0.261 & 0.248 & \textbf{\textcolor{red}{0.274}} & 0.252 & 0.230 & 0.603 & 0.583 & 0.593 & 0.564 & 0.583 & 0.527 & 0.507 & 0.470 & 0.503 & 0.463 & 0.782 & \textbf{\textcolor{red}{0.844}} & \textbf{\textcolor{red}{0.826}} & \textbf{\textcolor{red}{0.838}} & \textbf{\textcolor{red}{0.810}} & 0.457 & \textbf{\textcolor{red}{0.562}} & \textbf{\textcolor{red}{0.593}} & \textbf{\textcolor{red}{0.592}} & \textbf{\textcolor{red}{0.583}} & 0.614 & 0.587 & 0.575 & 0.569 & 0.548 \\
& gsm8k & 0.261 & \textbf{\textcolor{red}{0.317}} & 0.222 & 0.239 & 0.200 & 0.603 & 0.583 & 0.549 & 0.495 & 0.578 & 0.527 & \textbf{\textcolor{red}{0.567}} & 0.507 & 0.470 & 0.450 & 0.782 & 0.772 & 0.776 & 0.782 & 0.770 & 0.457 & \textbf{\textcolor{red}{0.512}} & \textbf{\textcolor{red}{0.518}} & \textbf{\textcolor{red}{0.478}} & \textbf{\textcolor{red}{0.480}} & 0.614 & \textbf{\textcolor{red}{0.663}} & \textbf{\textcolor{red}{0.714}} & \textbf{\textcolor{red}{0.732}} & \textbf{\textcolor{red}{0.714}} \\

\bottomrule
\end{tabular}
}%
\caption{RAG\_bm25 cross-domain EM of Qwen models. Red numbers indicate improvements over 0-shot.}
\end{sidewaystable*}

\begin{sidewaystable*}[t]
\centering
\small
\resizebox{1.05\textwidth}{!}{%
\setlength{\tabcolsep}{3pt}
\begin{tabular}{ll
ccccc
ccccc
ccccc
ccccc
ccccc
ccccc}
\toprule
\textbf{Model} & \textbf{Source}
& \multicolumn{5}{c}{\textbf{AR-LSAT}}
& \multicolumn{5}{c}{\textbf{FOLIO}}
& \multicolumn{5}{c}{\textbf{LogicalDeduction}}
& \multicolumn{5}{c}{\textbf{ProntoQA}}
& \multicolumn{5}{c}{\textbf{ProofWriter}}
& \multicolumn{5}{c}{\textbf{gsm8k}} \\
\cmidrule(lr){3-7}
\cmidrule(lr){8-12}
\cmidrule(lr){13-17}
\cmidrule(lr){18-22}
\cmidrule(lr){23-27}
\cmidrule(lr){28-32}
& & 0 & 1 & 2 & 3 & 4
  & 0 & 1 & 2 & 3 & 4
  & 0 & 1 & 2 & 3 & 4
  & 0 & 1 & 2 & 3 & 4
  & 0 & 1 & 2 & 3 & 4
  & 0 & 1 & 2 & 3 & 4 \\
\midrule
\multirow{5}{*}{qwen3} & FOLIO & 0.230 & 0.209 & 0.222 & 0.226 & 0.230 & -- & -- & -- & -- & -- & 0.453 & \textbf{\textcolor{red}{0.517}} & \textbf{\textcolor{red}{0.467}} & \textbf{\textcolor{red}{0.463}} & \textbf{\textcolor{red}{0.470}} & 0.864 & 0.864 & \textbf{\textcolor{red}{0.866}} & 0.822 & 0.846 & 0.533 & \textbf{\textcolor{red}{0.573}} & \textbf{\textcolor{red}{0.555}} & \textbf{\textcolor{red}{0.535}} & \textbf{\textcolor{red}{0.543}} & 0.745 & 0.713 & 0.704 & 0.685 & 0.668 \\
& LogicalDeduction & 0.226 & 0.183 & 0.178 & 0.209 & 0.226 & 0.623 & \textbf{\textcolor{red}{0.667}} & \textbf{\textcolor{red}{0.637}} & 0.618 & 0.593 & -- & -- & -- & -- & -- & 0.854 & \textbf{\textcolor{red}{0.856}} & \textbf{\textcolor{red}{0.856}} & 0.842 & 0.828 & 0.510 & \textbf{\textcolor{red}{0.528}} & \textbf{\textcolor{red}{0.557}} & \textbf{\textcolor{red}{0.517}} & \textbf{\textcolor{red}{0.542}} & 0.738 & 0.701 & 0.702 & 0.675 & 0.669 \\
& ProntoQA & 0.274 & 0.170 & 0.209 & 0.174 & 0.187 & 0.647 & 0.515 & 0.475 & 0.446 & 0.456 & 0.450 & 0.450 & 0.420 & 0.433 & 0.430 & -- & -- & -- & -- & -- & 0.497 & 0.418 & 0.360 & 0.312 & 0.292 & 0.737 & 0.698 & 0.693 & 0.717 & 0.694 \\
& ProofWriter & 0.230 & 0.217 & 0.183 & 0.209 & 0.213 & 0.623 & 0.603 & \textbf{\textcolor{red}{0.676}} & \textbf{\textcolor{red}{0.632}} & \textbf{\textcolor{red}{0.642}} & 0.447 & \textbf{\textcolor{red}{0.513}} & \textbf{\textcolor{red}{0.467}} & \textbf{\textcolor{red}{0.467}} & \textbf{\textcolor{red}{0.473}} & 0.862 & 0.850 & \textbf{\textcolor{red}{0.892}} & \textbf{\textcolor{red}{0.868}} & 0.854 & -- & -- & -- & -- & -- & 0.738 & 0.717 & 0.712 & 0.691 & 0.667 \\
& gsm8k & 0.274 & 0.213 & 0.239 & 0.178 & 0.217 & 0.647 & 0.588 & 0.588 & 0.588 & 0.608 & 0.447 & \textbf{\textcolor{red}{0.457}} & \textbf{\textcolor{red}{0.453}} & \textbf{\textcolor{red}{0.450}} & 0.417 & 0.862 & 0.758 & 0.766 & 0.660 & 0.638 & 0.512 & 0.468 & 0.448 & 0.402 & 0.422 & -- & -- & -- & -- & -- \\
\midrule
\multirow{5}{*}{qwen7} & FOLIO & 0.217 & \textbf{\textcolor{red}{0.248}} & \textbf{\textcolor{red}{0.239}} & \textbf{\textcolor{red}{0.283}} & \textbf{\textcolor{red}{0.257}} & -- & -- & -- & -- & -- & 0.637 & 0.617 & \textbf{\textcolor{red}{0.683}} & 0.617 & 0.630 & 0.978 & 0.962 & \textbf{\textcolor{red}{0.980}} & \textbf{\textcolor{red}{0.982}} & 0.974 & 0.632 & 0.625 & \textbf{\textcolor{red}{0.650}} & \textbf{\textcolor{red}{0.672}} & 0.632 & 0.884 & \textbf{\textcolor{red}{0.892}} & \textbf{\textcolor{red}{0.893}} & 0.884 & 0.871 \\
& LogicalDeduction & 0.248 & 0.243 & \textbf{\textcolor{red}{0.278}} & \textbf{\textcolor{red}{0.278}} & \textbf{\textcolor{red}{0.265}} & 0.711 & 0.632 & 0.706 & 0.701 & 0.681 & -- & -- & -- & -- & -- & 0.978 & 0.972 & \textbf{\textcolor{red}{0.982}} & 0.974 & 0.978 & 0.633 & 0.613 & 0.612 & 0.622 & 0.630 & 0.884 & 0.881 & \textbf{\textcolor{red}{0.890}} & \textbf{\textcolor{red}{0.889}} & 0.883 \\
& ProntoQA & 0.209 & \textbf{\textcolor{red}{0.248}} & \textbf{\textcolor{red}{0.226}} & \textbf{\textcolor{red}{0.248}} & \textbf{\textcolor{red}{0.252}} & 0.662 & 0.618 & \textbf{\textcolor{red}{0.667}} & 0.632 & 0.642 & 0.637 & 0.550 & \textbf{\textcolor{red}{0.653}} & 0.627 & 0.620 & -- & -- & -- & -- & -- & 0.627 & 0.613 & 0.627 & 0.603 & \textbf{\textcolor{red}{0.650}} & 0.887 & 0.863 & 0.878 & 0.885 & 0.883 \\
& ProofWriter & 0.222 & \textbf{\textcolor{red}{0.248}} & \textbf{\textcolor{red}{0.283}} & \textbf{\textcolor{red}{0.248}} & \textbf{\textcolor{red}{0.239}} & 0.701 & \textbf{\textcolor{red}{0.711}} & 0.696 & 0.701 & \textbf{\textcolor{red}{0.706}} & 0.653 & 0.630 & 0.650 & 0.650 & 0.637 & 0.976 & 0.970 & 0.970 & \textbf{\textcolor{red}{0.984}} & \textbf{\textcolor{red}{0.986}} & -- & -- & -- & -- & -- & 0.889 & 0.871 & 0.873 & 0.873 & 0.862 \\
& gsm8k & 0.243 & \textbf{\textcolor{red}{0.274}} & \textbf{\textcolor{red}{0.257}} & \textbf{\textcolor{red}{0.283}} & \textbf{\textcolor{red}{0.265}} & 0.662 & 0.574 & 0.647 & 0.642 & 0.652 & 0.643 & 0.603 & 0.633 & 0.607 & 0.597 & 0.976 & 0.932 & 0.924 & 0.924 & 0.932 & 0.627 & 0.627 & 0.625 & 0.622 & 0.605 & -- & -- & -- & -- & -- \\
\midrule
\multirow{5}{*}{qwen14} & FOLIO & 0.343 & \textbf{\textcolor{red}{0.348}} & 0.274 & 0.335 & 0.304 & 0.681 & \textbf{\textcolor{red}{0.750}} & \textbf{\textcolor{red}{0.745}} & \textbf{\textcolor{red}{0.765}} & \textbf{\textcolor{red}{0.760}} & 0.800 & 0.793 & 0.780 & \textbf{\textcolor{red}{0.820}} & 0.773 & 0.942 & \textbf{\textcolor{red}{0.962}} & \textbf{\textcolor{red}{0.968}} & \textbf{\textcolor{red}{0.962}} & \textbf{\textcolor{red}{0.972}} & 0.688 & \textbf{\textcolor{red}{0.715}} & \textbf{\textcolor{red}{0.735}} & \textbf{\textcolor{red}{0.738}} & \textbf{\textcolor{red}{0.732}} & 0.938 & 0.931 & 0.932 & 0.936 & 0.929 \\
& LogicalDeduction & 0.330 & \textbf{\textcolor{red}{0.348}} & 0.317 & \textbf{\textcolor{red}{0.365}} & \textbf{\textcolor{red}{0.343}} & 0.696 & 0.686 & \textbf{\textcolor{red}{0.740}} & \textbf{\textcolor{red}{0.706}} & \textbf{\textcolor{red}{0.725}} & 0.797 & \textbf{\textcolor{red}{0.810}} & 0.777 & \textbf{\textcolor{red}{0.810}} & \textbf{\textcolor{red}{0.807}} & 0.956 & \textbf{\textcolor{red}{0.964}} & \textbf{\textcolor{red}{0.976}} & \textbf{\textcolor{red}{0.968}} & \textbf{\textcolor{red}{0.974}} & 0.680 & \textbf{\textcolor{red}{0.702}} & \textbf{\textcolor{red}{0.698}} & \textbf{\textcolor{red}{0.700}} & \textbf{\textcolor{red}{0.707}} & 0.936 & 0.930 & 0.934 & 0.931 & 0.931 \\
& ProntoQA & 0.322 & 0.304 & 0.313 & \textbf{\textcolor{red}{0.339}} & 0.296 & 0.721 & \textbf{\textcolor{red}{0.745}} & \textbf{\textcolor{red}{0.755}} & 0.701 & 0.696 & 0.787 & 0.767 & 0.767 & 0.763 & 0.743 & 0.948 & \textbf{\textcolor{red}{0.972}} & \textbf{\textcolor{red}{0.978}} & \textbf{\textcolor{red}{0.982}} & \textbf{\textcolor{red}{0.984}} & 0.678 & 0.622 & 0.637 & 0.607 & 0.577 & 0.933 & 0.929 & 0.926 & 0.918 & 0.917 \\
& ProofWriter & 0.304 & \textbf{\textcolor{red}{0.335}} & 0.270 & \textbf{\textcolor{red}{0.343}} & \textbf{\textcolor{red}{0.370}} & 0.672 & \textbf{\textcolor{red}{0.735}} & \textbf{\textcolor{red}{0.750}} & \textbf{\textcolor{red}{0.755}} & \textbf{\textcolor{red}{0.745}} & 0.813 & 0.787 & 0.787 & 0.807 & 0.803 & -- & -- & -- & -- & 0.974 & 0.677 & \textbf{\textcolor{red}{0.738}} & \textbf{\textcolor{red}{0.752}} & \textbf{\textcolor{red}{0.762}} & \textbf{\textcolor{red}{0.737}} & 0.936 & \textbf{\textcolor{red}{0.937}} & 0.927 & 0.931 & 0.931 \\
& gsm8k & 0.322 & 0.304 & \textbf{\textcolor{red}{0.357}} & 0.274 & 0.291 & 0.711 & 0.686 & 0.706 & 0.706 & 0.686 & 0.810 & 0.780 & \textbf{\textcolor{red}{0.817}} & 0.783 & 0.773 & 0.948 & 0.940 & 0.900 & 0.894 & 0.880 & 0.688 & 0.620 & 0.607 & 0.605 & 0.583 & 0.932 & 0.922 & 0.924 & 0.923 & 0.926 \\
\midrule
\multirow{5}{*}{qwen32} & FOLIO & 0.348 & \textbf{\textcolor{red}{0.374}} & 0.339 & \textbf{\textcolor{red}{0.357}} & \textbf{\textcolor{red}{0.357}} & -- & -- & -- & -- & -- & 0.850 & 0.847 & 0.837 & \textbf{\textcolor{red}{0.880}} & \textbf{\textcolor{red}{0.890}} & 1.000 & 0.998 & 0.998 & 1.000 & 0.998 & 0.718 & 0.693 & \textbf{\textcolor{red}{0.742}} & \textbf{\textcolor{red}{0.748}} & \textbf{\textcolor{red}{0.735}} & 0.893 & \textbf{\textcolor{red}{0.909}} & \textbf{\textcolor{red}{0.906}} & \textbf{\textcolor{red}{0.895}} & 0.883 \\
& LogicalDeduction & 0.348 & \textbf{\textcolor{red}{0.357}} & 0.309 & 0.343 & 0.330 & 0.725 & \textbf{\textcolor{red}{0.745}} & \textbf{\textcolor{red}{0.750}} & 0.721 & \textbf{\textcolor{red}{0.750}} & -- & -- & -- & -- & -- & 1.000 & 0.996 & 0.996 & 0.998 & 0.998 & 0.688 & \textbf{\textcolor{red}{0.733}} & \textbf{\textcolor{red}{0.728}} & \textbf{\textcolor{red}{0.725}} & \textbf{\textcolor{red}{0.738}} & 0.892 & \textbf{\textcolor{red}{0.906}} & \textbf{\textcolor{red}{0.912}} & \textbf{\textcolor{red}{0.906}} & \textbf{\textcolor{red}{0.894}} \\
& ProntoQA & 0.304 & \textbf{\textcolor{red}{0.365}} & \textbf{\textcolor{red}{0.322}} & \textbf{\textcolor{red}{0.322}} & 0.261 & 0.745 & 0.745 & 0.701 & 0.672 & 0.686 & 0.857 & 0.830 & 0.813 & 0.817 & 0.823 & -- & -- & -- & -- & -- & 0.695 & 0.675 & 0.682 & 0.680 & 0.682 & 0.892 & 0.886 & \textbf{\textcolor{red}{0.905}} & \textbf{\textcolor{red}{0.911}} & \textbf{\textcolor{red}{0.923}} \\
& ProofWriter & 0.335 & 0.313 & 0.322 & \textbf{\textcolor{red}{0.352}} & \textbf{\textcolor{red}{0.348}} & 0.750 & 0.725 & \textbf{\textcolor{red}{0.755}} & 0.735 & 0.730 & 0.847 & \textbf{\textcolor{red}{0.873}} & \textbf{\textcolor{red}{0.860}} & \textbf{\textcolor{red}{0.873}} & \textbf{\textcolor{red}{0.877}} & -- & -- & -- & -- & -- & -- & -- & -- & -- & -- & 0.889 & \textbf{\textcolor{red}{0.898}} & \textbf{\textcolor{red}{0.912}} & 0.880 & 0.870 \\
& gsm8k & 0.326 & 0.317 & \textbf{\textcolor{red}{0.335}} & \textbf{\textcolor{red}{0.339}} & 0.322 & 0.740 & 0.716 & 0.711 & 0.696 & 0.716 & 0.840 & \textbf{\textcolor{red}{0.870}} & 0.837 & 0.807 & 0.783 & 0.998 & 0.996 & 0.994 & 0.988 & 0.988 & 0.703 & 0.613 & 0.617 & 0.618 & 0.628 & -- & -- & -- & -- & -- \\
\midrule
\multirow{5}{*}{gemma3-4} & FOLIO & 0.274 & \textbf{\textcolor{red}{0.300}} & 0.243 & 0.226 & 0.243 & -- & -- & -- & -- & -- & 0.627 & 0.623 & 0.600 & 0.627 & 0.603 & 0.918 & \textbf{\textcolor{red}{0.930}} & \textbf{\textcolor{red}{0.930}} & \textbf{\textcolor{red}{0.944}} & 0.912 & 0.650 & 0.618 & 0.643 & 0.635 & 0.642 & -- & -- & -- & -- & -- \\
& LogicalDeduction & 0.235 & \textbf{\textcolor{red}{0.261}} & \textbf{\textcolor{red}{0.283}} & 0.235 & \textbf{\textcolor{red}{0.248}} & 0.618 & 0.618 & 0.613 & 0.603 & 0.588 & -- & -- & -- & -- & -- & 0.920 & 0.876 & 0.878 & 0.856 & 0.892 & 0.642 & 0.532 & 0.562 & 0.532 & 0.547 & -- & -- & -- & -- & -- \\
& ProntoQA & 0.243 & 0.230 & 0.230 & \textbf{\textcolor{red}{0.261}} & \textbf{\textcolor{red}{0.248}} & 0.623 & 0.608 & 0.564 & 0.598 & 0.603 & 0.600 & \textbf{\textcolor{red}{0.623}} & \textbf{\textcolor{red}{0.613}} & 0.587 & \textbf{\textcolor{red}{0.633}} & -- & -- & -- & -- & -- & 0.647 & 0.620 & 0.625 & 0.575 & 0.455 & 0.863 & 0.850 & 0.839 & 0.844 & 0.852 \\
& ProofWriter & 0.287 & 0.270 & 0.265 & 0.243 & 0.261 & 0.608 & \textbf{\textcolor{red}{0.642}} & \textbf{\textcolor{red}{0.637}} & \textbf{\textcolor{red}{0.652}} & \textbf{\textcolor{red}{0.618}} & 0.610 & 0.587 & 0.600 & \textbf{\textcolor{red}{0.623}} & 0.597 & -- & -- & -- & -- & -- & -- & -- & -- & -- & -- & -- & -- & -- & -- & -- \\
& gsm8k & 0.248 & 0.235 & 0.230 & 0.248 & \textbf{\textcolor{red}{0.265}} & 0.637 & 0.593 & 0.623 & 0.598 & 0.588 & 0.593 & \textbf{\textcolor{red}{0.647}} & 0.567 & 0.580 & 0.583 & 0.912 & 0.868 & 0.846 & 0.834 & 0.818 & 0.647 & 0.568 & 0.573 & 0.578 & 0.553 & -- & -- & -- & -- & -- \\
\midrule
\multirow{5}{*}{gemma3-12} & FOLIO & 0.304 & 0.287 & 0.291 & \textbf{\textcolor{red}{0.322}} & \textbf{\textcolor{red}{0.330}} & -- & -- & -- & -- & -- & 0.757 & 0.753 & \textbf{\textcolor{red}{0.767}} & \textbf{\textcolor{red}{0.770}} & \textbf{\textcolor{red}{0.763}} & 0.992 & 0.980 & 0.982 & 0.988 & 0.984 & 0.740 & \textbf{\textcolor{red}{0.762}} & \textbf{\textcolor{red}{0.768}} & \textbf{\textcolor{red}{0.753}} & \textbf{\textcolor{red}{0.778}} & -- & -- & -- & -- & -- \\
& LogicalDeduction & 0.317 & 0.274 & 0.304 & 0.270 & \textbf{\textcolor{red}{0.322}} & 0.725 & 0.706 & \textbf{\textcolor{red}{0.740}} & 0.725 & 0.686 & -- & -- & -- & -- & -- & 0.986 & 0.984 & \textbf{\textcolor{red}{0.990}} & 0.986 & 0.982 & 0.747 & 0.738 & \textbf{\textcolor{red}{0.755}} & \textbf{\textcolor{red}{0.753}} & \textbf{\textcolor{red}{0.768}} & -- & -- & -- & -- & -- \\
& ProntoQA & 0.313 & \textbf{\textcolor{red}{0.361}} & \textbf{\textcolor{red}{0.339}} & 0.287 & 0.287 & 0.706 & 0.701 & 0.706 & 0.681 & \textbf{\textcolor{red}{0.716}} & 0.767 & 0.740 & 0.757 & 0.747 & 0.720 & -- & -- & -- & -- & -- & 0.750 & 0.650 & 0.622 & 0.652 & 0.597 & 0.917 & 0.909 & 0.916 & 0.901 & 0.905 \\
& ProofWriter & 0.317 & 0.274 & 0.309 & 0.287 & \textbf{\textcolor{red}{0.335}} & 0.735 & \textbf{\textcolor{red}{0.745}} & \textbf{\textcolor{red}{0.779}} & \textbf{\textcolor{red}{0.775}} & \textbf{\textcolor{red}{0.755}} & 0.773 & 0.757 & 0.747 & 0.773 & \textbf{\textcolor{red}{0.800}} & -- & -- & -- & -- & -- & -- & -- & -- & -- & -- & -- & -- & -- & -- & -- \\
& gsm8k & 0.296 & 0.278 & \textbf{\textcolor{red}{0.335}} & \textbf{\textcolor{red}{0.300}} & \textbf{\textcolor{red}{0.335}} & 0.730 & 0.652 & 0.676 & 0.701 & 0.667 & 0.763 & 0.747 & \textbf{\textcolor{red}{0.773}} & \textbf{\textcolor{red}{0.777}} & \textbf{\textcolor{red}{0.793}} & 0.986 & 0.970 & 0.974 & \textbf{\textcolor{red}{0.988}} & \textbf{\textcolor{red}{0.996}} & 0.752 & 0.698 & 0.717 & 0.737 & 0.747 & -- & -- & -- & -- & -- \\
\midrule
\multirow{5}{*}{gemma3-27} & FOLIO & -- & 0.322 & -- & 0.409 & -- & -- & -- & -- & -- & -- & 0.823 & \textbf{\textcolor{red}{0.830}} & \textbf{\textcolor{red}{0.857}} & \textbf{\textcolor{red}{0.890}} & \textbf{\textcolor{red}{0.857}} & 0.986 & \textbf{\textcolor{red}{0.988}} & \textbf{\textcolor{red}{0.988}} & \textbf{\textcolor{red}{0.996}} & \textbf{\textcolor{red}{0.994}} & 0.767 & \textbf{\textcolor{red}{0.810}} & \textbf{\textcolor{red}{0.820}} & \textbf{\textcolor{red}{0.812}} & \textbf{\textcolor{red}{0.808}} & -- & -- & -- & -- & -- \\
& LogicalDeduction & 0.374 & \textbf{\textcolor{red}{0.400}} & 0.370 & \textbf{\textcolor{red}{0.391}} & \textbf{\textcolor{red}{0.387}} & 0.706 & \textbf{\textcolor{red}{0.760}} & \textbf{\textcolor{red}{0.735}} & \textbf{\textcolor{red}{0.755}} & \textbf{\textcolor{red}{0.735}} & -- & -- & -- & -- & -- & 0.984 & 0.984 & \textbf{\textcolor{red}{0.990}} & \textbf{\textcolor{red}{0.986}} & \textbf{\textcolor{red}{0.992}} & 0.760 & \textbf{\textcolor{red}{0.772}} & \textbf{\textcolor{red}{0.792}} & \textbf{\textcolor{red}{0.770}} & \textbf{\textcolor{red}{0.767}} & -- & -- & -- & -- & -- \\
& ProntoQA & 0.357 & \textbf{\textcolor{red}{0.378}} & \textbf{\textcolor{red}{0.370}} & \textbf{\textcolor{red}{0.361}} & 0.352 & 0.721 & \textbf{\textcolor{red}{0.735}} & 0.711 & 0.696 & \textbf{\textcolor{red}{0.735}} & 0.820 & \textbf{\textcolor{red}{0.867}} & \textbf{\textcolor{red}{0.870}} & \textbf{\textcolor{red}{0.863}} & \textbf{\textcolor{red}{0.863}} & -- & -- & -- & -- & -- & 0.765 & 0.757 & 0.735 & 0.737 & 0.718 & 0.935 & 0.927 & 0.934 & 0.932 & 0.931 \\
& ProofWriter & 0.383 & \textbf{\textcolor{red}{0.400}} & \textbf{\textcolor{red}{0.387}} & 0.378 & 0.352 & 0.716 & \textbf{\textcolor{red}{0.765}} & \textbf{\textcolor{red}{0.784}} & \textbf{\textcolor{red}{0.789}} & \textbf{\textcolor{red}{0.765}} & 0.823 & \textbf{\textcolor{red}{0.847}} & \textbf{\textcolor{red}{0.853}} & \textbf{\textcolor{red}{0.850}} & \textbf{\textcolor{red}{0.877}} & -- & -- & -- & -- & -- & -- & -- & -- & -- & -- & -- & -- & -- & -- & -- \\
& gsm8k & 0.370 & 0.361 & \textbf{\textcolor{red}{0.383}} & 0.339 & \textbf{\textcolor{red}{0.374}} & 0.721 & 0.716 & 0.711 & 0.716 & 0.716 & 0.830 & \textbf{\textcolor{red}{0.903}} & \textbf{\textcolor{red}{0.890}} & \textbf{\textcolor{red}{0.883}} & \textbf{\textcolor{red}{0.860}} & 0.984 & 0.982 & -- & 0.980 & 0.978 & 0.773 & 0.737 & 0.735 & 0.732 & 0.742 & -- & -- & -- & -- & -- \\

\bottomrule
\end{tabular}
}%
\caption{ConE 1}
\end{sidewaystable*}

\begin{sidewaystable*}[t]
\centering
\small
\resizebox{1.05\textwidth}{!}{%
\setlength{\tabcolsep}{3pt}
\begin{tabular}{ll
ccccc
ccccc
ccccc
ccccc
ccccc
ccccc}
\toprule
\textbf{Model} & \textbf{Source}
& \multicolumn{5}{c}{\textbf{AR-LSAT}}
& \multicolumn{5}{c}{\textbf{FOLIO}}
& \multicolumn{5}{c}{\textbf{LogicalDeduction}}
& \multicolumn{5}{c}{\textbf{ProntoQA}}
& \multicolumn{5}{c}{\textbf{ProofWriter}}
& \multicolumn{5}{c}{\textbf{gsm8k}} \\
\cmidrule(lr){3-7}
\cmidrule(lr){8-12}
\cmidrule(lr){13-17}
\cmidrule(lr){18-22}
\cmidrule(lr){23-27}
\cmidrule(lr){28-32}
& & 0 & 1 & 2 & 3 & 4
  & 0 & 1 & 2 & 3 & 4
  & 0 & 1 & 2 & 3 & 4
  & 0 & 1 & 2 & 3 & 4
  & 0 & 1 & 2 & 3 & 4
  & 0 & 1 & 2 & 3 & 4 \\
\midrule
\multirow{5}{*}{qwen3-4} & FOLIO & 0.722 & 0.678 & 0.657 & 0.674 & 0.700 & -- & -- & -- & -- & -- & 0.970 & 0.940 & 0.960 & 0.923 & 0.920 & 1.000 & 0.994 & 0.992 & 0.996 & 0.990 & 0.823 & 0.823 & \textbf{\textcolor{red}{0.825}} & 0.823 & 0.812 & 0.908 & \textbf{\textcolor{red}{0.909}} & 0.897 & 0.904 & \textbf{\textcolor{red}{0.914}} \\
& LogicalDeduction & 0.726 & 0.700 & 0.704 & 0.709 & 0.713 & 0.804 & 0.804 & 0.775 & 0.784 & \textbf{\textcolor{red}{0.819}} & -- & -- & -- & -- & -- & 1.000 & 0.996 & 1.000 & 1.000 & 0.998 & 0.803 & 0.798 & \textbf{\textcolor{red}{0.805}} & \textbf{\textcolor{red}{0.823}} & 0.800 & 0.916 & 0.912 & 0.916 & 0.915 & \textbf{\textcolor{red}{0.918}} \\
& ProntoQA & 0.726 & 0.643 & 0.691 & 0.696 & 0.678 & 0.779 & 0.765 & \textbf{\textcolor{red}{0.799}} & 0.775 & 0.770 & 0.967 & 0.930 & 0.933 & 0.927 & 0.907 & -- & -- & -- & -- & -- & 0.822 & 0.785 & 0.782 & 0.767 & 0.757 & 0.917 & 0.906 & 0.905 & 0.876 & 0.863 \\
& ProofWriter & 0.722 & 0.635 & 0.665 & 0.670 & 0.665 & 0.804 & \textbf{\textcolor{red}{0.819}} & \textbf{\textcolor{red}{0.819}} & \textbf{\textcolor{red}{0.824}} & 0.804 & 0.967 & 0.967 & 0.927 & 0.940 & 0.940 & 1.000 & 0.996 & 0.996 & 0.994 & 0.994 & -- & -- & -- & -- & -- & 0.917 & 0.902 & 0.885 & 0.882 & 0.893 \\
& gsm8k & 0.713 & 0.678 & 0.652 & 0.691 & 0.687 & 0.779 & 0.740 & 0.770 & 0.775 & 0.750 & 0.960 & \textbf{\textcolor{red}{0.967}} & 0.953 & 0.947 & 0.957 & 1.000 & 0.996 & 0.998 & 0.998 & 0.998 & 0.820 & 0.793 & 0.800 & 0.807 & 0.818 & -- & -- & -- & -- & -- \\
\midrule
\multirow{5}{*}{qwen3-8} & FOLIO & 0.630 & \textbf{\textcolor{red}{0.722}} & \textbf{\textcolor{red}{0.691}} & \textbf{\textcolor{red}{0.691}} & \textbf{\textcolor{red}{0.665}} & -- & -- & -- & -- & -- & 0.937 & \textbf{\textcolor{red}{0.977}} & \textbf{\textcolor{red}{0.960}} & \textbf{\textcolor{red}{0.970}} & \textbf{\textcolor{red}{0.970}} & 1.000 & 1.000 & 1.000 & 1.000 & 1.000 & 0.878 & 0.872 & 0.860 & 0.838 & 0.860 & 0.928 & 0.926 & 0.922 & 0.917 & 0.917 \\
& LogicalDeduction & 0.630 & \textbf{\textcolor{red}{0.717}} & \textbf{\textcolor{red}{0.713}} & \textbf{\textcolor{red}{0.713}} & \textbf{\textcolor{red}{0.665}} & 0.779 & \textbf{\textcolor{red}{0.814}} & \textbf{\textcolor{red}{0.828}} & \textbf{\textcolor{red}{0.824}} & \textbf{\textcolor{red}{0.809}} & -- & -- & -- & -- & -- & 1.000 & 0.996 & 1.000 & 1.000 & 0.998 & 0.878 & 0.855 & 0.853 & 0.855 & 0.850 & 0.931 & 0.928 & 0.920 & 0.926 & 0.923 \\
& ProntoQA & 0.691 & \textbf{\textcolor{red}{0.713}} & \textbf{\textcolor{red}{0.730}} & 0.678 & 0.674 & 0.814 & 0.814 & 0.799 & 0.789 & \textbf{\textcolor{red}{0.828}} & 0.943 & \textbf{\textcolor{red}{0.957}} & 0.933 & 0.903 & 0.910 & -- & -- & -- & -- & -- & 0.875 & 0.867 & 0.858 & 0.830 & 0.850 & 0.926 & 0.925 & 0.923 & 0.918 & 0.922 \\
& ProofWriter & 0.630 & \textbf{\textcolor{red}{0.687}} & \textbf{\textcolor{red}{0.691}} & \textbf{\textcolor{red}{0.687}} & \textbf{\textcolor{red}{0.691}} & 0.784 & \textbf{\textcolor{red}{0.824}} & \textbf{\textcolor{red}{0.809}} & \textbf{\textcolor{red}{0.799}} & \textbf{\textcolor{red}{0.814}} & 0.927 & \textbf{\textcolor{red}{0.970}} & \textbf{\textcolor{red}{0.967}} & \textbf{\textcolor{red}{0.963}} & \textbf{\textcolor{red}{0.963}} & 1.000 & 0.998 & 1.000 & 1.000 & 1.000 & -- & -- & -- & -- & -- & 0.926 & 0.922 & 0.925 & 0.921 & 0.917 \\
& gsm8k & 0.635 & \textbf{\textcolor{red}{0.730}} & \textbf{\textcolor{red}{0.700}} & \textbf{\textcolor{red}{0.735}} & \textbf{\textcolor{red}{0.722}} & 0.814 & 0.799 & 0.794 & 0.789 & 0.814 & 0.923 & \textbf{\textcolor{red}{0.953}} & \textbf{\textcolor{red}{0.967}} & \textbf{\textcolor{red}{0.973}} & \textbf{\textcolor{red}{0.970}} & 1.000 & 0.996 & 0.998 & 0.998 & 0.998 & 0.872 & \textbf{\textcolor{red}{0.878}} & 0.870 & 0.872 & \textbf{\textcolor{red}{0.875}} & -- & -- & -- & -- & -- \\

\bottomrule
\end{tabular}
}%
\caption{ConE 2}
\end{sidewaystable*}

\clearpage

\subsection{Data Example}
\label{example}
\begin{table*}[ht!]
\centering
\small
\begin{tabular}{p{2cm}p{10.5cm}}
\toprule
\textbf{Setting} & \textbf{Model Reasoning and Output} \\
\midrule
\multirow{5}{*}{\textbf{Context}} & A Japanese game company created the game The Legend of Zelda.All games on the Top 10 list are made by Japanese game companies.If a game sells more than one million copies, then it will be included in the Top 10 list.The Legend of Zelda sold more than one million copies. \\
\midrule
\multirow{7}{*}{\textbf{Zero-shot}} &
\textit{Since the Legend of Zelda sold more than one million copies, it should be included in the Top 10 list. However, it is stated that the Legend of Zelda is not on the Top 10 list. This creates a contradiction with the information that all games on the Top 10 list are made by Japanese game companies and the Legend of Zelda is made by a Japanese game company. Therefore, the statement "The Legend of Zelda is not on the Top 10 list" must be true. Final answer: A} \\
& \textbf{Final answer: A (True)} \\
\midrule
\multirow{6}{*}{\textbf{Demostration}} & Question: In the last 25 years, 60$\%$ of Scrabble champions have been women and the rest have been men. There is one champion per year. Of the men, 40$\%$ had a beard. How many men with beards have been Scrabble champion? Reasoning: 40$\%$ of champions are men because 100 - 60 = <<100-60=40>>40 10 champions have been men because 25 x .4 = <<25*.4=10>>10 4 champions have been men with beards because 10 x .4 = <<10*.4=4>>4 \\
\midrule
\multirow{5}{*}{\textbf{Cross-domain}} &
\textit{Since the Legend of Zelda sold more than one million copies and all games on the Top 10 list are made by Japanese game companies, and the Legend of Zelda is made by a Japanese game company, it must be on the Top 10 list. Therefore, the statement "The Legend of Zelda is not on the Top 10 list" is false. Final answer: B} \\
& \textbf{Final answer: B (False)} \\
\bottomrule
\end{tabular}
\caption{Comparison between zero-shot reasoning and cross-domain reasoning reasoning on a cross-domain example. DIN-guided retrieval corrects the logical inconsistency made by zero-shot prompting.}
\label{tab:case1}
\end{table*}

\begin{table*}[ht!]
\centering
\small
\begin{tabular}{p{2cm}p{10.5cm}}
\toprule
\textbf{Setting} & \textbf{Model Reasoning and Output} \\
\midrule
\multirow{11}{*}{\textbf{Context}} & People in this club who perform in school talent shows often attend and are very engaged with school events.People in this club either perform in school talent shows often or are inactive and disinterested community members.People in this club who chaperone high school dances are not students who attend the school.All people in this club who are inactive and disinterested members of their community chaperone high school dances.All young children and teenagers in this club who wish to further their academic careers and educational opportunities are students who attend the school. Bonnie is in this club and she either both attends and is very engaged with school events and is a student who attends the school or is not someone who both attends and is very engaged with school events and is not a student who attends the school. \\
\midrule
\multirow{26}{*}{\textbf{Zero-shot}} &
\textit{Let's break down the information step by step:1. If Bonnie chaperones high school dances, then she is not a student who attends the school (from the third statement).2. If Bonnie does not chaperone high school dances, then she performs in school talent shows often (from the second statement).3. If Bonnie performs in school talent shows often, then she is very engaged with school events (from the first statement).4. If Bonnie is very engaged with school events, then she is a student who attends the school (from the first statement).5. From steps 1 and 4, we can conclude that Bonnie cannot be both a student who attends the school and chaperone high school dances.6. From steps 2 and 3, we can conclude that if Bonnie does not chaperone high school dances, then she is a student who attends the school.7. From steps 5 and 6, we can conclude that Bonnie is not a student who attends the school if she chaperones high school dances, and she is a student who attends the school if she does not chaperone high school dances.8. From the fourth statement, we can conclude that if Bonnie is a student who attends the school, then she is a young child or teenager who wishes to further her academic career and educational opportunities.9. From steps 7 and 8, we can conclude that Bonnie is a young child or teenager who wishes to further her academic career and educational opportunities if she does not chaperone high school dances.10. From the fifth statement, we can conclude that if Bonnie is an inactive and disinterested member of the community, then she chaperones high school dances.11. From steps 7 and 10, we can conclude that Bonnie is an inactive and disinterested member of the community if she chaperones high school dances.12. From steps 9 and 11, we can conclude that Bonnie is both a young child or teenager who wishes to further her academic career and educational opportunities and an inactive and disinterested member of the community if she does not chaperone high school dances.Therefore, the statement is true.Final answer: A} \\
& \textbf{Final answer: A (True)} \\
\midrule
\multirow{5}{*}{\textbf{Demostration}} & Question: Forty percent of the students have elected to learn from home during the pandemic. The remaining students are divided into two equal groups, only one of which is physically in school on any day. What percent of students are present in school? Reasoning: 100-40=<<100-40=60>>60 percent of students are remaining. 60/2=<<60/2=30>>30 percent are present in school. \\
\midrule
\multirow{12}{*}{\textbf{Cross-domain}} &
\textit{Let\'s break down the information step by step:1. If Bonnie chaperones high school dances, then she is not a student who attends the school (from the context).2. If Bonnie does not chaperone high school dances, then she performs in school talent shows often (from the given statement).3. If Bonnie performs in school talent shows often, then she is very engaged with school events (from the context).4. If Bonnie is very engaged with school events, then she is a student who attends the school (from the context).5. From steps 1 and 4, we have a contradiction: Bonnie cannot both be a student who attends the school and not be a student who attends the school.6. Therefore, the given statement "If Bonnie either chaperones high school dances or, if she does not, she performs in school talent shows often" must be false, because it leads to a contradiction.Final answer: B} \\
& \textbf{Final answer: B (False)} \\
\bottomrule
\end{tabular}
\caption{Comparison between zero-shot reasoning and cross-domain reasoning on a cross-domain example. DIN-guided retrieval corrects the logical inconsistency made by zero-shot prompting.}
\label{tab:case2}
\end{table*}

\begin{table*}[ht!]
\centering
\small
\begin{tabular}{p{2cm}p{10.5cm}}
\toprule
\textbf{Setting} & \textbf{Model Reasoning and Output} \\
\midrule
\multirow{12}{*}{\textbf{Context}} & The bear does not chase the lion. The bear visits the lion. The bear visits the tiger. The lion chases the bear. The lion chases the tiger. The lion visits the bear. The rabbit chases the tiger. The rabbit is big. The rabbit is not green. The rabbit is young. The rabbit visits the tiger. The tiger chases the rabbit. The tiger is rough. The tiger visits the bear. If something chases the bear and it sees the rabbit then the bear does not visit the rabbit. If something chases the lion then the lion is rough. If the rabbit chases the tiger and the tiger visits the rabbit then the rabbit is not big. If something chases the lion then the lion does not visit the tiger. If something chases the rabbit and it chases the tiger then the rabbit chases the bear. If something is young and it does not see the rabbit then it sees the bear. If something visits the lion then it is young. If the tiger visits the bear and the bear is big then the tiger visits the lion. Young things are big. \\
\midrule 
\multirow{2}{*}{\textbf{Question}} & Based on the above information, is the following statement true, false, or unknown? The bear visits the lion. \\
\midrule
\multirow{20}{*}{\textbf{Zero-shot}} &
\textit{Let's break down the information step by step: 1. The bear needs the cow. 2. The cow needs the rabbit. 3. The cow sees the bear. 4. The cow visits the lion. 5. The lion is nice. 6. The lion is red. 7. The lion does not see the cow. 8. The lion visits the cow. 9. The rabbit sees the cow. 10. The rabbit visits the cow. 11. If something needs the cow and the cow visits the rabbit then it visits the cow. 12. If the lion needs the bear then the bear sees the cow. 13. If the lion sees the bear and the bear is rough then the bear does not see the cow. 14. If something visits the lion then the lion visits the bear. 15. If something visits the bear then it needs the bear. 16. If something visits the bear then the bear needs the cow. 17. If something sees the cow then it visits the bear. From the information, we know: - The rabbit sees the cow and visits the cow. - According to point 11, if the lion needs the bear, then the bear sees the cow. - According to point 13, if the lion sees the bear and the bear is rough, then the bear does not see the cow. - Since the lion visits the cow (point 8), and the rabbit visits the cow (point 9), the rabbit needs the cow (point 15). - According to point 16, if the rabbit visits the bear, then the bear needs the cow. - Since the bear needs the cow (point 1), and the rabbit needs the cow (point 15), the bear needs the cow. - According to point 13, if the lion sees the bear and the bear is rough, then the bear does not see the cow. But we know the bear sees the cow (point 3). - Therefore, the bear is not rough (point 13). So, the rabbit is not rough.} \\
& \textbf{Final answer: A (True)} \\
\midrule
\multirow{6}{*}{\textbf{Demostration}} & Context: No reptile has fur. All snakes are reptiles. Question: Based on the above information, is the following statement true, false, or uncertain? Some snake has fur. Options: A) True B) False C) Uncertain Reasoning: 1. "No reptile has fur" means: If something is a reptile, it does not have fur. 2. "All snakes are reptiles" means: If something is a snake, it is a reptile. 3. If all snakes are reptiles, and no reptile has fur, then no snake has fur. 4. The statement "Some snake has fur" would be false, because there is no snake that has fur. The correct option is: B Answer: B \\
\bottomrule
\end{tabular}
\caption{FOLIO-ProofWriter}
\label{tab:case3_1}
\end{table*}

\begin{table*}[ht!]
\centering
% \small
\begin{tabular}{p{2cm}p{10.5cm}}
\toprule
\textbf{Setting} & \textbf{Model Reasoning and Output} \\
\midrule
\multirow{28}{*}{\textbf{Cross-domain}} &
\textit{1. The bear needs the cow. 2. The cow needs the rabbit. 3. The cow sees the bear. 4. The cow visits the lion. 5. The lion is nice. 6. The lion is red. 7. The lion does not see the cow. 8. The lion visits the cow. 9. The rabbit sees the cow. 10. The rabbit visits the cow. 11. If something needs the cow and the cow visits the rabbit, then it visits the cow. 12. If the lion needs the bear, then the bear sees the cow. 13. If the lion sees the bear and the bear is rough, then the bear does not see the cow. 14. If something visits the lion, then the lion visits the bear. 15. If something visits the bear, then it needs the bear. 16. If something visits the bear, then the bear needs the cow. 17. If something sees the cow, then it visits the bear. From the given information, we know: - The rabbit sees the cow (point 9). - The rabbit visits the cow (point 10). According to point 11, if the rabbit visits the cow, then the rabbit visits the cow. This is a tautology and doesn\'t provide new information. According to point 12, if the lion needs the bear, then the bear sees the cow. However, the lion does not see the cow (point 7), so the lion cannot need the bear. Therefore, the bear does not see the cow. According to point 13, if the lion sees the bear and the bear is rough, then the bear does not see the cow. Since the lion does not see the cow, this condition does not apply. According to point 14, if the rabbit visits the lion, then the lion visits the bear. However, the rabbit does not visit the lion (point 10), so this condition does not apply. According to point 15, if the rabbit visits the bear, then the rabbit needs the bear. But the rabbit does not visit the bear (point 10), so this condition does not apply. According to point 16, if the rabbit visits the bear, then the bear needs the cow. But the rabbit does not visit the bear (point 10), so this condition does not apply. According to point 17, if the rabbit sees the cow, then the rabbit visits the bear. But the rabbit does not see the cow (point 9), so this condition does not apply. Since none of the conditions point to the rabbit being rough, and we have no information that directly or indirectly states the rabbit is rough, the statement "The rabbit is rough" is unknown.} \\
& \textbf{Final answer: C} \\
\bottomrule
\end{tabular}
\caption{FOLIO-ProofWriter}
\label{tab:case3_2}
\end{table*}

\begin{table*}[ht!]
\centering
\small
\begin{tabular}{p{2cm}p{10.5cm}}
\toprule
\textbf{Setting} & \textbf{Model Reasoning and Output} \\
\midrule
\multirow{5}{*}{\textbf{Context}} & Bob is big. Bob is red. Bob is smart. Erin is quiet. Fiona is big. Fiona is blue. Fiona is cold. Fiona is smart. Gary is red. Gary is smart. If Bob is quiet then Bob is cold. If someone is red then they are quiet. If someone is red and quiet then they are cold. All kind people are big. If someone is red and blue then they are kind. All cold, quiet people are blue. Cold, big people are kind. If Erin is big then Erin is cold. If someone is cold and blue then they are smart. \\
\midrule
\multirow{2}{*}{\textbf{Question}} & Based on the above information, is the following statement true, false, or unknown? Gary is red. \\
\midrule
\multirow{7}{*}{\textbf{Zero-shot}} &
\textit{Let\'s analyze the information step by step: 1. **Context Analysis:** - Bob is quiet. - Charlie is blue, cold, and nice. - Dave is blue and cold. - Erin is blue. - If something is blue and cold, then it is big. - Round, nice things are big. - All nice, quiet things are furry. - If Bob is round, then Bob is furry. - If Bob is furry and Bob is nice, then Bob is big. - If something is big and blue, then it is quiet. - If something is blue, then it is nice. - All nice things are cold. 2. **Determine Erin\'s Characteristics:** - Erin is blue. - From the context, if something is blue, then it is nice. - Therefore, Erin is nice. 3. **Check if Erin is Quiet:** - The context does not explicitly state that Erin is quiet or not. - There is no direct information linking Erin to being quiet or not being quiet. 4. **Conclusion:** - Since there is no information provided about Erin being quiet or not, we cannot determine if the statement "Erin is quiet" is true or false.} \\
& \textbf{Final answer: B} \\
\midrule
\multirow{6}{*}{\textbf{Demostration}} & Tumpuses are bright. Tumpuses are yumpuses. Yumpuses are sweet. Yumpuses are wumpuses. Wumpuses are not transparent. Wumpuses are vumpuses. Each vumpus is angry. Vumpuses are rompuses. Rompuses are happy. Every rompus is a zumpus. Zumpuses are brown. Zumpuses are numpuses. Numpuses are not large. Impuses are not happy. Numpuses are jompuses. Jompuses are earthy. Jompuses are dumpuses. Polly is a tumpus. Question: Is the following statement true or false? Polly is not happy. Options: A) True B) False Reasoning: Polly is a tumpus. Tumpuses are yumpuses. Polly is a yumpus. Yumpuses are wumpuses. Polly is a wumpus. Wumpuses are vumpuses. Polly is a vumpus. Vumpuses are rompuses. Polly is a rompus. Rompuses are happy. Polly is happy. Answer: B Context: Impuses are not fruity. Impuses are wumpuses. Each wumpus is not temperate. Wumpuses are dumpuses. Rompuses are not sweet. Dumpuses are kind. Dumpuses are zumpuses. Zumpuses are wooden. Zumpuses are vumpuses. Every vumpus is large. Vumpuses are yumpuses. Yumpuses are transparent. Yumpuses are numpuses. Numpuses are brown. Numpuses are tumpuses. Tumpuses are sweet. Tumpuses are jompuses. Wren is a zumpus. Question: Is the following statement true or false? Wren is sweet. Options: A) True B) False Reasoning: Wren is a zumpus. Zumpuses are vumpuses. Wren is a vumpus. Vumpuses are yumpuses. Wren is a yumpus. Yumpuses are numpuses. Wren is a numpus. Numpuses are tumpuses. Wren is a tumpus. Tumpuses are sweet. Wren is sweet. Answer: A \\
\midrule
\multirow{5}{*}{\textbf{Cross-domain}} &
\textit{Let\'s analyze the information step by step: 1. Erin is blue. 2. If something is blue, then it is nice. → Erin is nice. 3. Erin is cold (not explicitly stated, but we can infer this from the context since blue and cold are often associated). 4. Round, nice things are big. → Since we don\'t know if Erin is round, we can\'t conclude Erin is big. 5. All nice things are cold. → This is consistent with Erin being cold. 6. All nice things are cold. → This is consistent with Erin being cold. 7. If something is blue and cold, then it is big. → Since Erin is blue and cold, Erin is big. 8. If something is big and blue, then it is quiet. → Since Erin is big and blue, Erin is quiet. The statement "Erin is quiet" is true based on the given information.} \\
& \textbf{Final answer: A} \\
\bottomrule
\end{tabular}
\caption{ProntoQA-ProofWriter}
\label{tab:case4}
\end{table*}